\documentclass[10pt,twocolumn,letterpaper]{article}

\usepackage[pagenumbers]{cvpr} % To force page numbers, e.g. for an arXiv version

\definecolor{cvprblue}{rgb}{0.21,0.49,0.74}
\usepackage[pagebackref,breaklinks,colorlinks,allcolors=cvprblue]{hyperref}
\usepackage{times}
\usepackage{epsfig}
\usepackage{graphicx}
\usepackage{amsmath}
\usepackage{amssymb}
\usepackage{booktabs}
\usepackage{appendix}
\usepackage{multirow}
\usepackage{bbding}
\usepackage{makecell}
\usepackage{bm}
\usepackage[accsupp]{axessibility}

\newcommand{\tight}[1]{\hspace{-.7mm}#1\hspace{-.7mm}}

\def\paperID{241} % *** Enter the Paper ID here
\def\confName{CVPR}
\def\confYear{2025}

\title{AnyEdit: Mastering Unified High-Quality Image Editing for Any Idea}
\author{Qifan Yu$^1$\footnotemark[1] \quad Wei Chow$^1$\footnotemark[1]\quad  Zhongqi Yue$^2$\footnotemark[1] \quad Kaihang Pan$^1$ \quad Yang Wu$^3$ \quad Xiaoyang Wan$^1$ \\Juncheng Li$^1$\footnotemark[2]\quad Siliang Tang$^1$\quad Hanwang Zhang$^2$\quad Yueting Zhuang$^1$\footnotemark[2]\\
\small $^1$Zhejiang University, $^2$Nanyang Technological University, $^3$Ant Group\\
{\tt\small \{yuqifan, 3210103790, kaihangpan, junchengli, siliang, yzhuang\}@zju.edu.cn}\\
{\tt\small nickyuezhongqi@gmail.com, wy306396@antgroup.com}\\\\
\large\url{https://dcd-anyedit.github.io/}}

\begin{document}

\maketitle
\renewcommand{\thefootnote}{\fnsymbol{footnote}} %将脚注符号设置为fnsymbol类型，即特殊符号表示
\footnotetext[1]{Equal Contribution.} %对应脚注[1]
\footnotetext[2]{Corresponding Author.}
\renewcommand{\thefootnote}{\arabic{footnote}}
\begin{figure*}[t]
    \centering
    \vspace{-2mm}
    \includegraphics[width=1.\linewidth]{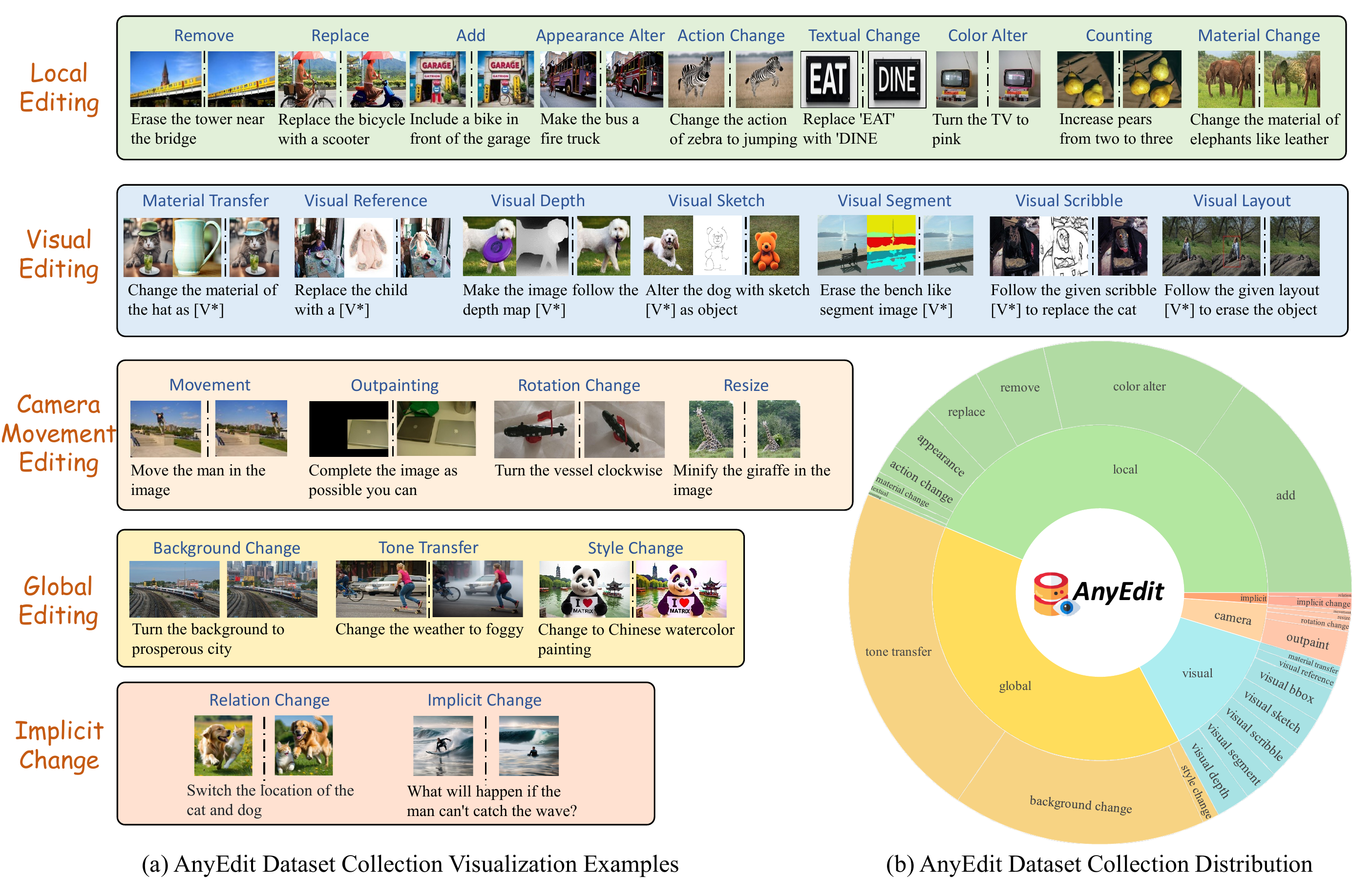}
    \vspace{-5mm}
    \caption{Examples of \textbf{AnyEdit} at scale. We comprehensively categorize image editing tasks into 5 groups based on different editing capabilities: (a) \textit{Local Editing} which focuses on region-based editing~({\color{LimeGreen}green area}); (b) \textit{Global Editing} which focuses on the full range of image rendering~({\color{Goldenrod}yellow area}); (c) \textit{Camera Move Editing} which focuses on viewpoints changing instead of scenes~({\color{gray}gray area}); (d) \textit{Implicit Editing} which requires commonsense knowledge to complete complex editing~({\color{orange}orange area}); (e) \textit{Visual Editing} which encompasses additional visual inputs, addressing the requirements for multi-modal editing~({\color{ProcessBlue}blue area}).} 
    \vspace{-5mm}
    \label{example}
\end{figure*}
\begin{abstract}
Instruction-based image editing aims to modify specific image elements with natural language instructions. However, current models in this domain often struggle to execute complex user instructions accurately, as they are trained on low-quality data with limited editing types. We present \textbf{AnyEdit}, a comprehensive multi-modal instruction editing dataset, comprising 2.5 million high-quality editing pairs spanning over 20 editing types and five domains. We ensure the diversity and quality of the AnyEdit collection through three aspects: initial data diversity, adaptive editing process, and automated selection of editing results. Using the dataset, we further train a novel AnyEdit Stable Diffusion with task-aware routing and learnable task embedding for unified image editing. Comprehensive experiments on three benchmark datasets show that AnyEdit consistently boosts the performance of diffusion-based editing models. This presents prospects for developing instruction-driven image editing models that support human creativity. 
\end{abstract}    
\section{Introduction}
\label{sec:intro}
With the development of multi-modal datasets~\cite{lin2014microsoft,schuhmann2022laion} and generative frameworks~\cite{goodfellow2014generative,dinh2014nice}, recent text-to-image generation~(T2I) models~\cite{betker2023improving,nichol2021glide,ramesh2022hierarchical,rombach2022high,saharia2022photorealistic, yu2023interactive, yang2024mastering, bai2024meissonic} have shown promising results in generating high-fidelity photo-realistic images. To enhance the controllability of generated images, image editing~\cite{zhu2017unpaired,chen2018language,ling2021editgan,crowson2022vqgan,sohn2023styledrop,hertz2022prompt} aims to modify desired target elements and retain unrelated contents, which has achieved significant improvements for T2I models. 
More recently, instruction-based image editing~\cite{hertz2022prompt,brooks2023instructpix2pix,cao2023masactrl, crowson2022vqgan, yang2024mastering} pave the way to conveniently edit images using natural language instructions without complex descriptions or region-specific masks. However, the scarcity of high-quality instruction editing datasets makes it difficult to develop powerful instruction-guided image editing models.
% Compared to text-to-image generation, instruction-based image editing is more challenging because it requires a comprehensive understanding of both linguistic and visual content. 
% Therefore, it is crucial to build a high-quality, large-scale instruction editing dataset for enhancing the capabilities of instruction-guided image editing models. 

Despite several prior works~\cite{brooks2023instructpix2pix, ge2024seed,sheynin2024emu,zhao2024ultraedit,zhang2024magicbrush,yang2024editworld,fu2023guiding} showcasing promising instruction-following editing datasets and achieving effective results, they still struggle with drawbacks due to limited editing types and low data quality.
% , which hinder the model's ability to robustly execute a wide variety of editing tasks.
These methods lack support for editing instructions based on complex multi-modal perception~(\textit{e.g.}, spatial composition~\cite{zhou2024migc}, viewpoint~\cite{yu2023mvimgnet}, commonsense understanding~\cite{bitton2023breaking}) and reference-based editing instructions~(\textit{e.g.}, customization~\cite{chen2024anydoor, yang2023paint}, image transfer~\cite{zhang2023adding}), limiting the model's ability to understand interleaved editing instructions and robustly execute a wide variety of editing tasks. 
\
To address the above limitations, this work, \textbf{AnyEdit}, first provides a comprehensive catalog of editing instructions and systematically introduces a unified editing framework for various editing tasks.
% which contains diverse instruction distribution and categorizes the instructions grounded by comprehensive scenarios. 
As shown in figure~\ref{example}, AnyEdit is divided into five classes based on editing capabilities, with each class containing multiple task types. Regarding the general editing scenarios in previous work~\cite{zhang2024magicbrush,zhao2024ultraedit,ge2024seed}, AnyEdit categorizes them into \textbf{Local Editing} and \textbf{Global Editing} based on their granularity to construct more precise instructions. 
To accommodate the editing intentions for spatial perception, we further include \textbf{Camera Movement Editing}. For commonsense understanding, we introduce \textbf{Implicit Editing}, which replaces direct intentions with more complex instructions that require imaginative thinking. In addition, to accommodate reference inputs from the visual modality, we incorporate \textbf{Visual Editing} to construct multi-modal editing instructions.

Although AnyEdit has enriched editing types, there exist three principal challenges for high-quality dataset collection:
(1) First, we observe that current editing datasets overlook inherent data biases when preparing initial image-text pairs~\cite{friedrich2023fair, yu2023visually, naik2023social}, leading to an imbalance of concepts within the dataset. For instance, ``\textit{change the tree in the park to a camel}” may bring undesired sand because ``sand” and ``camel” frequently co-occur in the training data. Thus, we introduce counterfactual synthetic scenes to enrich real-world data with tail concepts and their diverse combinations, achieving a balanced distribution of concepts in the dataset and enhancing generalization in editing tasks. (2) Secondly, current data collection methods struggle to accommodate diverse input formats and editing requirements due to the shared pipeline for all instructions. Therefore, we propose an adaptive editing pipeline that automatically selects the appropriate data pipeline for each task. 
% allowing AnyEdit to generate high-quality edited data with fine-grained precision control. 
(3) Third, there exist inherent drawbacks in the collected triplets of original images, edit instructions, and edited images due to the randomness of generation~(\textit{i.e.}, low resolution~\cite{schuhmann2022laion}, high noise~\cite{hasan2022noise}, and misalignment between text and images~\cite{thrush2022winoground,wu2023human}). 
% for more granular editing instructions. 
Thus, we develop an instruction validation pre-filter and image assessment post-filter strategy to filter out unsatisfactory results. In this way, AnyEdit automatically comprises 2.5M high-quality editing data across 25 distinct editing types~(c.f., Tab.~\ref{table1}), showing a 28.9\% improvement in visual similarity and 18.8\% in semantic similarity over SOTA datasets~(c.f., Tab.~\ref{dataset-statistic}). This automated process opens up the possibility of scaling the high-quality editing dataset in a low-resource manner.

% However, the quality of images in existing real-image datasets is inconsistent, exhibiting inherent deficiencies such as low resolution, high noise, and misalignment between images and textual descriptions, which impede the development of complex editing instructions. By incorporating synthetic images, we obtain more precise image representations for detailed editing, facilitating the generation of higher-quality complex editing instructions (e.g., action changes, implicit changes). This approach not only ensures the model’s ability to generalize and adapt to user intentions but also enhances the quality of the complex editing instructions produced.
% Furthermore, we observe that the widespread use of T2I models to generate source and target images~\cite{friedrich2023fair, naik2023social}, as well as the use of general real image data~\cite{lin2014microsoft, liu2024improved}, exhibits significant domain bias. This results in the image editing dataset being potentially quite unbalanced and incomplete. 
\begin{table*}[!t]
\centering
\vspace{-0.3cm}
\resizebox{0.98\textwidth}{!}{
    \begin{tabular}{l|cc|ccccc|ccc|c}
    \hline
    \multirow{2}{*}{Editing Dataset} & \multirow{2}{*}{\#Size} & \multirow{2}{*}{\#Types} &\multicolumn{5}{c|}{Instruction Type} & \multicolumn{3}{c|}{Target Scenario} & \multirow{2}{*}{Open Source } \\

   && & Local  & Global  &  Camera Movement & Implicit  & Visual         & Real Image & Synthetic Image & Synthetic Scene &         \\ \hline
    MagicBrush~\cite{zhang2024magicbrush}    & 10K     & 5     & {\color{ForestGreen}\checkmark}   & {\color{ForestGreen}\checkmark} & {\color{red}$\times$} & {\color{red}$\times$} & {\color{red}$\times$} & {\color{ForestGreen}\checkmark}  & {\color{red}$\times$} & {\color{red}$\times$}  & {\color{ForestGreen}\checkmark} \\
    InstructPix2Pix~\cite{brooks2023instructpix2pix} & 313K    &4     & {\color{ForestGreen}\checkmark}   & {\color{ForestGreen}\checkmark} & {\color{red}$\times$} & {\color{red}$\times$} & {\color{red}$\times$} & {\color{red}$\times$}  & {\color{ForestGreen}\checkmark} & {\color{red}$\times$}  & {\color{ForestGreen}\checkmark} \\
    EMU-Edit~\cite{sheynin2024emu}       & -     & 8       & {\color{ForestGreen}\checkmark}   & {\color{ForestGreen}\checkmark} &  {\color{red}$\times$} & {\color{ForestGreen}\checkmark} &
    {\color{red}$\times$} & {\color{red}$\times$}  & {\color{ForestGreen}\checkmark} & {\color{red}$\times$}  & {\color{red}$\times$} \\
    HQ-Edit~\cite{hui2024hq}        & 197K   & 6      & {\color{ForestGreen}\checkmark}   & {\color{ForestGreen}\checkmark} & {\color{red}$\times$} & {\color{red}$\times$} & {\color{red}$\times$} & {\color{red}$\times$}  & {\color{ForestGreen}\checkmark} & {\color{red}$\times$}  & {\color{ForestGreen}\checkmark} \\
    SEED-Data-Edit~\cite{ge2024seed}  & 3.7M     & 6    & {\color{ForestGreen}\checkmark}   & {\color{ForestGreen}\checkmark} & {\color{ForestGreen}\checkmark} & {\color{red}$\times$} & {\color{red}$\times$} & {\color{ForestGreen}\checkmark}  & {\color{ForestGreen}\checkmark} & {\color{red}$\times$}  & {\color{ForestGreen}\checkmark} \\
    EditWorld~\cite{yang2024editworld}      & 8.6K     &1    & {\color{red}$\times$}   & {\color{red}$\times$} & {\color{red}$\times$} & {\color{ForestGreen}\checkmark} & {\color{red}$\times$} & {\color{ForestGreen}\checkmark}  & {\color{red}$\times$} & {\color{red}$\times$}  & {\color{ForestGreen}\checkmark} \\
    UltraEdit~\cite{zhao2024ultraedit}       & 4M    & 9       & {\color{ForestGreen}\checkmark}   & {\color{ForestGreen}\checkmark} & {\color{red}$\times$} & {\color{red}$\times$} & {\color{red}$\times$}  & {\color{ForestGreen}\checkmark} & {\color{ForestGreen}\checkmark}  & {\color{red}$\times$} & {\color{ForestGreen}\checkmark}\\\hline
    AnyEdit         & 2.5M    & \textbf{25}       & {\color{ForestGreen}\checkmark}   & {\color{ForestGreen}\checkmark} & {\color{ForestGreen}\checkmark} & {\color{ForestGreen}\checkmark}   & {\color{ForestGreen}\checkmark}   & {\color{ForestGreen}\checkmark} & {\color{ForestGreen}\checkmark} & {\color{ForestGreen}\checkmark}  & {\color{ForestGreen}\checkmark} \\ \hline
    \end{tabular}
    }
    \caption{Comparison of existing image editing datasets. ``Real Image” means the original images are from real world, ``Synthetic Image" means they are from T2I models, ``Synthetic Scene” indicates both images and captions are generated to address the inherent data bias.}
    \vspace{-0.5cm}
    \label{table1}
\end{table*}

% However, these distinct pipelines are computationally time-consuming and cannot be directly used for instruction-based editing. 
% As AnyEdit has comprised diverse high-quality editing data, we further effectively develop a powerful instruction-based editing model for any modifying idea from humans based on it. Specifically, we propose a novel AnyEdit Stable Diffusion approach~(\textbf{AnySD}) with task-aware routing and learnable task embedding for unified image editing. 
To fully harness the potential of AnyEdit's high-quality editing data, we propose a novel AnyEdit Stable Diffusion approach (\textbf{AnySD}) that employs task-aware routing and learnable task embeddings to support various editing types within AnyEdit, thereby constructing a robust instruction-based editing model for handling diverse editing requests. The task-aware routing mechanism leverages AnyEdit’s diverse data to adjust the granularity of editing (\textit{e.g.}, local object or global style), while the learnable task embedding coordinates the inherent complexity across tasks within AnyEdit. In this way, AnySD fully harnesses the comprehensive and diverse data from AnyEdit to set new records on the MagicBrush and Emu-Edit benchmarks. 

Furthermore, to facilitate a comprehensive evaluation of instruction-based image editing, we manually curate a challenging benchmark, \textbf{AnyEdit-Test}, from our automatically collected AnyEdit dataset, comprising 1,250 high-quality editing pairs in diverse scenarios, across 25 distinct categories (50 per category). We notice that (1) Existing models often fail to properly modify or cause distortions in complex tasks (\textit{e.g.}, action change shown in Fig.~\ref{fig:main_case_part1}), exposing the limitations of current benchmarks for complex tasks. (2) Even for common tasks in AnyEdit-Test, some previous SOTA models show a notable performance drop compared to existing benchmarks (\eg, DINO’s decline in global editing in Tab.~\ref{table:2}), revealing the limitations of current benchmarks in multi-scene editing. This highlights the critical role of AnyEdit-Test in thoroughly assessing the capabilities of editing models. 
We will release it for public research.

% The task-aware routing mechanism controls the granularity of editing~(\textit{e.g.}, local object or global style) based on different tasks, while the learnable task embedding coordinates task complexities to achieve unified editing. 
% Equipped with an adaptive architecture trained across a wide range of editing types, AnySD achieves new records on the challenging MagicBrush and Emu-Edit benchmarks. 

Our main contributions are summarized as follows:
\begin{itemize}
    % \item To the best of our knowledge, AnyEdit is the most comprehensive editing dataset released to the community, encompassing not only general instructions but also more challenging abstract and multi-modal editing instructions.
    % \item We systematically categorize various instruction-based editing instructions in a novel perspective, incorporating more challenging camera movement, implicit editing, and visual editing instructions, which broaden their applicability and robustness in real-world scenarios.
    % \item We automatically curate the most comprehensive editing instruction dataset \textbf{AnyEdit} by leveraging task-adaptive pipelines in both realistic world and synthetic scenarios and propose AnySD to fully verify the potential of AnyEdit in developing a powerful image editing model.
    \item We systematically categorize various editing instructions in a novel perspective and innovatively introduce a unified editing framework that leverages an adaptive editing pipeline to automatically collect diverse high-quality editing data across different scenarios in a scalable way.
    \item We construct a multi-type, multi-scene dataset, AnyEdit, and its corresponding benchmark AnyEdit-Test for instruction-based editing in various scenarios, comprising 25 distinct and complex editing types to accommodate broader requirements of real-world editing.
    % \item Experiments over three benchmarks confirm that, with AnyEdit, AnySD shows steady improvement in instruction consistency and image fidelity on diverse editing types.   
    \item We use our proposed AnySD to fully unlock the potential of AnyEdit, achieving SOTA improvements in instruction adherence and image fidelity across diverse editing types.
    % , as demonstrated by quantitative and qualitative comparisons on three benchmarks.
 %    canonical diffusion-based image editing models can achieve more precise and robust editing capabilities on diverse tasks compared to existing models.
 % AnySD shows steady improvement in instruction consistency and content fidelity on diverse editing types
\end{itemize}
\section{Related Work}
% \begin{table*}[!t]
% \resizebox{0.99\textwidth}{!}{
% \begin{tabular}{lclccc|ccc}
% \hline
% \toprule
% \multicolumn{2}{c}{\multirow{2}{*}{Model Type}} &\multicolumn{1}{c}{\multirow{2}{*}{Methods}} & \multicolumn{3}{c}{Instance-level} &\multicolumn{3}{c}{Sentence-level} \\\cline{4-6} \cline{7-9}

% \noindent\textbf{Image Editing Dataset.} 
Table~\ref{table1} compares the existing Instruction-based image editing datasets including InstructPix2Pix~\cite{brooks2023instructpix2pix},
MagicBrush~\cite{zhang2024magicbrush}, EMU-Edit~\cite{sheynin2024emu}, HQ-Edit~\cite{hui2024hq}, SEED-Data-Edit~\cite{ge2024seed}, EDITWORLD~\cite{yang2024editworld}, and UltraEdit~\cite{zhao2024ultraedit}. Among these, the first four datasets use automatic methods to collect data, while the latter three incorporate human curation. InstructPix2Pix~\cite{brooks2023instructpix2pix} utilizes P2P~\cite{hertz2022prompt} for image editing, which makes it only suitable for synthetic images. MagicBrush~\cite{zhang2024magicbrush} hires crowd workers to annotate images from the MSCOCO~\cite{lin2014microsoft} dataset manually but only includes 10K editing pairs due to expensive labor expenses. HQ-Edit\cite{hui2024hq} also lacks fine-grained details and realism due to its diptych generation though it exploits GPT-4V~\cite{achiam2023gpt} and DALL-E~\cite{betker2023improving} to enhance descriptions. Both of them are lack of generalization. 
% 得益于多模态理解与生成的显著进展, 
% SEED-Data-Edit~\cite{ge2024seed} and UltraEdit~\cite{zhao2024ultraedit} have been proposed to enrich the types and turns of image editing recently. 
With the advancements in multi-modal understanding~\cite{li2023variational, li2023fine, lin2024vila, li2022fine, pan2024towards, yu2024hallucidoctor, pan2024i3, pan2023self} and generation~\cite{pan2024auto, chow2024unified, ge2024seedx, xiao2024omnigen}, SEED-Data-Edit~\cite{ge2024seed} and UltraEdit~\cite{zhao2024ultraedit} have been recently introduced to enrich the types and turns of image editing.
However, they only focus on traditional editing tasks, neglecting more complex and customized editing needs. Although EMU-Edit~\cite{sheynin2024emu} and EditWorld~\cite{yang2024editworld} endeavor to simulate physical world editing, they lack a unified definition of implicit editing. Moreover, these datasets fail to incorporate viewpoint changes and support multi-modal inputs. 
In AnyEdit, we combine five distinct groups of data, covering 25 editing types, which will be released to help the community. It is worth noting that AnyEdit is the only dataset that considers the data bias and introduces counterfactual synthetic scenes to balance the distribution of the dataset.
\begin{table}[!t]
    % \vspace{-0.1cm}
    \centering
    \resizebox{0.45\textwidth}{!}{
        \begin{tabular}{lcccc}
        \hline
        \textbf{Dataset}            & \textbf{\#Samples} & \textbf{\#Concepts} & \textbf{License}     & \textbf{Annotator} \\ \hline
       \multicolumn{5}{c}{\textbf{Real World Image Caption Paired Dataset}}\\\hline
        MS COCO~\cite{lin2014microsoft}              & 123,287 &  80           & CC BY 4.0            & Human             \\
         MVImgNet~\cite{yu2023mvimgnet}              & 31,783  & 238          & CC BY 4.0            & Human \\
         LLaVA-CC3M-Pretrain~\cite{liu2024visual}         & 519,176 & 31423             & Custom               & GPT-4V             \\\hline
       \multicolumn{5}{c}{\textbf{ Counterfactual Synthetic Scene Pair Dataset}} \\\hline
         AnyEdit-Composition & 22,936         & 500    & Custom               &  Composition            \\ \hline
        \end{tabular} 
    }
    \vspace{-0.1cm}
    \caption{Data preparation details for AnyEdit dataset collection.}
    \label{dataset-table}
    \vspace{-0.6cm}
\end{table}
\section{AnyEdit}
\begin{figure*}[t]
    \centering
    \includegraphics[width=1.\linewidth]{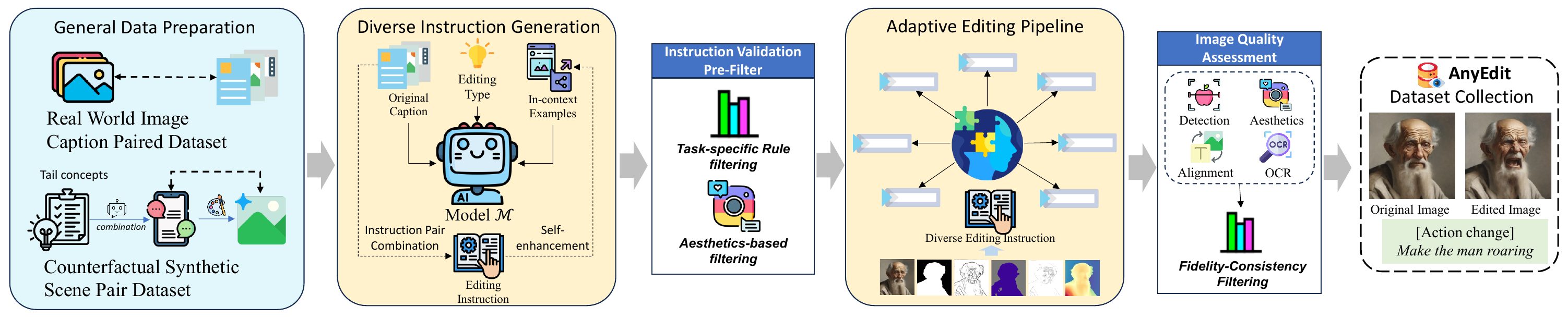}
    
  % \begin{subfigure}{1.\linewidth}
  %   \centering
  %   \includegraphics[width=1.\linewidth]{figures/total_pipeline.pdf}
  %   \caption{Base SGG}
  %   \label{sg2}
  % \end{subfigure}
    \caption{The comprehensive construction pipeline of AnyEdit. We summarize the general pipeline into five steps: (1) General data preparation from real-world image-text pairs and synthetic scenes. (2) Diverse instruction generation using LLM to produce high-quality editing instructions. (3) Pre-filtering for instruction validation. (4) Adaptive editing pipeline tailors specific editing methods for each edit type to generate high-quality edited images. (5) Image quality assessment ensures high-quality editing pairs for the AnyEdit Dataset.}

  \vspace{-4mm}  
  \label{pipeline}
\end{figure*}
% As shown in Fig.~\ref{example}(a), AnyEdit innovatively comprises five promising data groups with 20+ distinct editing types for completeness. 
% An overview of our construction pipeline is in Fig.~\ref{pipeline}. First, we formulate the original data preparation from real-world image-text pairs and synthetic scenes for concept balance~(\S~\ref{3.2}). Then, we focus on the diverse instruction generation~(\S~\ref{3.3}) and the adaptive editing pipeline incorporated with task-specific editing types~(\S~\ref{3.4}). Finally, we introduce instruction validation pre-filter and image assessment post-filter strategy for quality enhancement~(\S~\ref{3.5}). Next, we detailedly introduce each step.

% describe the diverse instruction generation process and the corresponding pre-filter strategy used to select valid original images. In Sec.~\ref{3.3}, we elaborate on each automated pipeline for precise image editing according to distinct tasks, and in Sec.~\ref{3.4} the post-filter strategy based on image quality assessment. In Sec.~\ref{3.5}, we introduce extra counterfactual image augmentation to balance the distribution of editing dataset.

% 1. 关于数据集文章整体的架构应该怎么展开，感觉现在这样子有点偏工程。文章定位是什么？现在合适的好像只有cvpr的，但是这类文章投cvpr您这里在写作布局上有什么建议吗？
% 2. 关于这一块在您看来方法层面有什么有意思且比较好入手的方法改进吗？因为我感觉只有数据集，现在的方法只是最baseline的做法，我感觉影响有限
% 3. 7月份有一篇很相近的papar发布了一个4M的editing数据集，它们主要做了tranditional的编辑和region-based（mask）的编辑，想听听您的意见如何和它们区别开

\subsection{Editing Type Definition}
\label{3.1}
To equip instruction-based editing models with comprehensive capabilities to follow any creative ideas, we compiled a multi-modal image editing dataset \textbf{AnyEdit} for instruction-based image editing, consisting of 2.5M high-quality editing pairs across five primary domains, as illustrated in Fig.~\ref{example}.
The dataset comprises editing tasks divided into five main categories, each containing various editing types: (1) Local editing: add, remove, replace, color alter, appearance alter, material change, action change, textual change, and counting. (2) Global editing: background change, tone transfer, and style change. (3) Camera movement editing: rotation change, out-painting, movement, and resize. (4) Implicit editing: implicit change and relation change. (5) Visual editing: visual reference, material transfer, and visual conditions~(\textit{i.e.}, depth, segmentation, scribble, sketch, mask). Specifically, local editing targets specific areas of an image without altering unrelated semantic content, while global editing affects the entire image. Camera movement editing extends this concept by manipulating the viewpoint of specific objects or the whole content within the scene. 
Furthermore, implicit editing involves hidden changes in state or interaction patterns that require comprehension. Visual editing, on the other hand, incorporates additional visual inputs as references alongside editing instructions. Figure~\ref{example}(a) shows examples of various editing types in each category and their detailed definitions are in Appendix ~\textcolor{red}{B.1}.
\subsection{Automatic Dataset Collection}
\noindent\textbf{General Data Preparation.} Previous studies have demonstrated that high-quality initial images facilitate the diversity of editing image creation~\cite{ge2024seed, zhao2024ultraedit}. To address real-world user demands for image editing in complex scenarios, we collect $\sim$680K real-world image-caption pairs from annotated datasets~(\textit{i.e.}, MSCOCO~\cite{lin2014microsoft}, LLaVA-CC3M-Pretrain~\cite{liu2024visual}) and multi-view image datasets~(\textit{i.e.}, MVImgNet~\cite{yu2023mvimgnet}).
Then, we enrich those brief captions by MLLMs (\textit{e.g.}, VILA~\cite{lin2024vila}) for the completeness of the descriptions. 
However, these image-text pairs suffer from inherent data bias, leading to the model falling short in domains not well covered by general knowledge~\cite{friedrich2023fair, naik2023social}. 
Thus, we introduce the Counterfactual Synthetic Scene Pair Dataset to balance the data distribution of the initial image-text pairs. Specifically, we collect infrequent tail concepts from internet data and combine multiple concepts and contexts to generate a description by LLaMA-3B~\cite{dubey2024llama}. Subsequently, we invoke off-the-shelf T2I models to produce the initial images. In this manner, we enrich the original dataset by incorporating rare concept combinations, resulting in $\sim$700K high-quality and diverse image-caption pairs for the AnyEdit dataset collection, as illustrated in Table~\ref{dataset-table}.

% Specifically, we collect infrequent tail concepts from internet data and manually filter them to retain the top 100 reasonable concepts, along with five corresponding contexts for each. We then combine multiple concepts and contexts to generate a description and invoke off-the-shelf T2I models to produce the initial images. In this way, we enriched the original dataset with rare concept combinations, ultimately obtaining 700K high-quality, diverse image-caption pairs. This drives the model to focus on the intended editing rather than prior co-occurrences, preventing undesired changes.

\noindent\textbf{Diverse Instruction Generation.} It aims to create varied editing instructions and corresponding edited caption outputs based on the caption of the initial image.
As illustrated in Figure~\ref{pipeline}, we leverage the public Llama3-8b~\cite{dubey2024llama} model to convert original captions to diverse editing instructions. To address the limitations in instruction diversity and consistency when generating editing instructions, we integrate intuitive type constraints with LLM generation and employ in-context examples to develop a task-specific agent tailored to each editing type. Furthermore, we integrate the generated editing instructions with the original captions to create instruction pairs, which serve as in-context examples for iterative self-enhancement, thereby gradually increasing the diversity and complexity of the instructions. Details of the prompt constraints and examples are in Appendix \textcolor{red}{B.2}.

\noindent\textbf{Adaptive Editing Pipelines.} Traditional instruction-editing datasets~\cite{sheynin2024emu, zhao2024ultraedit} rely on fixed pipelines~\cite{hertz2022prompt} or time-consuming manual filtering, making it difficult to efficiently generate high-quality edited images for complex editing types and various input formats. Here we propose an adaptive view of editing pipelines that customizes the edited image according to the specific type of editing. Specifically, we design 9 core pipelines for generating local, global, camera movement, implicit, and visual editing data. 
During image editing generation, we input editing instruction pairs along with the original images and their variants into the adaptive editing pipeline. The pipeline dynamically selects tailored solutions based on the editing type to generate images that align with the intended edits. Additionally, we incorporate extra constraints~(\textit{e.g.}, dilated masks, layouts, and geometry guidance) into the U-Net layers within the diffusion process to achieve more precise semantic alignment and artifact reduction.
Full details of the image construction process for each editing type can be found in Appendix \textcolor{red}{B.3}.

\noindent\textbf{Data Quality Enhancement.} Since the quality of editing data is critical for robust editing model training in AnyEdit, we further introduce a comprehensive filtering strategy for data quality enhancement. It consists of two steps:
\textit{instruction validation pre-filter} and \textit{image quality post-filter}.

\textit{(i) Instruction Validation Pre-filter.} 
We notice that partial editing instructions from LLMs sometimes introduce ambiguities that adversely impact the edited image (\textit{e.g.}, changing appearance in ``color alter" editing or changing action of the static desk in ``action change" editing). Simultaneously, low-quality initial images (\textit{e.g.}, low resolution, bad aspect ratio, lack of aesthetics) consistently lead to unsatisfactory editing results, even after time-consuming rounds of selection. Thus, we employ task-specific heuristic rules to validate various instructions, ensuring consistency and performing aesthetic evaluations to guarantee that aesthetically balanced images are used for the editing process.

\textit{(ii) Image Quality Post-filter.} After collecting all the editing results, we conducted an automatic data evaluation to filter out unsatisfactory images that did not meet the criteria for data generation. First, we utilize CLIP filtering metrics~\cite{sheynin2024emu, zhao2024ultraedit} to assess the alignment between edited images $I_e$ and their output captions $T_e$, ensuring consistency with editing instructions and desired modifications in target regions. Subsequently, we utilize CLIP image similarity to compare the original image $I_o$ and the edited image $I_e$, ensuring the fidelity of unedited image elements at the pixel level. Additionally, we compute the L1 distance between $I_o$ and $I_e$ to maintain the global structure of the original images. Finally, we employ CLIP directional similarity~\cite{zhang2024magicbrush}, which quantifies the alignment between modifications in images and the associated changes in captions, to verify the AnyEdit's instruction-following capability. We further apply the object detector to validate the presence of edited objects~(in Local Editing) and the VLM to assess the modification of global images~(in Global Editing).
\begin{table}[!t]
    \centering
    \resizebox{0.48\textwidth}{!}{
        \begin{tabular}{lcccc}
        \hline
        \multirow{2}{*}{Dataset}&\multicolumn{2}{c}{Semantic Similarity}  & \multicolumn{2}{c}{Visual Similarity} \\
        & CLIPin & CLIPout & SSIM & DINOv2 \\ \hline
        IP2P~\cite{brooks2023instructpix2pix}&0.2650&0.2694&0.5826&0.6859 \\
        UltraEdit~\cite{zhao2024ultraedit} & 0.2834 &0.3049&0.6401&0.7231\\
        AnyEdit~(Ours) & \textbf{0.3288} & \textbf{0.3133} & \textbf{0.6638}&\textbf{0.9053}\\
         
         \hline
        \end{tabular} 
    }\caption{Overall dataset quality comparison between AnyEdit and other datasets using automatic semantic and visual metrics.}
    \label{dataset-statistic}
    \vspace{-0.5cm}
\end{table}
\subsection{Characteristics and Statistics}\label{3.3}
Benefiting from our promising automated dataset collection approach, AnyEdit comprises 2.5 million high-quality editing pairs across 25 distinct editing types. AnyEdit encompasses a broader range of domains, including complex editing tasks such as viewpoint editing, implicit editing, and visual editing, and it incorporates a richer variety of scenes, including conceptually rich synthetic scenes~(c.f., Tab.~\ref{table1}). Moreover, the data distribution of AnyEdit in Fig.~\ref{example}(b)  reflects a broad variety of edit types, with coverage across domains. Quantitatively, we assess AnyEdit’s data quality based on semantic similarity and visual similarity metrics~(c.f., Tab.~\ref{dataset-statistic}). The significant improvement~(+25.2\% in DINOv2 and +16.0\% in CLIPin compared to the UltraEdit) shows AnyEdit's effectiveness in maintaining pixel-level consistency and accurately reflecting editing instructions.
% To further distinguish AnyEdit and other existing datasets, we elaborate on the dataset details in Table~\ref{table1}. Through our meticulous data curation process, we maintain a high-quality editing instruction dataset for comprehensive tasks and categories. The data distribution of AnyEdit is provided in Figure~\ref{example}(b), with its quantitative evaluation displayed in Table 3, demonstrating AnyEdit's significant advantages in diversity and data quality. 
\section{Method}
% In this section, we propose that AnySD cope with various tasks in AnyBench. Firstly, we introduce how to edit images by three conditions (original image, language prompt, visual prompt) in Section~\ref{sec4:1}. Then we further put forward the MoE Architecture of the AnySD in Section~\ref{sec4:2} and finally give details of its training strategy in Section~\ref{sec4:3}.

\subsection{Architecture}~\label{sec4:2}
\begin{figure}[!t]
  \vspace{-4mm}
    \centering
    \includegraphics[width=1.\linewidth]{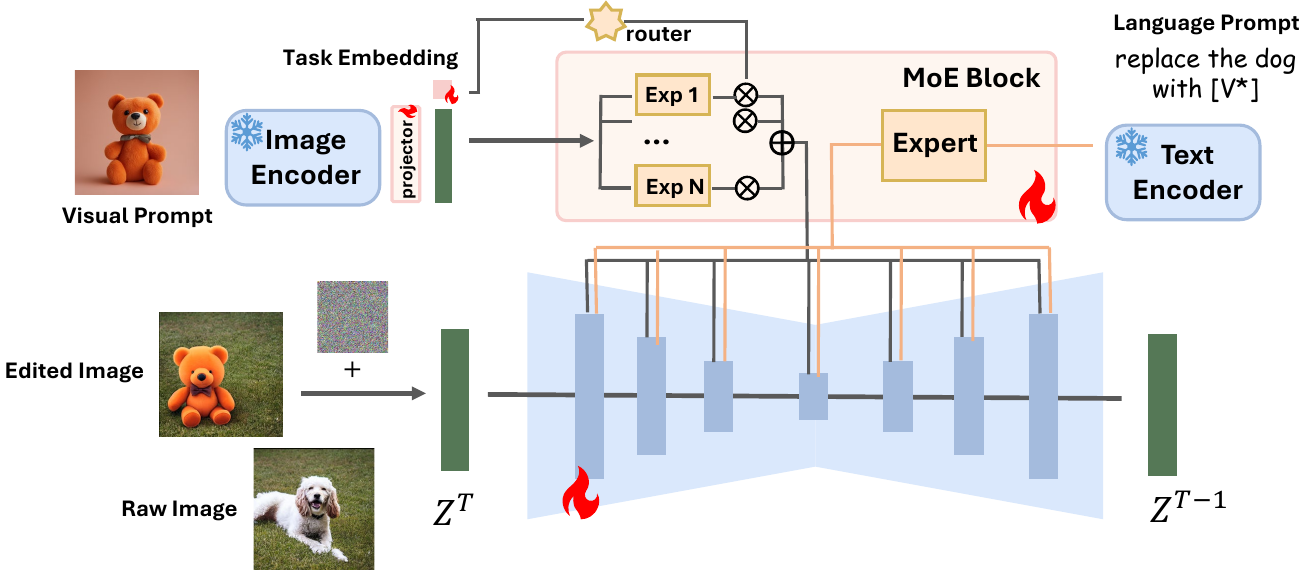}
    \caption{\textbf{Architecture of AnySD}. AnySD is a novel architecture to supports three conditions (original image, editing instruction, visual prompt) for various editing tasks. }\label{fig:model}
  \vspace{-4mm}
\end{figure}

Since AnyEdit contains a wide range of editing instructions across various domains, it holds promising potential for developing a powerful editing model to address high-quality editing tasks. However, training such a model has three extra challenges: (a) aligning the semantics of various multi-modal inputs; (b) identifying the semantic edits within each domain to control the granularity and scope of the edits; (c) coordinating the complexity of various editing tasks to prevent catastrophic forgetting. To this end, we propose a novel AnyEdit Stable Diffusion approach~(\textbf{AnySD}) to cope with various editing tasks in the real world. As illustrated in Figure~\ref{fig:model}, AnySD includes three designs: visual prompt projector, task-aware routing, and learnable task embedding. Then, we will introduce each AnySD design.
% Then, we will introduce each module of AnySD in detail.

\noindent\textbf{Visual Prompt Projector.} To align the semantics of multi-modal inputs, we first use a visual prompt projector to align these image features $z_V$ from the frozen CLIP image encoder with the instruction embeddings to obtain visual prompt $c_V$. We then integrate these embeddings into the timestep embeddings of the UNet through cross-attention interactions to provide visual prompt conditioning during general image editing diffusion~\cite{brooks2023instructpix2pix}. The diffusion objective is then adjusted as follows:
% process editing instructions for visual prompt conditioning, we first use a frozen CLIP image encoder to extract image features $z_V$. Then, we use a trainable MLP projector to align these features with the instruction embeddings, injecting them into the UNet through cross-attention interactions. The weights of the visual prompt projector are optimized jointly during the training process. Concretely, we integrate the aligned visual prompt $c_V$ into timestep embeddings for visual prompt conditioning during general image editing diffusion~\cite{brooks2023instructpix2pix}.
\vspace{-2mm}
\begin{equation}
L = \mathbb{E}_{\mathcal{E}(x), \mathcal{E}(c_I), c_T, c_V, \epsilon \sim \mathcal{N}(0, 1), t }\Big[ \Vert \epsilon - \epsilon_\theta(z_{t}, t, \mathcal{E}(c_I), c_T, c_V)) \Vert_{2}^{2}\Big]
\label{eq:loss}
\end{equation}
% \vspace{-1mm}
where $\epsilon \in N(0, 1)$ is the noise added by the diffusion process and $[x, c_I , c_T, (c_V)]$ is a quadruple of editing instruction, with $c_V$ only being applicable for visual editing.

\begin{figure*}[!t]
    % 使用 minipage 环境将图和表并排显示
    \hfill
    \begin{minipage}{0.75\textwidth}
\resizebox{1.0\textwidth}{!}{
    \begin{tabular}{l c c c c c c c c}
    \hline
    \multirow{2}{*}{} & \multicolumn{4}{c}{EMU-Edit Test} & \multicolumn{4}{c}{MagicBrush Test} \\
    \cline{2-9}
    & CLIPim $\uparrow$ & CLIPout $\uparrow$ & L1 $\downarrow$ & DINO $\uparrow$ & CLIPim $\uparrow$ & CLIPout $\uparrow$ & L1 $\downarrow$ & DINO $\uparrow$ \\
    \hline
        PnP~\citep{pnp}                                  & 0.521 & 0.089 & 0.089 & 0.304 & 0.568 & 0.101 & 0.289 & 0.220 \\
        Null-Text~\citep{mokady2023null}                 & 0.761 & 0.236 & 0.075 & 0.678 & 0.752 & 0.263 & 0.077 & 0.664 \\
        InstructPix2Pix~\citep{brooks2023instructpix2pix} & 0.834 & 0.219 & 0.121 & 0.762 & 0.837 & 0.245 & 0.093 & 0.767 \\
        EMU-Edit~\citep{sheynin2024emu}                  & 0.859 & 0.231 & 0.094 & 0.819 & 0.897 & 0.261 & 0.052 & 0.879 \\
        UltraEdit (SD3)~\citep{zhao2024ultraedit}                 & 0.845 & 0.283 & 0.071 & 0.794 & -     & -     & -     & -     \\
        SD1.5 w/ MagicBrush~\citep{zhang2024magicbrush}           & 0.838 & 0.222 & 0.100 & 0.776 & 0.883 & 0.261 & 0.058 & 0.871 \\\hline
        SD1.5 w/ AnyEdit (Ours) & 0.866~\footnotesize{(\textcolor{ForestGreen}{+3.3\%})} & 0.284~\footnotesize{(\textcolor{ForestGreen}{+27.9\%})} & 0.095~\footnotesize{(\textcolor{ForestGreen}{-5.0\%})} & 0.812~\footnotesize{(\textcolor{ForestGreen}{+4.6\%})} & 0.892~\footnotesize{(\textcolor{ForestGreen}{+1.0\%})} & 0.273~\footnotesize{(\textcolor{ForestGreen}{+4.6\%})} & 0.057~\footnotesize{(\textcolor{ForestGreen}{-1.7\%})} & 0.877 ~\footnotesize{(\textcolor{ForestGreen}{+0.7\%})} \\
        \textbf{AnySD w/ AnyEdit (Ours)}                             & \textbf{0.872}~\footnotesize{(\textcolor{ForestGreen}{+4.1\%})}     & \textbf{0.285}~\footnotesize{(\textcolor{ForestGreen}{+28.4\%})}     & \textbf{0.070}~\footnotesize{(\textcolor{ForestGreen}{-30.0\%})}     & \textbf{0.821}~\footnotesize{(\textcolor{ForestGreen}{+5.8\%})}     & \textbf{0.898}~\footnotesize{(\textcolor{ForestGreen}{+1.7\%})}     & \textbf{0.275}~\footnotesize{(\textcolor{ForestGreen}{+5.4\%})}     & \textbf{0.051}~\footnotesize{(\textcolor{ForestGreen}{-12.1\%})}     & \textbf{0.881}~\footnotesize{(\textcolor{ForestGreen}{+1.1\%})}     \\
    \hline
    \end{tabular}
}
\captionof{table}{Comparison of methods on EMU-Edit~\citep{sheynin2024emu} and MagicBrush~\citep{zhang2024magicbrush} benchmark. We show performance improvements over SOTA models of the same architecture, with only training data differences.}\label{table:1}
    \end{minipage}
     \begin{minipage}{0.24\textwidth}
        \centering
        \vspace{-4mm}
        \includegraphics[width=\textwidth]
        {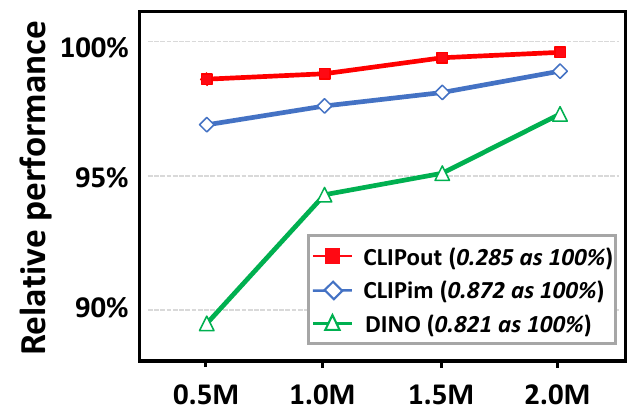}  % 替换为你的图片文件
        \vspace{-7mm}
        \caption{Dataset scales effect.}
        \label{fig:dataset_scale}
    \end{minipage}
    \vspace{-0.3cm}
\end{figure*}

\noindent\textbf{Task-aware Routing.} Since each editing type exhibits different granularities and scopes of edits, we propose a task-aware routing strategy with Mixture of Experts~(MoE)~\cite{masoudnia2014mixture} blocks to meet task-specific editing requirements.
% Since each editing type requires the model to selectively focus on specific elements, such as ensuring visual consistency in visual editing or adjusting style while preserving the original image structure in style transfer. Therefore, we propose a task-aware routing strategy with Mixture of Experts~(MoE)~\cite{masoudnia2014mixture} blocks to control each editing type's granularity and specific elements. 
Specifically, the visual condition $c_V$ is integrated into the frozen UNet layers with decoupled cross-attention to avoid disrupting the edit instruction condition. Each MoE block shares the same text attention layer but has diverse visual attention weights for $c_V $ by the router based on the task embedding. Therefore, we introduce a cross-attention layer for each expert and add the distributed results from all experts to the output of text cross-attention as follows,
\begin{equation}
\begin{aligned}
\mathbf{Z}^{new} &= \text{Softmax}\left(\frac{\mathbf{Q}\mathbf{K}^{\top}}{\sqrt{d}}\right)\mathbf{V} 
+ \text{Softmax}\left(\frac{\mathbf{Q}(\mathbf{K}')^{\top}}{\sqrt{d}}\right)\mathbf{V}' 
\end{aligned}
\end{equation}
where $\mathbf{K}'=\boldsymbol{c}_{v}\mathbf{W}'_k$ is the query matrix from the visual prompt and $\mathbf{W}'_k$ is the corresponding weight matrix.

\noindent\textbf{Learnable Task Embeddings.}
To guide the generation process toward the appropriate granularity for each editing task, we learn a task embedding $v_i$ during training. Unlike previous work~\cite{sheynin2024emu}, we insert task embeddings before the MoE block rather than integrate them into the whole denoising process to alleviate query confusion between instructions and editing types. The task embedding $v_i$ has two main roles: (1) It concatenates with $z_v$ (set to zero when no visual prompt is provided) to form $c_V$ for MoE blocks; (2) it inputs to the router \citep{zhou2022mixture}, distributing weights across MoE experts. More details of architecture are in Appendix \textcolor{red}{G.1}

\subsection{Training and Inference}~\label{sec4:3}
To enhance AnySD's ability to handle diverse conditioning for editing, we introduce CFG~\cite{ho2022classifier} extending from InstructPix2Pix~\cite{brooks2023instructpix2pix} for three conditioning. Furthermore, we structure our training into two stages for AnySD, ensuring that the diffusion model can thoroughly understand general editing knowledge and develop fine-grained task-specific skills.
% structure our training into two phases. This approach ensures that the model can thoroughly comprehend general instruction capabilities while developing fine-grained skills for various tasks. 

% training details can be seen in Appendix~\ref{app:train_details}
\textbf{Stage I: Instruction Understanding}. In this phase, we freeze the task-aware router and only pre-train the UNet layer to align with editing instructions. Additionally, We set additional conditions to zero tensors by CFG to enhance the model’s instruction-following capability.
% Additionally, we randomly omit original images and editing instructions to enhance CFG process.

\textbf{Stage II: Task Tuning}. In this phase, we further fine-tune $\mathbf{W}'_k$ and $\mathbf{W}'_v$ within the MoE block, task-aware router, visual prompt projector, and task embeddings to adapt the model to the task-specific editing granularity.
% In cases where the image condition is omitted, we simply set the CLIP image embedding to zero.

\textbf{Inference}. 
% For a fair comparison, we adopt the diffusion editing model from InstructPix2Pix~\citep{brooks2023instructpix2pix}, using Stable Diffusion 1.5~\citep{sd1.5} as the backbone. We infer in $512\times 512$.
In the inference stage, we predict the editing type with LLMs~(\textit{e.g.}, LLaMA-3) when given the input instruction. Then, we apply our AnySD for editing.
\begin{table}[!t]
\centering
\resizebox{0.49\textwidth}{!}{
    \begin{tabular}{l l c c c c }
    \hline
    % \cline{2-9}
    && CLIPim $\uparrow$ & CLIPout $\uparrow$ & L1 $\downarrow$ & DINO $\uparrow$\\
    \hline
        \multirow{7}{*}{\rotatebox[]{90}{\small{Local}}}         &Null-Text~\citep{mokady2023null} &0.773   &0.270   &0.245    &0.641    \\
        &InstructPix2Pix~\citep{brooks2023instructpix2pix} &0.753   &0.274   &0.164    &0.562    \\
        &MagicBrush~\citep{zhang2024magicbrush}   &0.823   &0.293   &0.120    &0.698    \\
        &HIVE$^w$~\citep{zhang2023hive} &0.814   &0.294   &0.184    &0.651    \\
        &HIVE$^c$~\citep{zhang2023hive} &\underline{0.845}   &\underline{0.299}   &0.114    &\underline{0.766}    \\
        &UltraEdit~(SD3)~\citep{zhao2024ultraedit}  &0.831   &\textbf{0.308}   &\underline{0.112}    &0.731    \\
        &\textbf{AnyEdit (Ours)}  &\textbf{0.863}   &0.297   &\textbf{0.094}    &\textbf{0.788}    \\
    \hline
        \multirow{7}{*}{\rotatebox[]{90}{\small{Global}}} &Null-Text~\citep{mokady2023null} &0.753   &0.277   &0.270    &0.613    \\
        &InstructPix2Pix~\citep{brooks2023instructpix2pix} &0.747   &0.261   &\underline{0.180}    &0.523    \\
        &MagicBrush~\citep{zhang2024magicbrush}   &0.731   &0.278   &0.233    &0.493    \\
        &HIVE$^w$~\citep{zhang2023hive} &0.762   &0.287   &0.196    &0.579    \\
        &HIVE$^c$~\citep{zhang2023hive} &\underline{0.787}   &0.294   &0.253    &0.576    \\
        &UltraEdit (SD3)~\citep{zhao2024ultraedit}  &0.772   &\underline{0.297}   &0.191    &\underline{0.619}    \\
        
        &\textbf{AnyEdit (Ours)}  &\textbf{0.788}   &\textbf{0.301}   &\textbf{0.159}    &\textbf{0.647}   \\
    \hline
        \multirow{6}{*}{\rotatebox[]{90}{\small{Camera Movement}}} &
        InstructPix2Pix~\citep{brooks2023instructpix2pix} &0.700   &-   &0.178    &0.477    \\
        &MagicBrush~\citep{zhang2024magicbrush}   &0.765   &-   &0.170    &0.589    \\
        &HIVE$^w$~\citep{zhang2023hive} &0.779   &-   &0.175    &0.619    \\
        &HIVE$^c$~\citep{zhang2023hive} &\textbf{0.838}   &-   &\underline{0.171}    &\underline{0.739}    \\
        &UltraEdit (SD3)~\citep{zhao2024ultraedit}  &0.802   &-   &0.340    &0.514    \\
        &\textbf{AnyEdit (Ours)}  &\underline{0.833}   &-   &\textbf{0.110}    &\textbf{0.745}    \\
    \hline
        \multirow{6}{*}{\rotatebox[]{90}{\small{Implicit}}} &
        InstructPix2Pix~\citep{brooks2023instructpix2pix} &0.794   &\underline{0.288}   &0.190    &0.558    \\
        &MagicBrush~\citep{zhang2024magicbrush}   &\underline{0.865}   &0.280   &0.149    &0.711    \\
        &HIVE$^w$~\citep{zhang2023hive} &0.821   &0.284   &0.161    &0.635    \\
        &HIVE$^c$~\citep{zhang2023hive} &0.862   &0.284   &0.137    &\underline{0.728}    \\
        &UltraEdit (SD3)~\citep{zhao2024ultraedit}  &0.856   &0.281   &\underline{0.135}    &0.703    \\
        &\textbf{AnyEdit (Ours)}  &\textbf{0.867}   &\textbf{0.289}   &\textbf{0.130}    &\textbf{0.733 }   \\
    \hline
       \multirow{2}{*}{\rotatebox[]{90}{\small{Visual}}} &
        Uni-controlnet~\citep{zhao2023uni} &0.717   &0.249   &0.260    &0.442    \\
        &\textbf{AnyEdit (Ours)}  &\textbf{0.801}      &\textbf{0.258}    &\textbf{0.145}  &\textbf{0.628}   \\
    \hline
    \end{tabular}
}
\caption{Comparison of methods on AnyEdit-Test benchmark.}\label{table:2}
\vspace{-0.4cm}
\end{table}
\section{Experiments}
% Our experiments evaluate the quality of AnyEdit in accurately following user instructions while maintaining the visual fidelity of source images
% We first evaluate AnyEdit on the wide-ranged editing benchmarks~(\S~\ref{5.2}). 
% Additionally, we examine its promising expandability to more challenging editing tasks with our AnyEdit-Test benchmark

In this section, we first assess AnyEdit and AnySD on popular standard editing benchmarks~(\S~\ref{5.2}), demonstrating the high quality of the AnyEdit dataset and the superiority of the AnySD architecture. Additionally, we extend the evaluation to the more challenging AnyEdit-Test benchmark~(\S~\ref{5.3}) to show the promising expandability of our approach, better aligning with the creative editing demands in real-world scenarios. We further present qualitative results (\S~\ref{5.4}) and conduct in-depth analysis (\S~\ref{5.5}) to illustrate the scalability and broader applicability of AnyEdit.

\begin{figure*}[!t]
\vspace{-0.3cm}
    \centering
        \includegraphics[width=1.\linewidth]{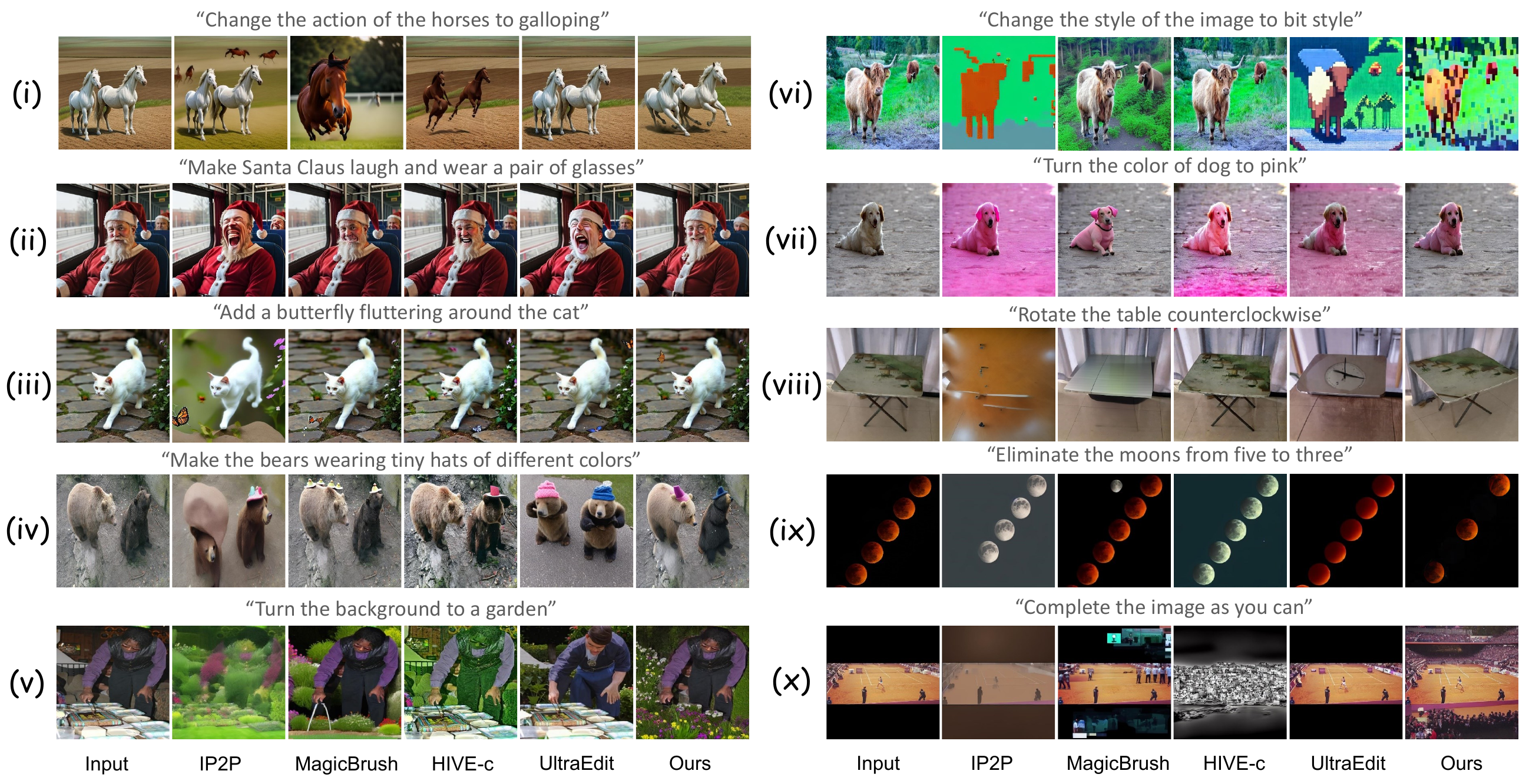}
        \vspace{-5mm}
    \caption{Qualitative evaluation of 10 distinct and complex tasks on AnyEdit-Test, such as \textit{action change}~(i), \textit{appearance alter}~(iv), \textit{rotation change}~(viii), \textit{counting}~(ix), and \textit{outpainting}~(x), demonstrates that our method yields promising results across these editing tasks.}\label{fig:main_case_part1}
    \vspace{-0.4cm}
\end{figure*}

\subsection{Experimental Setup}\label{5.1}
\noindent\textbf{Settings.} For a fair comparison, we adopt Stable-Diffusion 1.5~\cite{rombach2022high} as the backbone and follow the settings of InstructPix2Pix~\cite{brooks2023instructpix2pix} to train our AnySD. Notably, we only use data from AnyEdit for training, without incorporating any additional datasets. More details are in Appendix \textcolor{red}{G.5}.

\noindent\textbf{Benchmarks \& Metrics.} We access our method across two popular benchmarks: Emu Edit Test~\cite{sheynin2024emu} and MagicBrush~\cite{zhang2024magicbrush}. These standard benchmarks evaluate editing models by comparing edited results with ground truths. Additionally, we manually selected 50 high-quality editing data from each editing type in AnyEdit, creating AnyEdit-Test for a more challenging and comprehensive evaluation. Notably, AnyEdit-Test is not visible during training.
See Appendix \textcolor{red}{E} for details of AnyEdit-Test. Following prior work~\citep{zhang2024magicbrush, sheynin2024emu, fu2023guiding}, we adopt semantic similarity~(\textit{i.e.}, CLIPim and CLIPout) and visual similarity~(\textit{i.e.}, DINO and L1 distance) metrics to evaluate the effectiveness of AnySD trained on AnyEdit for instruction-based image editing.

\noindent\textbf{Baselines.} We use the following baselines: 1) specialized image editing methods: PnP~\cite{pnp}, Null-Text~\cite{mokady2023null}. 2) instruction-based methods: it directly edits images with natural language, including InstructPix2Pix~\cite{brooks2023instructpix2pix}, MagicBrush~\cite{zhang2024magicbrush}, 
HIVE~\cite{zhang2023hive},
EMU-Edit~\cite{sheynin2024emu}, UltraEdit~\cite{zhao2024ultraedit}. 3) visual condition methods: it targets visual editing, including Uni-controlnet~\cite{zhao2023uni}. More details are in Appendix \textcolor{red}{G.6}.

\subsection{Main Results on Standard Image Editing}\label{5.2}

We report the standard image editing results of AnyEdit and other baselines on EMU-Edit Test and MagicBrush benchmarks in Table~\ref{table:1}. Based on the experimental results, we have summarized the following conclusions: \textbf{\textit{(i)}} 
% Our method consistently demonstrates superior semantic performance compared to baseline models, achieving 0.872 for CLIPim and 0.285 for CLIPout on the EMU-Edit Test. 
Our SD-1.5 with AnyEdit, which only changes the training data to AnyEdit, consistently demonstrates superior semantic performance in both edit alignment and content preservation compared to SOTA methods, even without additional mask supervision~(0.872 for CLIPim and 0.285 for CLIPout on the EMU-Edit Test). It highlights AnyEdit's effectiveness in mastering high-quality image editing, validating its \textbf{high-quality editing data with significant semantic alignment and underlying clear editing structure}.
% which is largely due to it comprising well-aligned editing data provided by AnyEdit. 
\textbf{\textit{(ii)}} Our AnySD model, trained on AnyEdit using the AnySD architecture, further surpasses SOTA methods in both semantic and visual similarity~(0.872 of CLIPim on EMU-Edit Test and 0.881 of DINO on MagicBrush Test), setting new records on MagicBrush and Emu-Edit benchmarks. This demonstrates \textbf{the superiority of AnySD in following editing instructions while preserving unchanged image elements}, thanks to its task-aware architecture that learns task-specific knowledge from the diverse editing types in AnyEdit, enhancing the model's cross-task editing capabilities.

\subsection{Comparison on AnyEdit-Test Benchmark}\label{5.3}
Table~\ref{table:2} presents the results of the AnyEdit-Test benchmark, where each instruction is designed to rigorously evaluate AnyEdit’s adaptability across a wider range of challenging editing scenarios. We provide further results of each editing category in Appendix \textcolor{red}{F}. It can be observed that \textbf{\textit{(i)}} 
most baselines struggle to effectively handle more complex editing tasks that are rarely in standard benchmarks~(0.190 v.s. 0.121 on average L1), especially for implicit editing that requires reasoning abilities. This illustrates \textbf{the importance of AnyEdit-Test for evaluating the performance of editing models on complex tasks}. \textbf{\textit{(ii)}} Even for common editing tasks, state-of-the-art models show a significant decline in consistency performance on AnyEdit-Test~(-3.5\% on CLIPim and -19.2\% on DINO of UltraEdit). This underscores \textbf{the limitations of existing benchmarks in evaluating multi-scene editing}. \textbf{\textit{(iii)}} In contrast, AnyEdit significantly outperforms SOTA methods across all editing categories, demonstrating its scalability and robustness in handling complex tasks across diverse scenarios. \textbf{\textit{(iv)}} Traditional methods often struggle to handle visual editing effectively due to additional visual inputs. In such cases, even when compared to Uni-ControlNet, which is pre-trained with diverse visual conditions, AnyEdit consistently performs better in visual editing tasks. It shows the efficacy of AnyEdit in handling vision-conditioned editing instructions. We provide more qualitative evidence in Sec.~\ref{5.4}.

\begin{figure}[!t]
\vspace{-0.5cm}
    \centering
        \includegraphics[width=1.\linewidth]{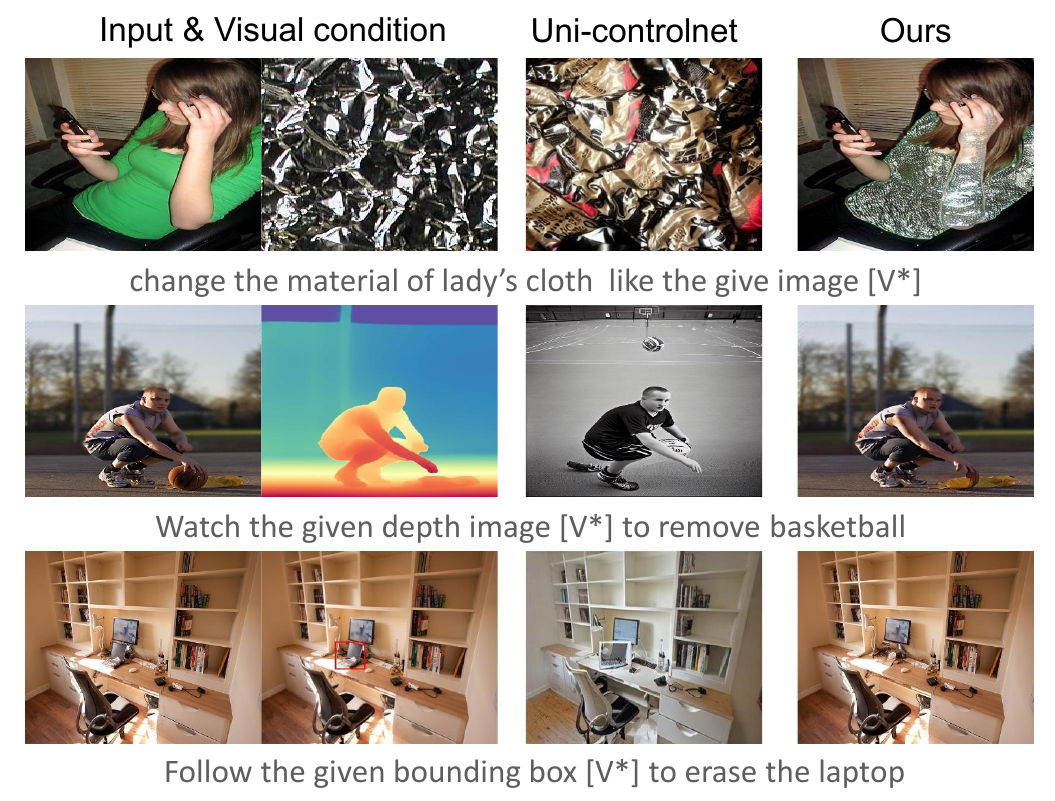}
        \vspace{-0.5cm}
    \caption{Comparison of our image editing method against Uni-controlnet~\cite{zhao2023uni} for visual editing tasks on AnyEdit-Test.}\label{fig:main_case_part2}
    \vspace{-0.6cm}
\end{figure}
\subsection{Qualitative Evaluation}\label{5.4}
Due to the limitations of quantitative metrics in evaluating editing tasks, we perform qualitative evaluations to further assess the effectiveness of our approach, as shown in Figure~\ref{fig:main_case_part1}. Our key observations are: 1) Most baseline models, including the human-tuned HIVE-c\cite{zhang2023hive} and the SOTA UltraEdit with its extensive training data\cite{zhao2024ultraedit}, \textbf{still suffer from over-editing or misalignment when handling complex fine-grained instructions} (e.g., ``facial distortion" and ``missing glasses" in Fig.~\ref{fig:main_case_part1}(ii)). 2) Due to the limited diversity and quality of current datasets, previous methods ~(\textit{i.e.}, ip2p~\cite{brooks2023instructpix2pix}, MagicBrush~\cite{zhang2024magicbrush}, and UltraEdit) \textbf{struggle to generalize to novel editing types in diverse scenarios}~(\textit{e.g.}, failing to follow instructions in rotation change and counting tasks, or roughly altering objects instead of its fine-grained appearance in appearance alter task). 3) In contrast, our method can effectively \textbf{ensure editing accuracy in target regions and maintain consistency in irrelevant areas} even without any mask guidance~(Fig.~\ref{fig:main_case_part1}(i), (vii)). Also, our method can \textbf{automatically distinguish between foreground and background} to modify the background~(Fig.~\ref{fig:main_case_part1}(v)). Moreover, our method \textbf{successfully executes more complex editing instructions}~(\textit{e.g.}, style change in Fig.~\ref{fig:main_case_part1}(vi) and outpainting in Fig.~\ref{fig:main_case_part1}(x)).

Additionally, we visualize AnyEdit results on visual editing in Figure~\ref{fig:main_case_part2}. In this challenging setting, UniControl can either reflect only the pixel information from the visual condition or retain the semantics of the original image without performing any edits. In contrast, for various visual instructions, AnyEdit \textbf{consistently comprehends the pixel information in visual conditions and achieves reliable editing}. These promising visualization results confirm the effectiveness and high quality of the diverse editing data in AnyEdit. More qualitative results are shown in Appendix \textcolor{red}{H.2}.

\begin{table}[!t]
\vspace{-0.3cm}
\centering
\resizebox{0.46\textwidth}{!}{
    \begin{tabular}{ll c c c c }
    \hline
    % \cline{2-9}
    && CLIPim $\uparrow$ & CLIPout $\uparrow$ & L1 $\downarrow$ & DINO $\uparrow$\\
    \hline 
    &AnySD w/ AnyEdit & \textbf{0.872}   &\textbf{0.285}  &\textbf{0.070} &\textbf{0.821} \\
    \hline
       1 &w/o task-aware routing &0.838  &0.275  &0.154    &0.757    \\
       2 &w/o task emb &0.859  &0.282  &0.107 &0.809  \\ \hline
       3 & AnyEdit (w/o compsn.)  &0.868  &0.271  &0.099    &  0.785\\
    \hline
    \end{tabular}
}
\caption{Ablation study of our method on EMU-Edit Test~\citep{sheynin2024emu}.}\label{table:ablation}
\vspace{-0.6cm}
\end{table}
\subsection{In-depth Analysis}\label{5.5}
\textbf{AnySD Architecture}.
We investigate the effectiveness of each component and conduct the following experiments on EMU-Edit Test benchmark: 
(1) We remove the text-aware routing strategy in AnySD~(c.f. Rows 1 of Tab.~\ref{table:ablation}) and find that it leads to significant performance degradation~(0.838 v.s. 0.872 in CLIPim and 0.154 v.s. 0.070 in L1), demonstrating its crucial role for adapting diverse image editing tasks. (2) We remove task embeddings in AnySD and observe that it has little impact on semantic alignment but significantly affects visual consistency~(c.f. Rows 2 of Tab.~\ref{table:ablation}), suggesting that task embeddings control the perceptual granularity of pixel information during cross-attention.

\noindent\textbf{Analysis of Data Scaling in AnyEdit}.
In Figure~\ref{fig:dataset_scale}, we analyze the data scaling effect of AnyEdit on image editing capability. We observe: (1) For consistency metrics (\textit{i.e.}, CLIPim and DINO), performance improves progressively as data scale increases; (2) For editing accuracy metrics (\textit{i.e.}, CLIPout), we can achieve promising performance even with a small amount of data, indicating that AnyEdit excels in semantic alignment. (3) Furthermore, we remove the AnyEdit-Composition editing data in counterfactual synthesis scenarios, as shown in row 3 of Tab.~\ref{table:ablation}, the lack of conceptual balance in AnySD hinders its ability to generalize, leading to a decline in semantic performance~(-4.9\% of CLIPout). This confirms the efficacy of counterfactual synthetic scenes to generalization in editing tasks.

\section{Conclusion}
In this work, we present a novel perspective for categorizing editing tasks and introduce a unified framework that exploits adaptive pipelines to construct high-quality data for diverse editing tasks in a low-resource manner. Building on this, we propose AnyEdit, a multi-type, multi-scene instruction-based editing dataset comprising 2.5M editing samples across 25 distinct types, along with its benchmark, AnyEdit-Test, enabling a more comprehensive paradigm for unified image editing. Furthermore, we develop the powerful AnySD, unlocking the full potential of AnyEdit. Extensive experiments on standard benchmarks and the challenging AnyEdit-Test demonstrate that our method excels in high-quality image editing across diverse tasks and scenarios, accurately executing complex instructions while preserving image consistency in unmodified elements.

\noindent\textbf{Acknowledgment.} This work was supported by the National Natural Science Foundation of China (62436007), the Key R\&D Projects in Zhejiang Province (No. 2024C01106, 2025C01030), and the Zhejiang NSF (LRG25F020001). We thank all the reviewers for their valuable comments.

% , Key Research and Development Projects in Zhejiang Province (No. 2024C01106), the National Key Research and Development Project of China (2018AAA0101900), and Research funding from FinVolution Group. We thank all the reviewers for their valuable comments.

{
    \small
    \bibliographystyle{ieeenat_fullname}
    \bibliography{main}
}
\appendix
%%%%%%%%%%%%% 目录
% \addtocontents{toc}{\protect\setcounter{tocdepth}{3}}
% \hypersetup{linkcolor=black}
% % \tableofcontents
% {\small \tableofcontents} % Use \small or other size commands
% \hypersetup{linkcolor=red}
% \newpage
%%%%%%%%%%%%% 
\clearpage
\maketitlesupplementary
\section{Overview}
In this supplementary material, we present:

\begin{itemize}
	\item More detailed dataset collection process of AnyEdit~(Section \ref{2}).
	
	\item Statistic information of AnyEdit~(Section \ref{3}).
	
	\item Additional examples of AnyEdit~(Section \ref{4}).

         \item Detailed description of AnyEdit-Test Benchmark~(Section \ref{5}).
         \item Detailed experimental results of various editing types on AnyEdit-Test Benchmark~(Section \ref{6}).
         \item Implementation details~(Section \ref{7}).
         \item More qualitative results on various benchmarks and human evaluations~(Section \ref{8}).	
\end{itemize}

\begin{figure*}[t]
    \centering
    \includegraphics[width=1.\linewidth]{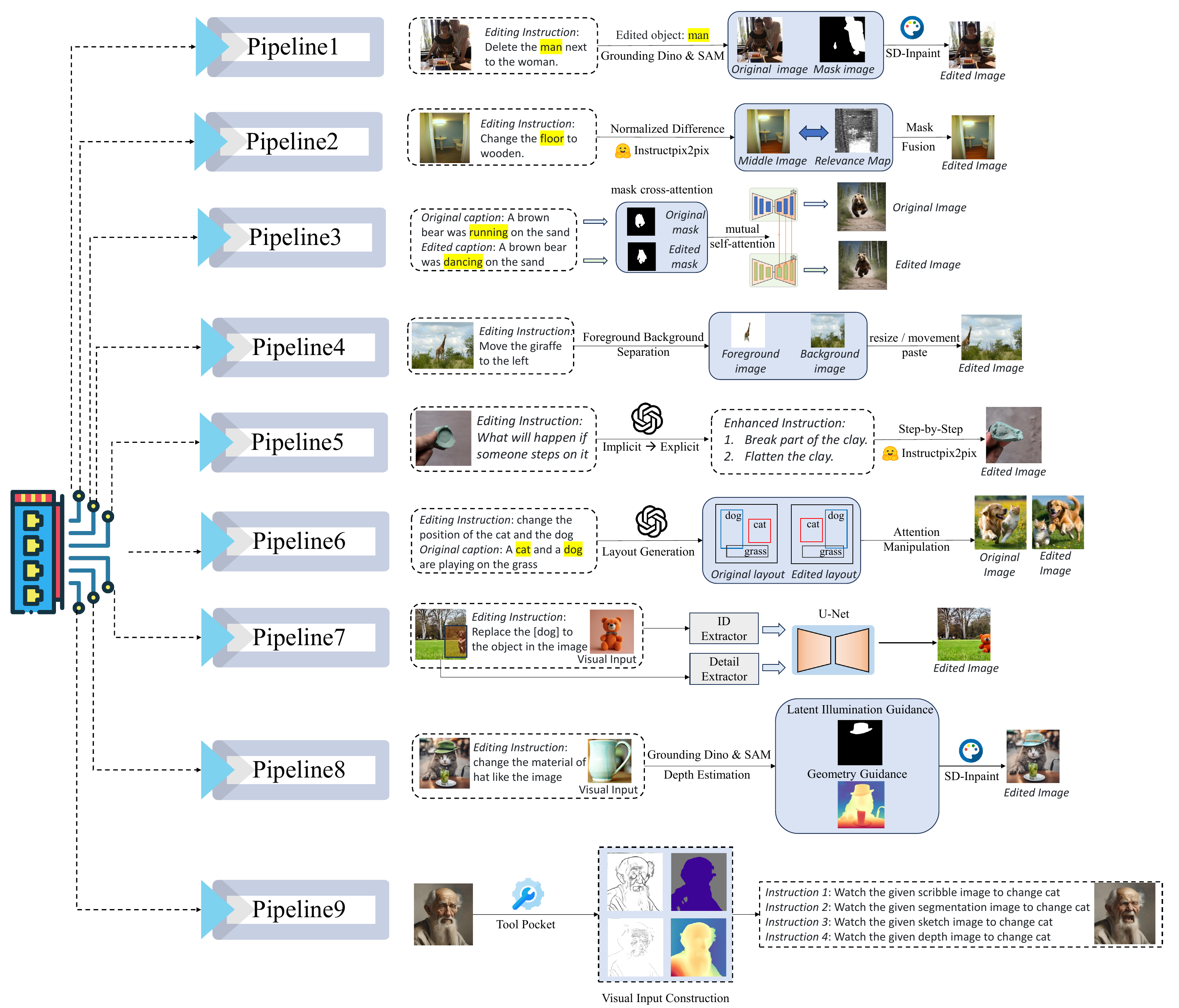}
    \vspace{-4mm}
    \caption{The illustration of detailed pipelines for each editing type in the AnyEdit dataset collection.} 
    \vspace{-4mm}
    \label{total_pipelines}
\end{figure*}
\section{Detailed Dataset Collection Process}\label{2}
\subsection{Editing Type Definition} Here, we explain the detailed definition of each editing task in AnyEdit, which comprises five primary categories with 25 distinct editing types, as shown in Table~\ref{dataset_info}.
\subsection{Diverse Instruction Generation.}
\subsubsection{Prompt Constraints}\label{prompt}
To address the limitations in instruction diversity and consistency during the process of Instruction Generation, we use prompt constraints to guide the LLM as a task-specific agent that responds in JSON format with diverse editing types. A key innovation is incorporating a task-specific user-LLM conversational history into the prompt, where we replace direct constraints with high-quality, hand-crafted examples. This approach enables the LLM to learn from ideal responses and improve subsequent generations. Specifically, we design templates for each task instructing the LLM to respond in the required format. These templates are tailored for each task and include four core elements: \textit{input}, \textit{output}, \textit{editing type}, and \textit{edit instruction}. We also define a set of action verbs for each task, ensuring the LLM’s response aligns with our guidelines and improving the quality and consistency of the generated output. The detailed prompt constraints are shown in Table~\ref{prompt_template}.
\subsubsection{In-context Examples}
As mentioned in Section 3.2, we employ in-context examples in the conversation history to develop a task-specific agent tailored to each editing type. For each new task, we initially generate a set of five in-context example instructions into prompt templates for instruction generation. After generating an editing instruction, we integrate it with its original caption to create instruction pairs, which are used to expand the in-context example pool. For each subsequent generation process, five in-context examples are randomly selected. This iterative self-enhancement mechanism exposes the generation process to a diverse range of examples, encouraging more varied and robust output responses. In this way, we maintain a cohesive conversational flow while progressively increasing the diversity and complexity of the generated instructions, facilitating the construction of the AnyEdit dataset.

\subsection{Adaptive Editing Pipelines}
This section will elaborate on the pipeline implementation details for various editing tasks in the AnyEdit datasets collection.
Figure~\ref{total_pipelines} shows the illustrations of the main specific pipelines in our Adaptive Editing Pipeline module to construct these various high-quality editing instructions adaptively.
\subsubsection{Local Editing}
\noindent\textbf{Remove.} As shown in \textcolor{red}{Pipeline1} of Figure~\ref{total_pipelines}, we first extract the mask of the edited object in the editing instruction by GroundingDINO~\cite{liu2023grounding} and Segment Anything~\cite{kirillov2023segment}. We then generate the target image using SD-Inpaint~\cite{rombach2022high}, with the original image and the mask produced during the process. To remove the target object, we set the prompt word to empty and the negative prompt word to edited object. Notably, we also apply dilation to the mask and perform Gaussian filtering to smooth it, ensuring a more natural blend with the surrounding contents. In addition, we merge the edited image, mask, and original image to retain image elements outside the content of the editing instructions.

\noindent\textbf{Replace.} As shown in \textcolor{red}{Pipeline1} of Figure~\ref{total_pipelines}, the process of generating data for replace type is similar to the removal. The only difference is
that we set the prompt word to the new object. In this way, our pipeline tends to produce a new object to replace the edited object instead of removing it.

\noindent\textbf{Add.} As shown in \textcolor{red}{Pipeline1} of Figure~\ref{total_pipelines}, the process of generating data for add type is similar to the removal. 
However, since the add edit is to add a new object to the original image, its correct placement is unknown. Thus, we reverse the process by first generating the image of the output caption as the edited image and then using the `remove' editing instruction to obtain the original image that does not include the newly added object. In this way, we obtain the original image and the edited image with the newly added object seamlessly integrated.

\noindent\textbf{Counting.} 
The counting-type editing introduces the concept of object quantity, performing the corresponding number of removal or additions iteratively based on the specified count in the instructions. Each step in the process is illustrated in the \textcolor{red}{Pipeline1} of Figure~\ref{total_pipelines}.

\noindent\textbf{Color Alter \& Appearance Alter.} As shown in \textcolor{red}{Pipeline2} of Figure~\ref{total_pipelines}, the key aspect of editing types like color alteration and appearance alteration lies in modifying only the object's attributes or appearance instead of altering or removing whole objects. Therefore, we introduce a Normalized Attention Difference~\cite{mirzaei2025watch} based on input-output caption discrepancies to identify the target editing mask. Based on this, we apply InstructPix2Pix~\cite{brooks2023instructpix2pix} for instruction-based editing, blending the original and edited images within the masked region to produce the final result, thereby minimizing element confusion.

\noindent\textbf{Action Change.} To achieve complex non-rigid image editing, we introduce a joint intervention mechanism involving mutual self-attention and masked cross-attention, as shown in \textcolor{red}{Pipeline3} of Figure~\ref{total_pipelines}. This approach addresses the limitations in the action change editing instructions, which sometimes fail to accurately execute editing intentions due to the need for fine-grained modifications.

\noindent\textbf{Textual Change.} To meet the demands for textual change, we collected captions containing text from the AnyWord-3M dataset~\citep{tuo2023anytext}, namely, ArT~\citep{chng2019icdar2019}, COCO-Text~\citep{COCO-Text}, RCTW~\citep{RCTW}, LSVT~\citep{LSVT}, MLT~\citep{MLT}, MTWI~\citep{MTWI}, ReCTS~\citep{ReCTS}. Following this, we generate editing instructions that alter only the text within the image, guided by specific type constraints and in-context examples in our diverse instruction generation. We ultimately generate corresponding images as the final result by using a text-specialized T2I model~(\textit{i.e.}, FLUX), with both the original caption and the edited caption maintained under the same seed.

\noindent\textbf{Material Change.} We reuse the original and edited images from Material Transfer in Visual Editing. However, we only utilize editing instructions to convey the editing intent without using material images as references. Specifically, we will change ``the material of [object] like the image" to ``change the material of [object] to [material category]".
\subsubsection{Global Editing}
\noindent\textbf{Background Change.} As shown in \textcolor{red}{Pipeline1} of Figure~\ref{total_pipelines}, we define background changes as modifications to the edited object ``background". 
To avoid unnecessary foreground modifications, we extract and invert all foreground masks from captions, then merge them with the background mask. Similar to the replacement instructions, we also apply dilation to the merged mask and perform Gaussian filtering to eliminate artifacts in the contour.

\noindent\textbf{Tone Transfer.} We define three types of changing scenes~(\textit{i.e.}, season, time, weather) involving the overall tone of the image. According to this, we generated editing instructions tailored to tone transfer and used InstructPix2Pix~\cite{brooks2023instructpix2pix} to edit the whole image, as shown in \textcolor{red}{Pipeline1} of Figure~\ref{total_pipelines}.

\noindent\textbf{Style Change.} We collect 50 desired style images and extract 2,500 images from the MSCOCO validation set as original images. Using an API of Prisma Art, we applied style transfer to obtain the edited results. Ultimately, we only retain the intuitive style as the style changes editing instructions, such as ``animated".
\subsubsection{Camera Movement Editing}
\noindent\textbf{Movement \& Resize.} As shown in \textcolor{red}{Pipeline4} of Figure~\ref{total_pipelines}, we first extract the foreground object and backgrounds separately according to the edited object. Here we use the ``remove" operation to ensure the pixel integrity of the background after removing the foreground. Then, we utilize the crop-and-paste operation to change the size of the edited object and the position of it in the edited image.

\noindent\textbf{Outpainting.}
To reduce the complexity of constructing data, we inversely designate the images from the initial dataset as extended images. Given the input caption, we randomly select an object within it and use GroundingDINO to extract its bounding box in the image. 
Then, we apply a mask to the areas outside the bounding box and obtain the original image that contains only selected elements. The extended and original images are then used to construct editing instructions for the outpainting type. 

\noindent\textbf{Rotation Change.} 
Since direct perspective rotation change of images is challenging, we extract related image pairs directly from MVImgNet~\cite{yu2023mvimgnet}, the Large-scale Dataset of Multi-view Images, to construct original images and edited images for rotation changes.
Then, we categorize the editing instructions as ``rotate the object clockwise" and ``rotate the object counterclockwise" according to changes in the camera's viewpoint, thereby constructing corresponding pairs of editing data.
\subsubsection{Implicit Editing}
\noindent\textbf{Implicit Change.} As shown in \textcolor{red}{Pipeline5} of Figure~\ref{total_pipelines}, we first elicit the world knowledge from LLMs to transform implicit instructions into explicit instructions, which directly convey executable editing intentions~(\textit{e.g.}, ``Flatten the clay" directly conveys the alteration in the clay's appearance without requiring additional interpretation). In this way, we use the instruction-based editing method to complete the explicit instructions step-by-step, thereby constructing edited images with implicit changes. We also enrich the dataset by using existing dynamic world editing datasets~\cite{yang2024editworld}.

\noindent\textbf{Relation Change.} To adjust the positional relationships of objects within images, our pipeline first generates layouts based on the original captions, as shown in \textcolor{red}{Pipeline6} of Figure~\ref{total_pipelines}. We can swap the positional relationship between two objects in the layout space to construct the edited layout. Subsequently, we adopt attention manipulation to the layout-to-image models~\cite{zhou2024migc} to generate original and edited images that alter only relative positioning without changing other content.
\subsubsection{Visual Editing}
\noindent\textbf{Image Reference.} 
We are the first to incorporate additional visual input into instruction-based image editing. To reduce the cost of automated synthetic data generation, we leverage zero-shot image customization~\cite{chen2024anydoor} to synthesize images containing visual concepts. We repurpose the edited objects from the remove or replace steps and the corresponding masks to guide the target positioning within the edited image. Additionally, we introduce an ID extractor to embed visual concepts into the target image and a detail extractor to preserve fine content details. Finally, We construct edited images containing the visual concepts in the image reference, as shown in the \textcolor{red}{pipeline7} of Figure~\ref{total_pipelines}.

\noindent\textbf{Material Transfer.} Similar to the image reference, the material transfer requires injecting the material reference into the target image to achieve the material transfer effect. Considering the compatibility between materials and target objects, we further introduce depth estimation and latent illumination guidance for seamless material fusion. The total process is shown in the \textcolor{red}{Pipeline8} of the Figure~\ref{total_pipelines}.

% \noindent\textbf{Style Transfer.} Similar to the collect pipeline in Style Change, the only difference is that for certain styles, we represent them directly with an image (e.g., an image of fallen leaves to represent an autumn style), rather than using a style descriptor in the editing instruction. The instruction takes the form: ``Change the style of the image to match the given image."

\noindent\textbf{Visual Condition.} To support a broader range of visual editing types, we incorporate additional condition images as reference images through tool pockets from ControlNet~\cite{zhang2023adding}. We use tools to generate the corresponding conditional images and construct the corresponding editing instructions by templates. Notably, the edited images originate from other editing instructions, where the visual condition constructs new instruction pairs without generating additional edited images.

\section{Statistics of AnyEdit}\label{3}
We present detailed dataset statistics for all editing types in AnyEdit in Table~\ref{append:dataset_stastics}.
\begin{table}[h]
\centering
\resizebox{0.48\textwidth}{!}{ % 调整表格宽度为页面宽度
\begin{tabular}{lcc}
\hline
Editing Type & \#Instruction & \#Image \\

\hline
\textbf{Local Editing} \\
\quad Remove& 116013 & 116013 \\
\quad Replace& 97219 & 97219 \\
\quad Add& 390049 & 390049 \\
\quad Color Alter& 337078 & 337078 \\
\quad Appearance Alter& 79720 & 79720 \\
\quad Material Change& 21646 & - \\
\quad Action Change& 47210 & 47210 \\
\quad Textual Change& 2500 & 2500 \\
\quad Counting& 698 & 698 \\
\hline
\textbf{Global Editing} \\
\quad Background Change& 413570 & 413570 \\
\quad Tone Transfer& 553919 & 553919 \\
\quad Style Change& 27488 & - \\
\hline
\textbf{Camera Movement Editing} \\
\quad Movement& 7724 & 7724 \\
\quad Outpaint& 57462 & 57462 \\
\quad Rotation Change& 17022 & 17022 \\
\quad Resize& 10219 & 10219 \\
\hline
\textbf{Implicit Editing} \\
\quad Implicit Change& 10000 & 10000 \\
\quad Relation Change& 410 & 410 \\
\hline
\textbf{Visual Editing} \\
\quad Visual Sketch& 55385 & - \\
\quad Visual Scribble& 55385 & - \\
\quad Visual Segmentation& 55385 & - \\
\quad Visual Depth& 55385 & - \\
\quad Visual Layout& 55385 & - \\
\quad Material Transfer& 21646 & 21646 \\
\quad Image Reference& 17885 & 17885 \\
% Style Transfer& 172431 & 172431 \\
\hline
\textbf{Total} & 2506403 & 2180350\\\hline
\end{tabular}
}
\caption{The detailed statistics of the AnyEdit dataset.}\label{append:dataset_stastics}
\end{table}

\section{More Examples of AnyEdit}\label{4}
\begin{figure*}[h]
    \centering
        \includegraphics[width=1.\linewidth]{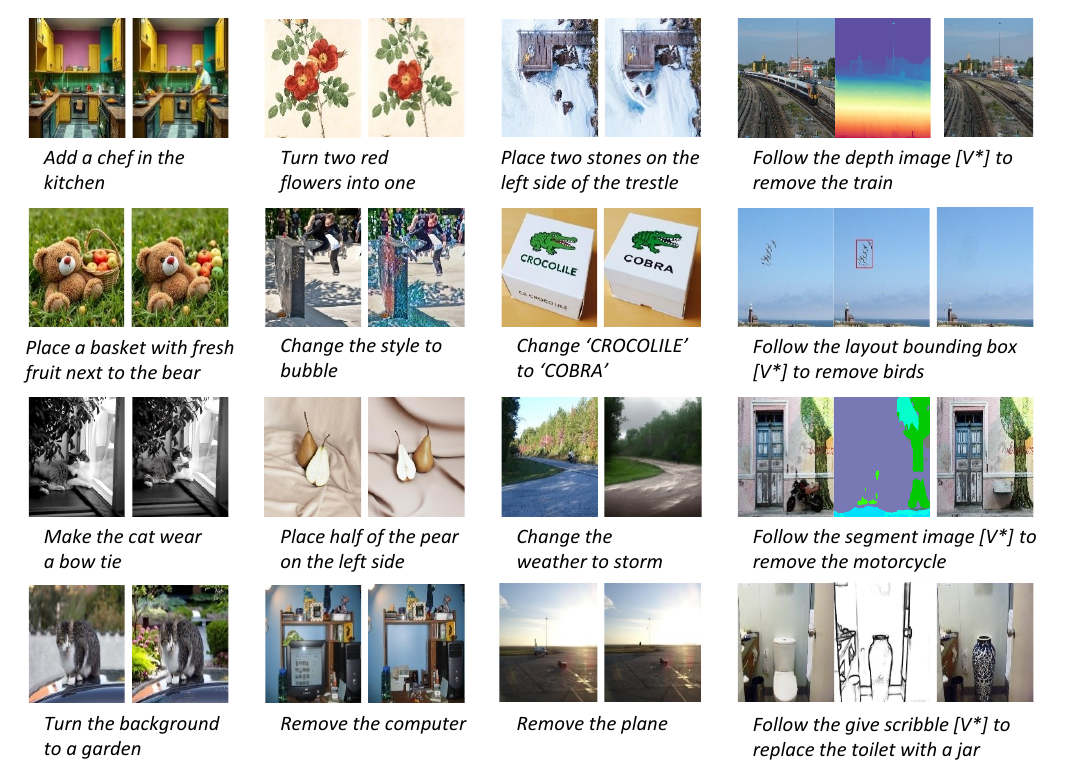}
    \caption{More Examples of AnyEdit dataset~(Part 1). textual instruction-based (first three columns) and visual instruction-based (last column) image editing.}\label{fig:more_anyedit_dataset_part1}
\end{figure*}

\begin{figure*}[h]
    \centering
        \includegraphics[width=1.\linewidth]{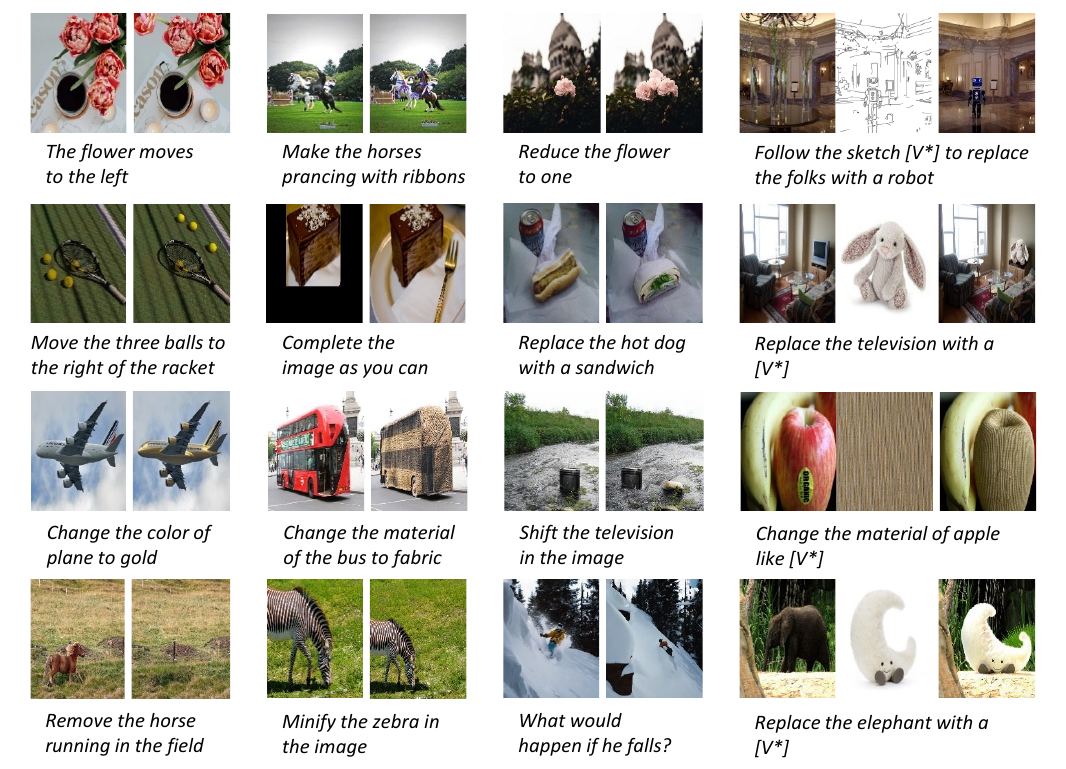}
    \caption{More Examples of AnyEdit dataset~(Part 2). textual instruction-based (first three columns) and visual instruction-based (last column) image editing.}\label{fig:more_anyedit_dataset_part2}
\end{figure*}

See figure~\ref{fig:more_anyedit_dataset_part1} and \ref{fig:more_anyedit_dataset_part2} for more data examples with various editing types in AnyEdit.
\section{AnyEdit-Test Benchmark}\label{5}
\begin{figure*}[ht]
\caption{More Examples of AnyEdit-Test with local editing categories.}
    \centering
    \vspace{-4mm}
        \includegraphics[width=0.9\linewidth]{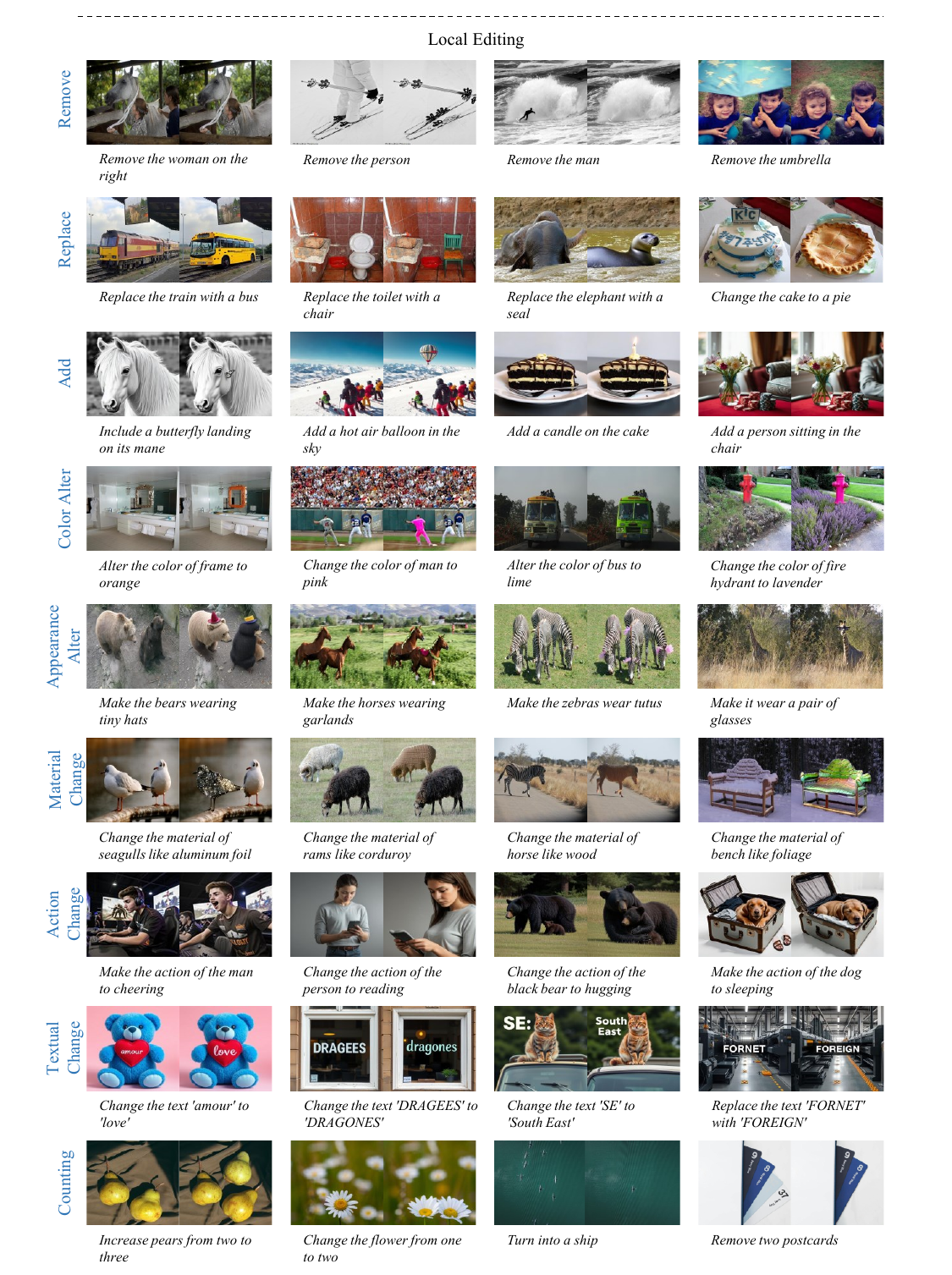}
  \vspace{-4mm} 
  \label{anytest-local}
\end{figure*}

\begin{figure*}[ht]
    \centering
        \includegraphics[width=0.9\linewidth]{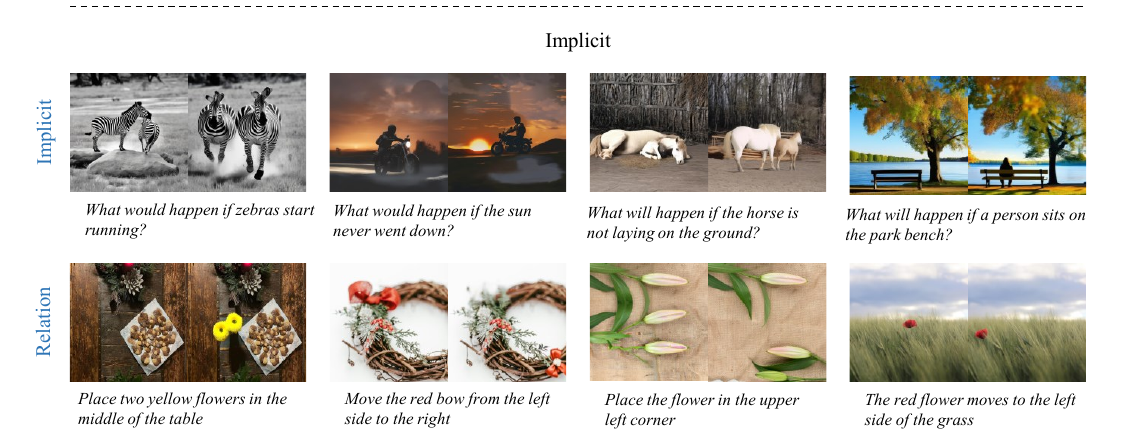}
  \vspace{-4mm}
  \caption{More Examples of AnyEdit-Test with implicit editing categories.}
    \label{anytest_implicit}
\end{figure*}

To comprehensively evaluate AnyEdit’s capabilities across a broader range of editing tasks, we carefully selected 50 example pairs from each type of editing task supported by AnyEdit. This selection process allowed us to construct a new test set, named \textbf{AnyEdit-Test}, designed specifically to provide a more rigorous assessment. The resulting dataset includes diverse and representative editing challenges, offering a well-rounded evaluation benchmark better to understand AnyEdit's performance across different task types. In this way, AnyEdit-Test not only broadens the scope of evaluation but also ensures that the test set includes a variety of editing complexities, thereby making the evaluation both more challenging and more insightful. In figures~\ref{anytest-local} to \ref{anytest_implicit}, the editing examples encompass all types from our AnyEdit dataset, all of which demonstrate excellent adherence to instructions and high visual fidelity. This further attests to the high quality and diversity of the editing data in AnyEdit. We will release this high-quality dataset and benchmark for community research.

\section{Detailed Experiments of AnyEdit-Test}\label{6}
We conduct detailed quantitative evaluations of 25 editing types in AnyEdit-Test, focusing on editing accuracy and content consistency.
Detailed results for the AnyEdit-Test benchmark can be seen in Table~\ref{app:exp2_table1}, \ref{app:exp2_table2}, \ref{app:exp2_table3}. We have the following observations: (1) Existing models often fail to ensure editing accuracy in complex tasks (\eg, significant reduction of CLIPim in action change, rotation, outpainting shown in Tab.~\ref{app:exp2_table1} \& ~\ref{app:exp2_table2}, exposing the limitations of current benchmarks for complex tasks. For fine-grained editing tasks, models frequently struggle to maintain the integrity of image content while making precise modifications~(\eg, L1 nearly doubled performance degradation in action change and textual change). These tasks demand both a high level of fine-grained control and the ability to preserve the original context. These limitations highlight a fundamental gap in existing benchmarks and AnyEdit-Test, which is more comprehensive for complex, real-world editing demands. (2) Even for common tasks in AnyEdit-Test, some previous SOTA models show a notable performance drop compared to existing benchmarks, revealing the limitations of current benchmarks in multi-scene editing. While many state-of-the-art models have achieved impressive results on conventional benchmarks, they struggle to generalize to more diverse, multi-scene editing tasks that are present in AnyEdit-Test. This performance drop highlights the limitations of traditional benchmarks when adapting to the increased diversity of multi-scene editing. In contrast, AnyEdit-Test introduces a broader range of editing scenes, making it a more accurate reflection of real-world scenarios.
\section{Implementation Details}\label{7}
\subsection{AnySD Architecture}
AnySD is a diffusion model designed to handle a broad range of editing tasks through language-based instructions. Given the distinct demands of each edit type, which require the model to selectively focus on different elements—such as faithfully preserving visual likeness in visual instructions or altering style while retaining the original image composition in style transfer—we adopt a Mixture of Experts (MoE) architecture~\citep{masoudnia2014mixture}.

% \textbf{Visual Prompt Projector}.
% For the visual prompt, we first utilize the frozen CLIP image encoder to extract image features. The CLIP model, a multimodal framework trained through contrastive learning on a large dataset of image-text pairs, generates image features that are well-aligned with their corresponding captions, effectively capturing the rich content and style of the images. To facilitate the decomposition of the image embedding, we employ a small trainable projector comprising a linear layer and Layer Normalization~\cite{ba2016layer}, which projects the image embedding into a sequence of features $z_v$. The dimensionality of these image features is matched to that of the text features in the pre-trained diffusion model.

% \textbf{Learned Task Embeddings}.
% To guide the generation process towards the appropriate task, we learn an embedding vector~\citep{sheynin2024emu} for each task in the dataset. During training, for a given sample from our dataset, we utilize the task index, $i$, to retrieve the corresponding embedding vector, $v_i$, from an embedding table, which we optimize jointly with the model weights. The embedding vector $v_i$ serves two functions: (1) It is concatenated directly with $z_v$ (where $z_v$ is set to a zero tensor in the absence of a visual prompt) and is used as condition $c_V$ for the UNet, thereby influencing the diffusion output; (2) It acts as input for the router~\citep{zhou2022mixture}, distributing weights among different experts in the MoE block.

% \textbf{MoE Block}. 
The visual condition $c_V$ is integrated into the pretrained UNet~\cite{ronneberger2015u} by the adapted modules with decoupled cross-attention to avoid disrupt the edit instruction condition. Each MoE block share the same language attention layer but diverse in the attention for $c_V$ and the weights are distributed by the router based on the task embedding.

In the original SD model, given the query features $\mathbf{Z}$ and the text features $\boldsymbol{z}_{t}$, the output of cross-attention $\mathbf{Z}'$ can be defined by the following equation:
\begin{equation}
\begin{split}
\mathbf{Z}'=\text{Attention}(\mathbf{Q},\mathbf{K},\mathbf{V}) = \text{Softmax}(\frac{\mathbf{Q}\mathbf{K}^{\top}}{\sqrt{d}})\mathbf{V},
\\
\end{split}
\end{equation}
where $\mathbf{Q}=\mathbf{Z}\mathbf{W}_q$, $\mathbf{K}=\boldsymbol{z}_{t}\mathbf{W}_k$, $\mathbf{V}=\boldsymbol{z}_{t}\mathbf{W}_v$ are the query, key, and values matrices of the attention operation respectively, and $\mathbf{W}_q$, $\mathbf{W}_k$, $\mathbf{W}_v$ are the weight matrices of the trainable linear projection layers.

To achieve separate attention mechinism, we add a new cross-attention layer for each cross-attention layer in the original UNet model to insert image features. Given the $\boldsymbol{c}_{V}$, the output of new cross-attention $\mathbf{Z}''$ is computed as follows:
\begin{equation}
    \begin{split}
    \mathbf{Z}''=\text{Attention}(\mathbf{Q},\mathbf{K}',\mathbf{V}') = \text{Softmax}(\frac{\mathbf{Q}(\mathbf{K}')^{\top}}{\sqrt{d}})\mathbf{V}',\\
    \end{split}
\end{equation}
where, $\mathbf{K}'=\boldsymbol{c}_{v}\mathbf{W}'_k$ and $\mathbf{V}'=\boldsymbol{c}_{v}\mathbf{W}'_v$ are the query, key, and values matrices from the image features. $\mathbf{W}'_k$ and $\mathbf{W}'_v$ are the corresponding weight matrices. In order to speed up the convergence, $\mathbf{W}'_k$ and $\mathbf{W}'_v$ are initialized from $\mathbf{W}_k$ and $\mathbf{W}_v$.
Then, we simply add the output of image cross-attention to the output of text cross-attention:
\begin{equation}
\begin{aligned}
\mathbf{Z}^{new} &= \text{Softmax}\left(\frac{\mathbf{Q}\mathbf{K}^{\top}}{\sqrt{d}}\right)\mathbf{V} 
+ \text{Softmax}\left(\frac{\mathbf{Q}(\mathbf{K}')^{\top}}{\sqrt{d}}\right)\mathbf{V}' \\
% \text{where} \quad &\mathbf{Q} = \mathbf{Z}\mathbf{W}_q, \quad \mathbf{K} = \boldsymbol{z}_{t}\mathbf{W}_k, \quad \mathbf{V} = \boldsymbol{z}_{t}\mathbf{W}_v, \\
% &\mathbf{K}' = \boldsymbol{c}_{i}\mathbf{W}'_k, \quad \mathbf{V}' = \boldsymbol{c}_{i}\mathbf{W}'_v
\end{aligned}
\end{equation}
\subsection{CFG for Three Conditionings}~\label{sec4:1}
AnySD is based on the latent diffusion model architecture~\cite{rombach2022high, song2020denoising, sd1.5} to support high-resolution image generation and incorporated variational autoencoder~\cite{kingma2013auto} with encoder $\mathcal{E}$ and decoder $\mathcal{D}$, with estimating the score~\cite{hyvarinen2005estimation} of a data distribution. 
To support image conditioning, we add additional input channels to the first convolutional layer on the simple text to image diffusion model~\cite{sd1.5}, concatenating $z_t$ and $\mathcal{E}(c_I)$, following InstructPix2Pix~\citep{brooks2023instructpix2pix}. 

For an image $x$, the diffusion process adds noise to the encoded latent $z = \mathcal{E}(x)$ producing a noisy latent $z_t$ where the noise level increases over timesteps $t \in T$. We learn a network $\epsilon_\theta$ that predicts the noise added to the noisy latent $z_t$ given original image conditioning $c_I$, text edit instruction conditioning $c_T$ and visual prompt conditioning $c_V$. We minimize the following latent diffusion objective:

\vspace{-4mm}
\begin{equation}
L = \mathbb{E}_{\mathcal{E}(x), \mathcal{E}(c_I), c_T, \epsilon \sim \mathcal{N}(0, 1), t }\Big[ \Vert \epsilon - \epsilon_\theta(z_{t}, t, \mathcal{E}(c_I), c_T, c_V)) \Vert_{2}^{2}\Big]
\end{equation}
\vspace{-1mm}

Classifier-free diffusion guidance (CFG)~\cite{ho2022classifier, liu2022compositional} effectively shifts probability mass toward data where an implicit classifier $p_{\theta}(c|z_t)$ assigns high likelihood to the conditioning $c$. Training for unconditional denoising is done by simply setting the conditioning to a fixed null value $c\tight{=}\varnothing$ at some frequency during training. At inference time, with a guidance scale $s\ge1$, the modified score estimate $\tilde{e_{\theta}}(z_t, c)$ is extrapolated in the direction toward the conditional $e_{\theta}(z_t, c)$ and away from the unconditional $e_{\theta}(z_t, \varnothing)$.

\vspace{-3.5mm}
\begin{equation}
    \tilde{e_{\theta}}(z_t, c) = e_{\theta}(z_t, \varnothing) + s \cdot (e_{\theta}(z_t, c) - e_{\theta}(z_t, \varnothing))
    \label{eq:cfg}
\end{equation}
\vspace{-3mm}

For our task, the score network $e_{\theta}(z_t, c_I, c_T, c_V)$  has three conditionings: the input image $c_I$, text instruction $c_T$ and visual prompt $c_V$.
We introduce two guidance scales, $s_I$, $s_T$ and $s_V$ which can be adjusted to trade off how strongly each condition.
Our modified score estimate is as follows: 

\vspace{-3.5mm}
\begin{equation}
\begin{split}
    \tilde{e_{\theta}}&(z_t, c_I, c_T, c_V) = \: e_{\theta}(z_t, \varnothing, \varnothing, \varnothing) \\ 
    &+ s_I \cdot (e_{\theta}(z_t, c_I, \varnothing, \varnothing) - e_{\theta}(z_t, \varnothing, \varnothing, \varnothing)) \\ 
    &+ s_T \cdot (e_{\theta}(z_t, c_I, c_T, \varnothing) - e_{\theta}(z_t, c_I, \varnothing, \varnothing)) \\ 
    &+ s_V \cdot (e_{\theta}(z_t, c_I, c_T, c_V) - e_{\theta}(z_t, c_I, c_T, \varnothing))
    \label{eq:cfg2}
\end{split}
\end{equation}

% See Appendix~\ref{sec:cfg2a} for details of our classifier-free guidance formulation. 
\subsection{Classifier-free Guidance Details}\label{sec:cfg2a}
As discussed in Section~\ref{sec4:1}, we apply classifier-free guidance with respect to three conditionings: the input image $c_I$ , the text instruction $c_T$ and the visual prompt with task embedding $c_V$. We introduce separate guidance scales $s_I$, $s_T$ and $s_V$ that enable separately trading off the strength of each conditioning.

When ignoring $c_V$, we can have the modified score estimate as InstructPix2Pix~\citep{brooks2023instructpix2pix}:
\vspace{-2.5mm}
\begin{equation*}
\begin{split}
    \tilde{e_{\theta}}(z_t, c_I, c_T) = &\: e_{\theta}(z_t, \varnothing, \varnothing) \\ &+ s_I \cdot (e_{\theta}(z_t, c_I, \varnothing) - e_{\theta}(z_t, \varnothing, \varnothing)) \\ &+ s_T \cdot (e_{\theta}(z_t, c_I, c_T) - e_{\theta}(z_t, c_I, \varnothing))
\end{split}
\end{equation*}
\vspace{0.1mm}

Below is the modified score estimate for our model with classifier-free guidance on three conditions (copied from Equation~\ref{eq:cfg2}):

\vspace{-2.5mm}
\begin{equation*}
\begin{split}
        \tilde{e_{\theta}}&(z_t, c_I, c_T, c_V) = \: e_{\theta}(z_t, \varnothing, \varnothing, \varnothing) \\ 
    &+ s_I \cdot (e_{\theta}(z_t, c_I, \varnothing, \varnothing) - e_{\theta}(z_t, \varnothing, \varnothing, \varnothing)) \\ 
    &+ s_T \cdot (e_{\theta}(z_t, c_I, c_T, \varnothing) - e_{\theta}(z_t, c_I, \varnothing, \varnothing)) \\ 
    &+ s_V \cdot (e_{\theta}(z_t, c_I, c_T, c_V) - e_{\theta}(z_t, c_I, c_T, \varnothing))
\end{split}
\end{equation*}
\vspace{0.1mm}

Our generative model learns $P(z|c_I,c_T)$, the probability distribution of image latents $z = \mathcal{E}(x)$ conditioned on an input image $c_I$, a text instruction $c_T$ and the visual prompt with task embedding $c_V$. We arrive at our particular classifier-free guidance formulation by expressing the conditional probability as follows:

\vspace{-2.5mm}
\begin{equation*}
\begin{split}
    P(z|c_T,c_I,c_V) &= \frac{P(z,c_T,c_I, c_V)}{P(c_T,c_I, c_V)} \\
    &= \frac{P(c_T|c_I,c_V,z)P(c_I|c_V,z)P(c_V|z)P(z)}{P(c_T,c_I, c_V)}
\end{split}
\end{equation*}
\vspace{0.1mm}

Diffusion models estimate the score~\cite{hyvarinen2005estimation} of the data distribution, i.e., the derivative of the log probability. Taking the logarithm gives us the following expression:

\vspace{-2.5mm}
\begin{equation*}
\begin{split}
\log(P(z|c_T,c_I,c_V)) = &\: \log(P(c_T|c_I,c_V,z)) \\& + \log(P(c_I|c_V,z)) \\ &+ \log(P(c_V|z))  +\log(P(z)\\ &- \log(P(c_T,c_I,c_V))
\end{split}
\end{equation*}
\vspace{0.1mm}

\noindent Taking the derivative and rearranging we attain:

\vspace{-2.5mm}
\begin{equation*}
\begin{split}
\nabla_z\log(P(z|c_T,c_I,c_V)) = &\: \nabla_z\log(P(z)) \\ &+ \nabla_z\log(P(c_V|z)) \\ &+ \nabla_z\log(P(c_i|c_V, z))\\ &+ \nabla_z\log(P(c_T|c_I,c_V,z))
\end{split}
\end{equation*}
\vspace{0.1mm}

This corresponds with the terms in our classifier-free guidance formulation in Equation~\ref{eq:cfg2}.

\subsection{Supporting Tasks for AnySD}\label{app:support_task}
In general, the editing tasks supported by AngSD align with those listed for AnyEdit-Test. For each task, we utilize a distinct learned task embedding of size $N$ (where $N$ matches the dimensionality of CLIP).

Additionally, there are substantial differences between various types of tasks. Consequently, we employ a Mixture of Experts (MoE) framework. Specifically, our expert categorization is detailed in Table~\ref{support_task}.

\begin{table}[h]
\centering
\resizebox{0.49\textwidth}{!}{
\begin{tabular}{p{0.3\linewidth}  p{0.65\linewidth}}
\hline
 \textbf{Expert} & \textbf{Supporting tasks} \\\hline
 \quad Expert 1 & tone transfer, background change, style transfer, style change\\
  \quad Expert 2 & movement, outpaint, resize, rotation\\
  \quad Expert 3 & visual bbox\\
  \quad Expert 4 & visual depth\\
  \quad Expert 5 & visual material transfer\\
  \quad Expert 6 & visual reference\\
  \quad Expert 7 & visual scribble\\
  \quad Expert 8 & visual segment\\
  \quad Expert 9 & visual sketch\\
  \hline
\end{tabular}
}
\caption{Expert division for various editing tasks of AnySD.}\label{support_task}
\end{table}

\subsection{Training Details}\label{app:train_details}
\textbf{Stage I: Instruction Understanding.} In this stage, we use the dataset type of background change, tone transfer, remove, replace, add, color, and appearance change in AnyEdit to enhance the model’s instruction-following capability. Following prior works~\cite{brooks2023instructpix2pix, zhang2024magicbrush, sheynin2024emu}, we train our image editing model for 110,000 steps using four 48GB NVIDIA A6000 GPUs for 280 hours. Specifically, the training is conducted at a resolution of 256 × 256 with a total batch size of 1024 (gradient\_accumulation\_steps=2, batch\_size=128 per GPU). We apply random horizontal flip augmentation and crop augmentation, where images are first randomly resized between 256 and 288 pixels, followed by cropping to 256 × 256. The model is trained with a learning rate of $10^{-4}$, without any learning rate warm-up. We initialize our model using the EMA weights from the Stable Diffusion 1.5 checkpoint~\citep{sd1.5} and adopt other training configurations from the publicly available Stable Diffusion codebase. Although the model is trained at a resolution of 256 × 256, it generalizes well to a resolution of 512 × 512 during inference. All results presented in this paper are generated at 512 × 512 resolution, with an Euler ancestral sampler and the denoising variance schedule proposed by~\cite{karras2022elucidating}.

\textbf{Stage II: Task Tuning}. In the second stage, we train our model on the entire AnyEdit dataset to adapt the model to the task-specific editing granularity. We utilize the task embedding and each expert is described in Appendix~\ref{app:support_task}. Unlike the first stage, we do not use EMA (Exponential Moving Average) for training~\cite{hunter1986exponentially}. Additionally, we set the training resolution to 512 × 512, compared to 256 × 256 in the first stage, to achieve better editing results for specific tasks. The model is trained with a learning rate of $10^{-4}$, without any learning rate warm-up. The second stage is trained for 400,000 steps using eight 48GB NVIDIA A6000 GPUs over approximately 150 hours.

\subsection{Baselines Details}\label{app:baseline}
We establish the models in Table~\ref{table:model_hp} as baselines, organized into two categories: instruction-based and specific image editing methods. The former utilizes natural instructions to guide the editing process, while the latter relies on global descriptions of the target image to enable editing. Instruction-based image editing methods include InstructPix2Pix~\cite{brooks2023instructpix2pix}, HIVE~\citep{zhang2023hive}, UltraEdit~\cite{zhao2024ultraedit}, EMU-Edit~\cite{sheynin2024emu}, and MagicBrush~\citep{kawar2023imagic}. The specific image editing methods include Null Text Inversion~\citep{mokady2023null}, while the visual condition editing methods include Uni-ControlNet~\citep{zhao2023uni}.

\textbf{Instruction-Based Editing Methods}:
\begin{itemize}
    \item \textit{InstructPix2Pix}~\citep{brooks2023instructpix2pix}: Utilizes automatically generated instruction-based image editing data to fine-tune Stable Diffusion~\cite{rombach2022high}, enabling instruction-based image editing during inference without requiring any test-time tuning. We use the official Hugging Face to implement it.
    \item \textit{HIVE}~\citep{zhang2023hive}: Trained with supplementary data akin to InstructPix2Pix, HIVE undergoes further fine-tuning using a reward model derived from human-ranked data. Notably, the edited output of HIVE is not square; instead, it is scaled to preserve the original aspect ratio, with the longer side resized to 512 pixels. We utilized two models, the weighted reward (SD1.5) and the conditional reward (SD1.5), referred to as \textbf{HIVE$^w$} and \textbf{HIVE$^c$}, respectively.
    \item \textit{UltraEdit}~\citep{zhao2024ultraedit}. It is trained on nearly 4 million instruction-based editing samples based on the Stable Diffusion 3~\citep{sd3} and supports free-form and mask-form inputs to enhance editing performance. To ensure comparison fairness, we utilize its freeform model for all experiments. Notably, since it is trained on SD3, its performance cannot accurately reflect the improvements brought by its editing data.
    \item \textit{EMU-Edit}~\citep{sheynin2024emu}. It is a fine-tuned editing model that integrates recognition and generation tasks. Although it provides promising results, the model is not open-sourced. Therefore, we only conduct comparisons of it and AnyEdit on public benchmarks to demonstrate the superiority of our approach.
    \item \textit{MagicBrush}~\citep{kawar2023imagic}: MagicBrush curates a well-structured editing dataset with detailed human annotations and fine-tunes its model on this dataset using the InstructPix2Pix~\cite{brooks2023instructpix2pix} framework. Therefore, we use this as a baseline to fairly compare the improvement in editing capabilities brought by the AnyEdit dataset in our experiments.
\end{itemize}

\textbf{Specific Editing Methods}:
\begin{itemize}
\item \textit{Null Text Inversion}~\citep{mokady2023null}: This method inverts the source image using the DDIM~\citep{song2020denoising} trajectory and performs edits during the denoising process by controlling cross-attention between text and image. Notably, Null Text Inversion requires that "attention replacement editing can only be applied to prompts of the same length." Therefore, if the input and output captions differ in length, we align the word count by truncating the longer caption. Additionally, it is worth mentioning that the official repository performs a center crop when processing non-square images, and we adhered to this setting.
\end{itemize}

\textbf{Visual Condition Methods}:
\begin{itemize}
\item  \textit{Uni-controlnet}~\citep{zhao2023uni}. Uni-ControlNet categorizes conditions into two groups: local and global. By adding only two additional adapters, the cost of fine-tuning and the model size are significantly reduced. For local controls, we introduce a multi-scale conditional injection strategy, while for global controls, a global condition encoder is used to convert them into conditional tokens, which then interact with the incoming features. To let it support visual reference editing, we use the HED condition as the channel of reference image input.
\end{itemize}

\begin{table}[!ht]
\centering
\resizebox{0.48\textwidth}{!}{
    \begin{tabular}{p{0.4\linewidth}  p{0.55\linewidth}}
    \hline
    Method & Configuration \\
    \hline
        InstructPix2Pix~\citep{brooks2023instructpix2pix} & num\_inference\_steps=10,\newline image\_guidance\_scale=1 \\
        MagicBrush~\citep{zhang2024magicbrush}   &seed=42, guidance\_scale=7 \newline num\_inference\_steps=20,\newline image\_guidance\_scale=1.5  \\
        UltraEdit~\citep{zhao2024ultraedit}& negative\_prompt="",
            \newline num\_inference\_steps=50,
            \newline image\_guidance\_scale=1.5,
            \newline guidance\_scale=7.5 \\
        HIVE$^w$~\citep{zhang2023hive} & steps=100 
            \newline text\_cfg\_scale=7.5,
            \newline image\_cfg\_scale=1.5\\
        HIVE$^c$~\citep{zhang2023hive} & steps=100 
            \newline text\_cfg\_scale=7.5,
            \newline image\_cfg\_scale=1.5\\
        Null-Text~\citep{mokady2023null}  & cross\_replace\_steps.default=0.8, \newline self\_replace\_steps= 0.5, 
        \newline blend\_words=None,
         \newline equilizer\_params=None \\
         Uni-controlnet~\citep{zhao2023uni} & num\_samples = 1
            \newline image\_resolution = 512
            \newline strength = 1
            \newline global\_strength = 1
            \newline low\_threshold = 100
            \newline high\_threshold = 200
            \newline value\_threshold = 0.1
            \newline distance\_threshold = 0.1
            \newline alpha = 6.2
            \newline ddim\_steps = 50
            \newline scale = 7.5
            \newline seed = 42
            \newline eta = 0.0
            \newline a\_prompt = 'best quality, extremely detailed'
            \newline n\_prompt = 'longbody, lowres, bad anatomy, bad hands, missing fingers, extra digit, fewer digits, cropped, worst quality, low quality'
            \\
    \hline
    \end{tabular}
}
\caption{\textbf{Configuration of the baselines for AnyEdit-Test}. We strictly adhered to the default hyperparameters provided in the official repositories or Huggingface implementations of these baseline models.}\label{table:model_hp}
\end{table}

\subsection{Details on Benchmarks and Metrics}\label{app:matrix}
\textbf{Metrics and code}.
For metrics evaluation, we closely follow the MagicBrush evaluation script without any modifications. Following previous works~\citep{bai2024meissonic, zhang2024magicbrush, zhao2024ultraedit}, we employ L1 metrics to measure pixel-level differences between the generated and ground truth images. Additionally, CLIP and DINO similarities are used to assess the overall similarity with the ground truth, while CLIP-T measures text-image alignment based on local descriptions and the CLIP embedding of generated images. Furthermore, CLIP text-image similarity between the edited image and the output caption, as well as CLIP text-image direction similarity (CLIPdir), are employed to evaluate the model's instruction-following ability. Specifically, CLIPdir measures the agreement between changes in caption embedding and changes in image embedding. While the Emu Edit Test eliminates bias and overfitting at the image level by not providing ground truth images, the evaluation metrics still implicitly assess the model’s editing capabilities.

\noindent\textbf{EMU-Edit-Test}.
We observe that the original EMU-Edit~\citep{sheynin2024emu} paper and dataset don't specify the versions of CLIP~\citep{clip} and DINO~\citep{caron2021emerging} used. To maintain consistency with other benchmarks, we follow the settings from the MagicBrush repository~\cite{zhang2024magicbrush}, modifying only the evaluation dataset to EMU-Edit-Test.

\noindent\textbf{MagicBrush-Test}. 
MagicBrush is designed to evaluate both the single-turn and multi-turn image editing capabilities of models. It provides annotator-defined instructions and editing masks, along with ground truth images generated by DALLE-2~\citep{ramesh2022hierarchical}, facilitating a more effective metric-based assessment of the model’s editing performance. However, the dataset suffers from inherent biases. During data collection, annotators are instructed to use the DALLE-2 image editing platform to generate the edited images, making the benchmark biased towards images and editing instructions that the DALLE-2 editor can successfully follow. This bias may limit the dataset’s diversity and complexity. The baseline results in Table \textcolor{cvprblue}{4} of the main paper correspond to EMU-Edit~\citep{sheynin2024emu}.

\section{Qualitative and Human Evaluations}\label{8}
\subsection{Human Evaluation}
We conduct comprehensive human evaluations to assess both the consistency and image quality of generated images across three tasks: multiple-choice comparison, pairwise comparison, and individual image assessment. For each task, we randomly sample 100 images from AnyEdit-Test (excluding the visual instruction component). These images are evenly distributed among evaluators, and where applicable, we report averaged scores. Specifically, we evaluate three methods, comparing our approach against four SOTA editing methods: InstructPix2Pix~\citep{brooks2023instructpix2pix}, MagicBrush~\citep{zhang2024magicbrush}, HIVE$^w$~\citep{zhang2023hive}, HIVE$^c$~\citep{zhang2023hive}, UltraEdit (SD3)~\citep{zhao2024ultraedit} and our method.

\textbf{Multiple-Choice Comparison}. In this task, evaluators select the best-edited image based on consistency and image quality. As shown in Table~\ref{tab:he_multichoice}, our method demonstrates superior performance, significantly surpassing the other methods, which emphasizes the effectiveness of training on our AnyEdit dataset. Notably, while MagicBrush and UltraEdit score highly in automated evaluations, their performance in human assessments is comparatively lower, especially in instruction consistency. This discrepancy highlights the limitations of current automatic metrics, which focus primarily on image quality and may not fully capture human preferences, underscoring the need for future research to develop more robust and aligned evaluation metrics.

\begin{table}[t]
\centering
\resizebox{0.4\textwidth}{!}{
    \begin{tabular}{l|ccccc}
    \toprule
         & \textbf{Consistency } & \textbf{Image Quality } \\
    \midrule
    MagicBrush~\citep{zhang2024magicbrush} &10 &9 \\
    HIVE$^w$~\citep{zhang2023hive} &15 &17 \\
    HIVE$^c$~\citep{zhang2023hive} &20 &21 \\
    UltraEdit (SD3)~\citep{zhao2024ultraedit} &13 &17 \\
        AnySD &\textbf{42} &\textbf{36} \\
    \bottomrule
    \end{tabular}
}
\caption{Multi-choice comparison of four methods. The numbers represent the frequency of each method being chosen as the best for each aspect.}\label{tab:he_multichoice}
\end{table}

\textbf{One-on-One Comparison}.The one-on-one comparison provides a detailed and nuanced assessment of the edited results by juxtaposing them with robust baselines. Evaluators are instructed to select the preferred option based on both consistency and image quality. As shown in Table~\ref{tab:he_comparative}, AnySD consistently outperforms the alternatives in both aspects, with a majority of evaluators favoring AnySD's results in these direct comparisons.

\begin{table}[t]
\centering
\resizebox{0.4\textwidth}{!}{
    \begin{tabular}{c|cc}
    \toprule
     & \textbf{Consistency } & \textbf{Image Quality } \\
    \midrule
    MagicBrush~\citep{zhang2024magicbrush} &0.35  & 0.27\\
    HIVE$^w$~\citep{zhang2023hive} & 0.42 & 0.41 \\
    HIVE$^c$~\citep{zhang2023hive} & 0.47 & 0.48 \\
    UltraEdit (SD3)~\citep{zhao2024ultraedit} & 0.35 & 0.37 \\
    \bottomrule
    \end{tabular}
}
\caption{One-on-one comparisons. The numbers in the table indicate the percent of each method being chosen as the better option compared with the AnySD's results.}\label{tab:he_comparative}
\end{table}

\textbf{Individual Evaluation}. 
The individual evaluation utilizes a 5-point Likert scale to gather subjective feedback on image quality generated by four distinct models. Evaluators rate each image from 1 to 5, focusing on both consistency and overall quality. As shown in Table~\ref{tab:he_standalone}, the results clearly indicate that AnySD outperforms the other baselines, underscoring the advantages of training or fine-tuning models on the AnyEdit dataset.

\begin{table}[t]
\centering
\resizebox{0.4\textwidth}{!}{
    \begin{tabular}{l|cc}
    \toprule
     & \textbf{Consistency } & \textbf{Image Quality } \\
    \midrule
    MagicBrush~\citep{zhang2024magicbrush} & 3.3 & 3.1 \\
    HIVE$^w$~\citep{zhang2023hive} & 4.1  & 3.8 \\
    HIVE$^c$~\citep{zhang2023hive} & 4.2 & 4.2 \\
    UltraEdit (SD3)~\citep{zhao2024ultraedit}  & 3.7 & 4.0 \\
        AnySD &\textbf{4.3} & \textbf{4.4}\\
    \bottomrule
    \end{tabular}
}
\caption{Individual evaluation using a 5-point Likert scale. 
The numbers in the table represent the average scores calculated for each aspect.}\label{tab:he_standalone}
\end{table}

\subsection{Qualitative Evaluation on Different Benchmarks}
\begin{figure*}[h]
    \centering
        \includegraphics[width=1.\linewidth]{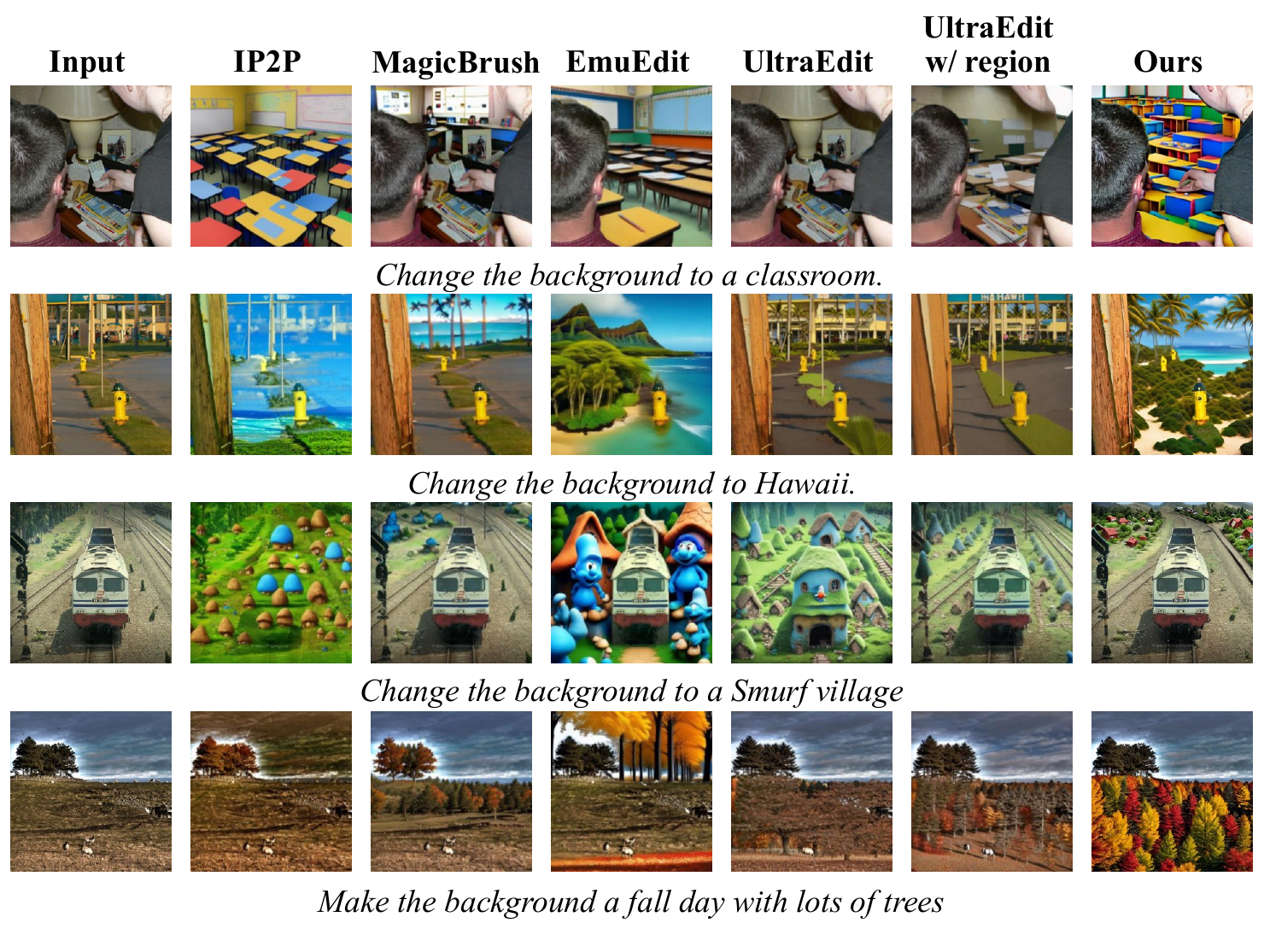}

    \caption{More qualitative results of the EMU-Edit Test for the editing of background change.}\label{append:editing_result_1}

\end{figure*}

\noindent\textbf{Detailed Results for EMU-Edit Test.}
More qualitative results of the EMU-Edit Test are shown in Figure~\ref{append:editing_result_1}. We observe that AnySD can effectively distinguish between the foreground and background of an image solely based on editing instructions, accurately modifying the background while preserving the content of the foreground.

\begin{figure*}[h]
    \centering
        \includegraphics[width=1.\linewidth]{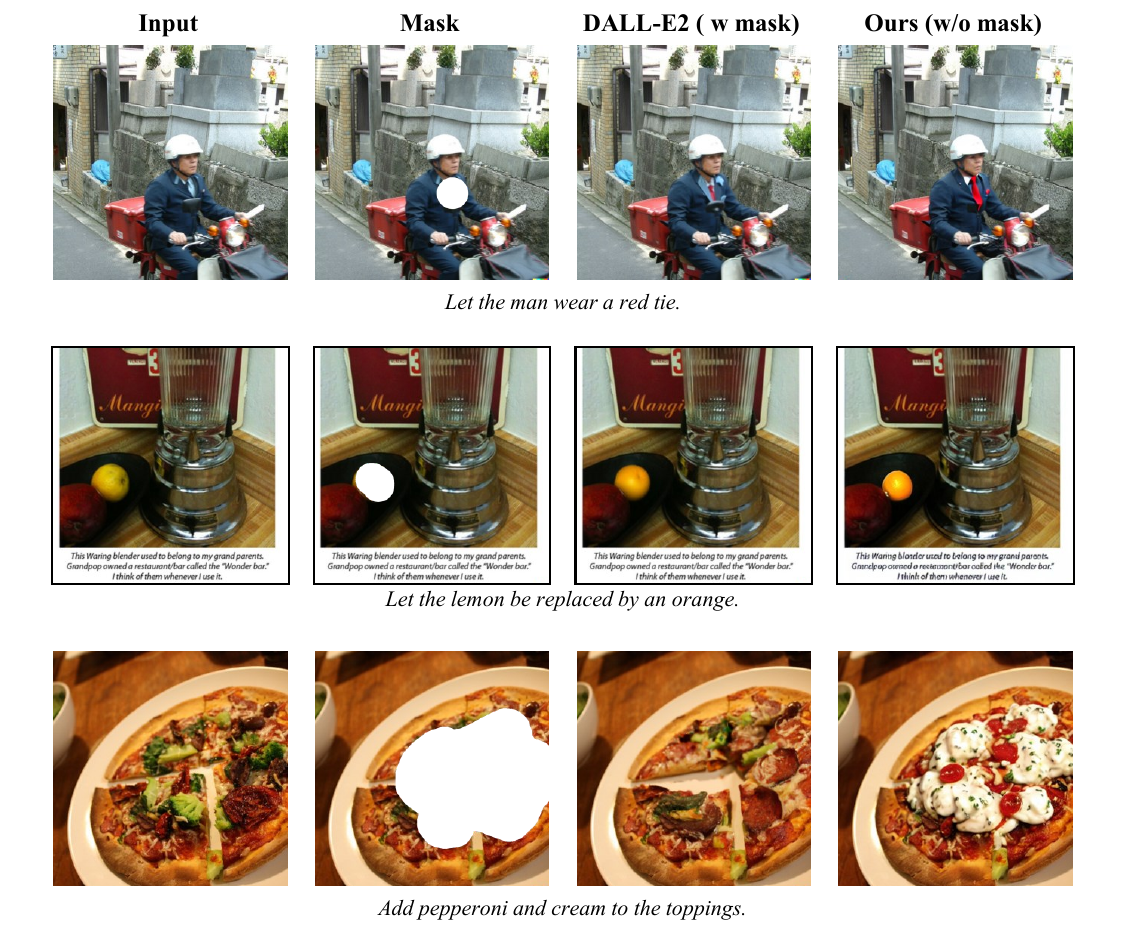}

    \caption{More qualitative results of the MagicBrush Test for local editing. The mask is used solely to supervise the editing process in DALLE-2~\cite{ramesh2022hierarchical} and is not provided as input to our method.}\label{append:editing_result_0}

\end{figure*}

\noindent\textbf{Detailed Results for MagicBrush Benchmark.}
More qualitative results of the MagicBrush Test are shown in Figure~\ref{append:editing_result_0}. We visually compare the performance of our method on local editing with the SOTA mask-based Editing model~(\textit{i.e.}, DALLE-2~\cite{ramesh2022hierarchical}). We notice that even without masks as supervision signals, our method accurately performs edits in specific regions, benefiting from the well-aligned editing data provided by AnyEdit.

\begin{figure*}[h]
    \centering
        \includegraphics[width=0.8\linewidth]{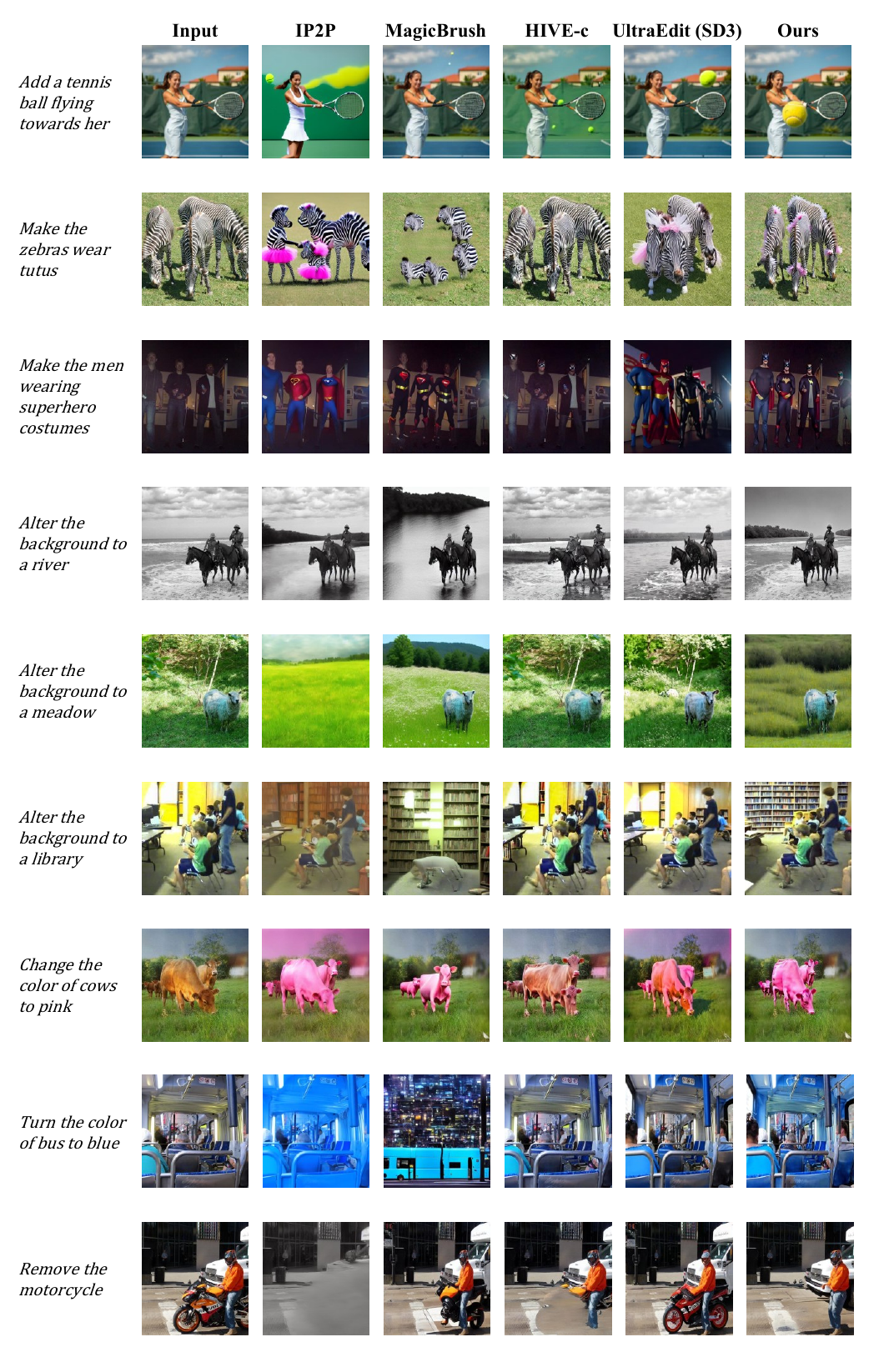}

    \caption{More qualitative evaluation of our model trained on AnyEdit across AnyEdit-Test benchmark~(Part I).}\label{append:editing_result_2}

\end{figure*}

\begin{figure*}[h]
    \centering
        \includegraphics[width=0.8\linewidth]{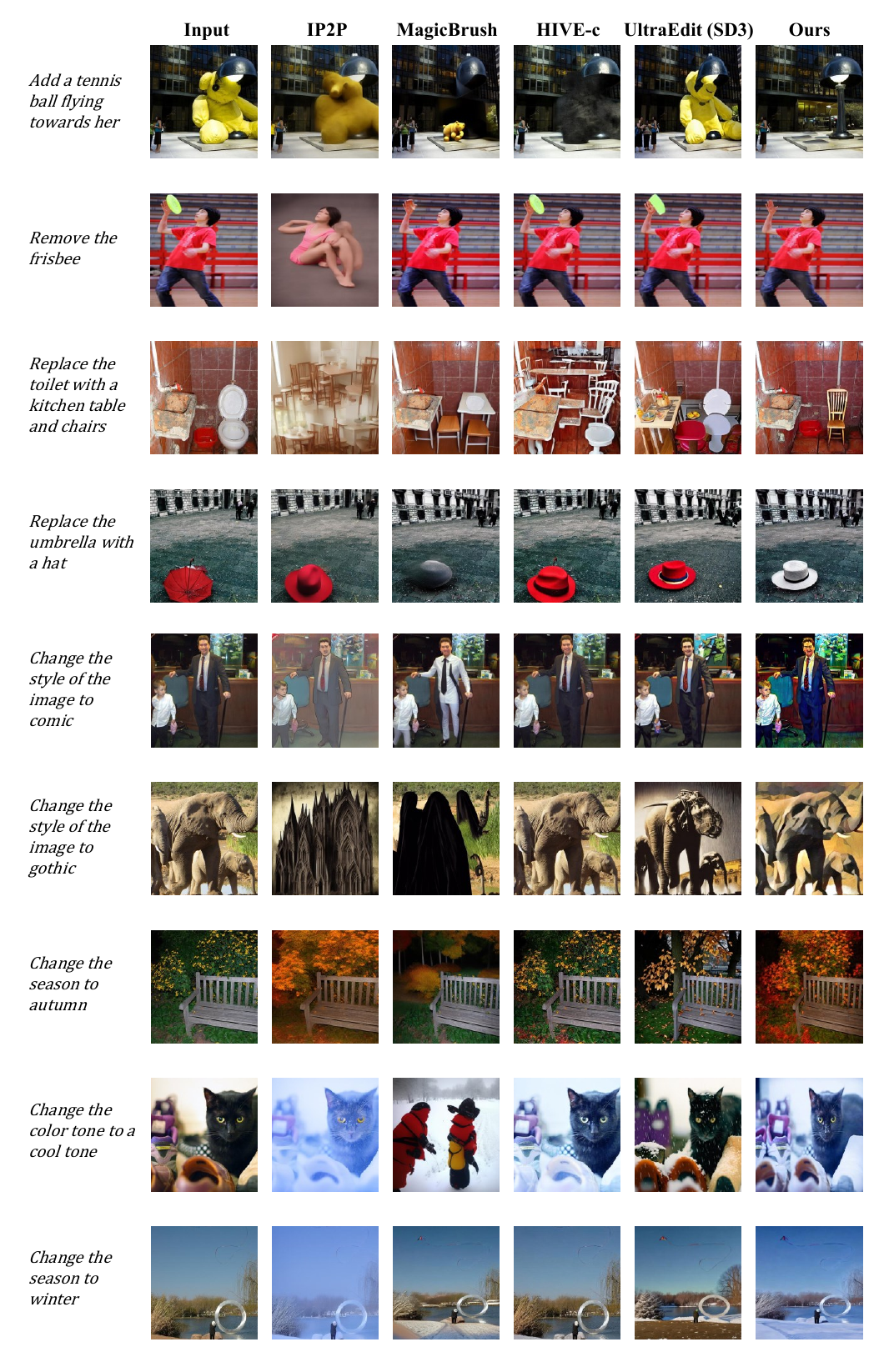}

    \caption{More qualitative evaluation of our model trained on AnyEdit across AnyEdit-Test benchmark~(Part II).}\label{append:editing_result_3}

\end{figure*}

\begin{figure*}[h]
    \centering
        \includegraphics[width=0.8\linewidth]{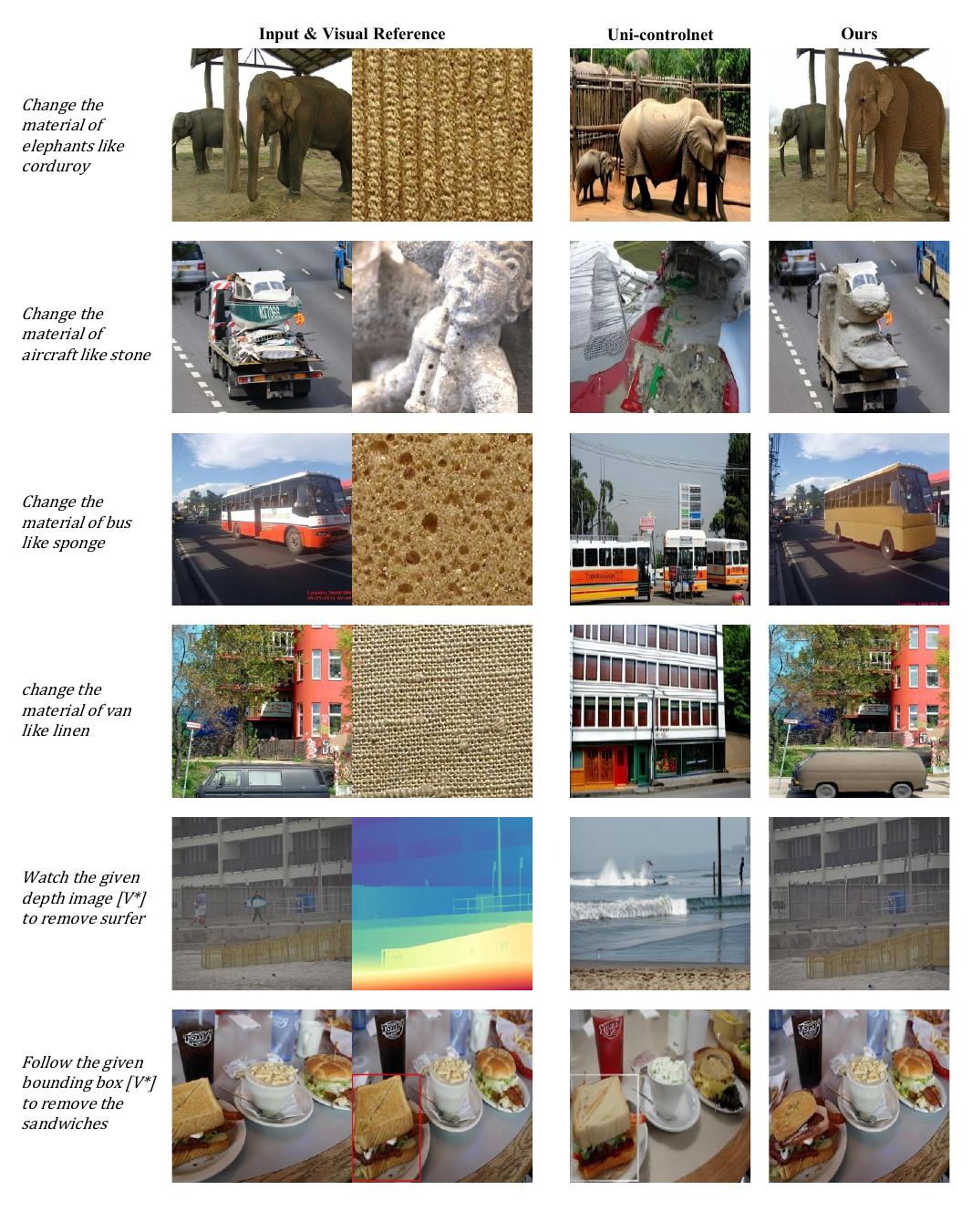}

    \caption{More qualitative evaluation of our model trained on AnyEdit across AnyEdit-Test benchmark~(Part III).}\label{append:editing_result_4}

\end{figure*}

\noindent\textbf{Detailed Results for AnyEdit-Test.}
More qualitative results of the AnyEdit-Test are shown in Figure~\ref{append:editing_result_2}, \ref{append:editing_result_3}, \ref{append:editing_result_4}.
% \zw{more example Figure on each benchmark}
% Figures 10 and 11 showcase qualitative examples of single-turn and multi-turn editing tasks on MagicBrush, while Figure 12 presents qualitative examples from the Emu Edit test across various editing tasks.

\begin{figure*}[h]
    \centering
        \includegraphics[width=1.\linewidth]{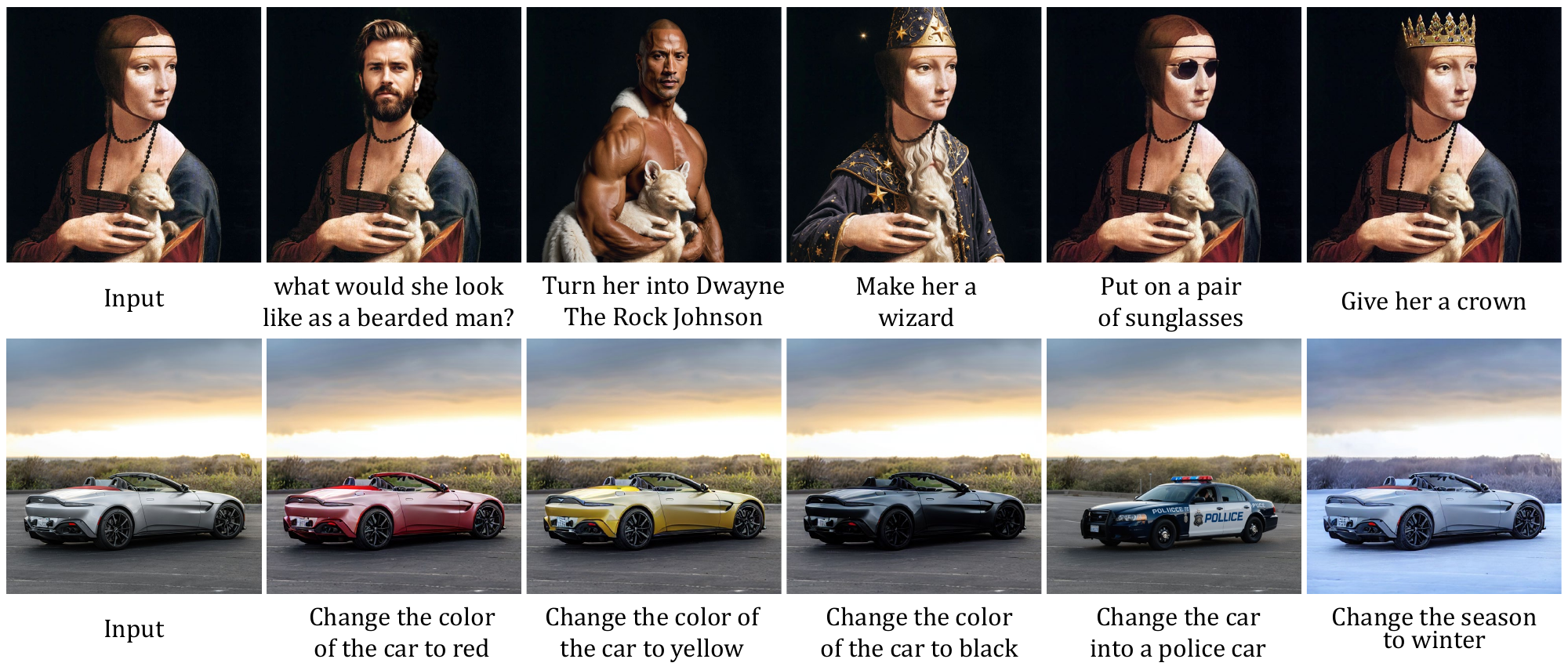}
        \includegraphics[width=1.\linewidth]{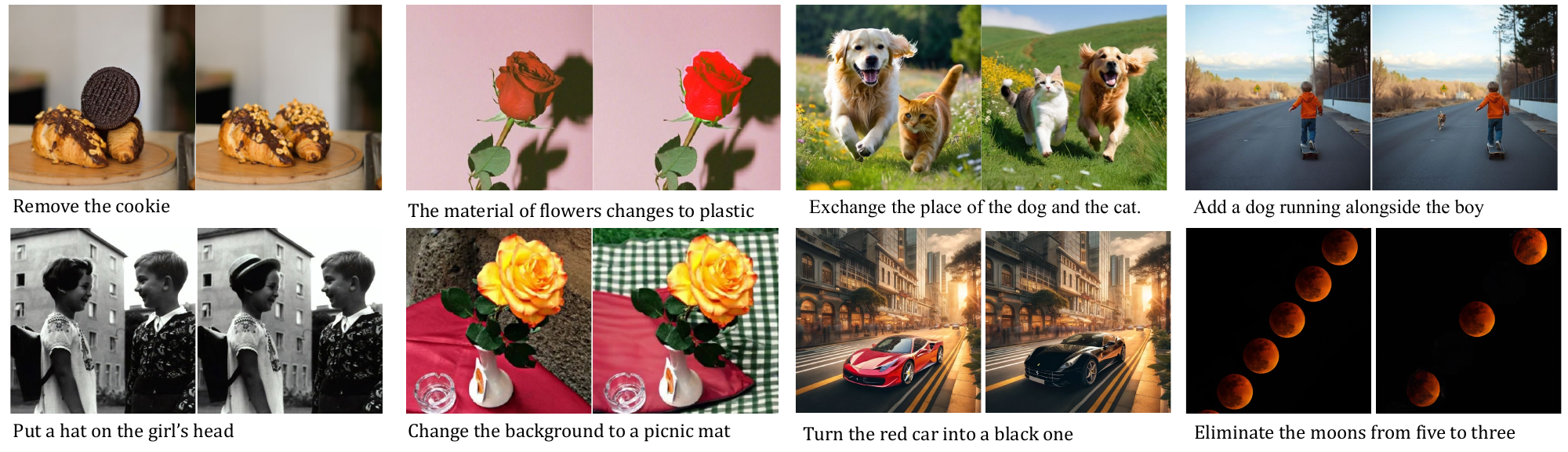}
        \includegraphics[width=1.\linewidth]{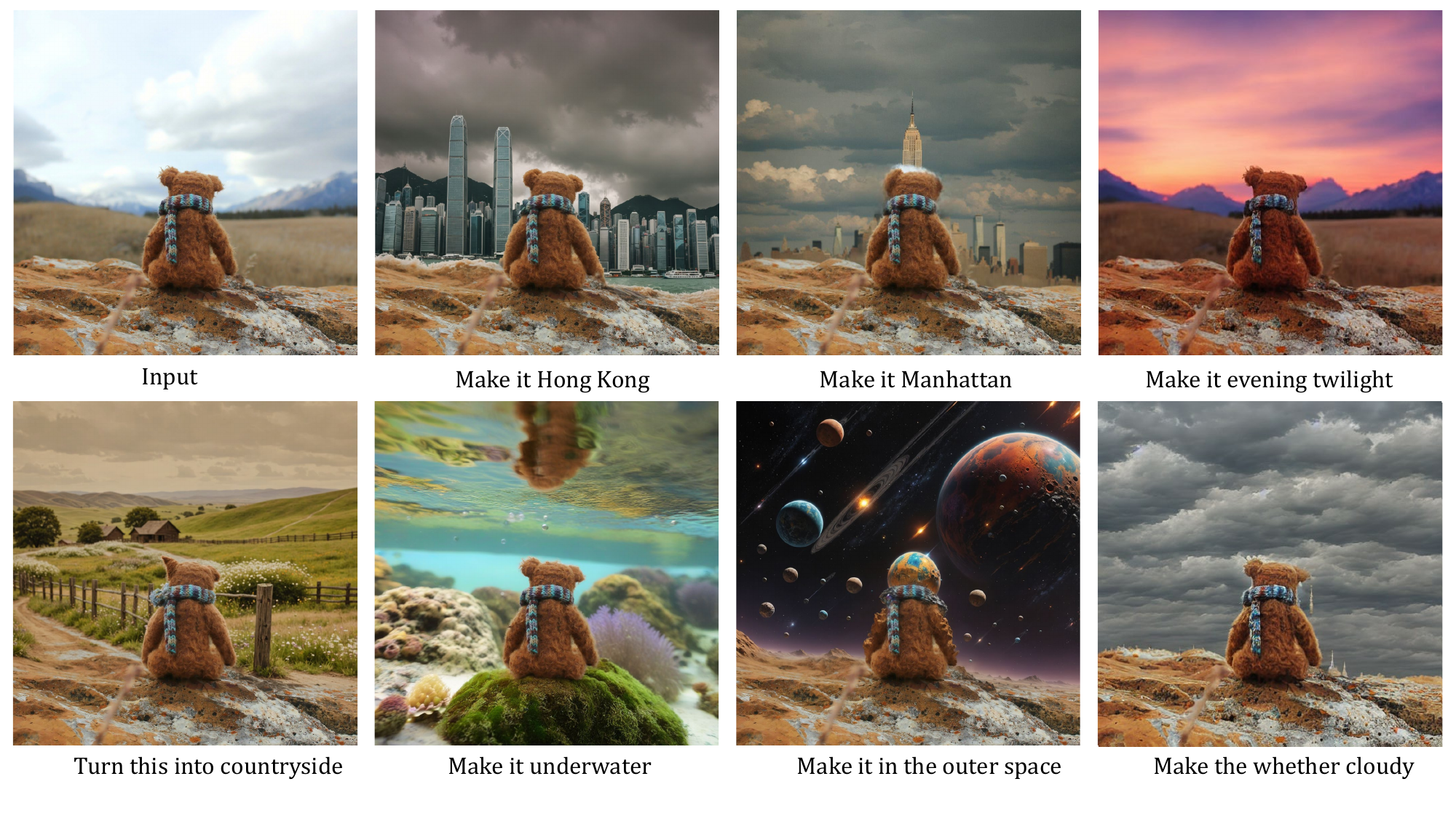}
      \vspace{-10mm}
    \caption{Qualitative evaluation of using real images as user inputs for the robustness of our editing model. }\label{fig:main_case1}
    \vspace{-8mm}
\end{figure*}

\noindent\textbf{More qualitative results of high-quality image editing from AnySD}. 
In Figure~\ref{fig:main_case1}, we visualize AnySD editing results on a wide variety of images.
We provide different editing instructions for the same image and observe that our method consistently achieves high-quality and fine-grained editing. For example, it successfully modifies underwater reflections and performs appearance modifications involving world knowledge. It effectively demonstrates the high quality of the AnyEdit dataset and the superiority of the AnySD architecture.

\begin{figure*}[ht]
    \centering
        \includegraphics[width=1.\linewidth]{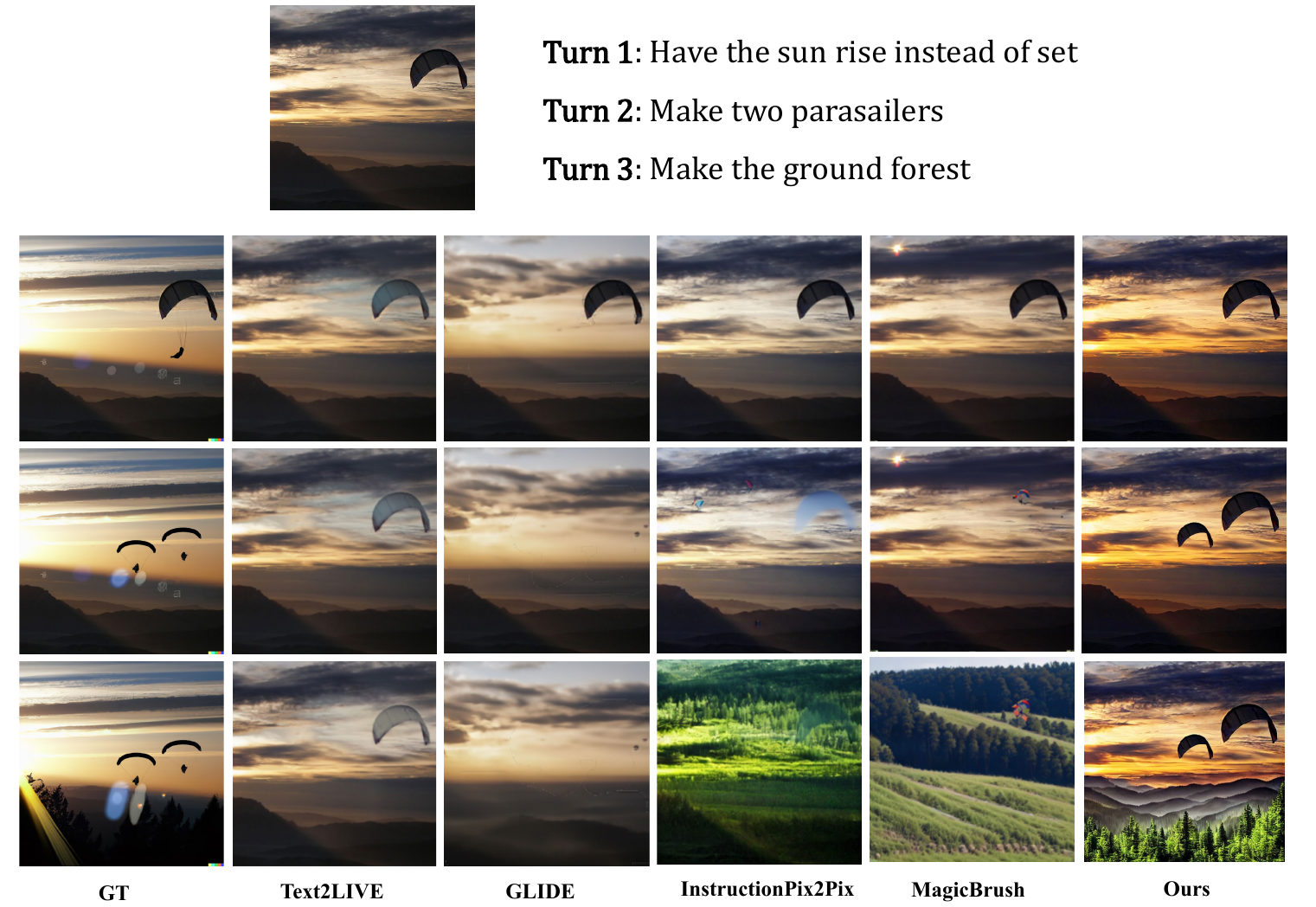}
    \caption{Qualitative evaluation of multi-turn editing scenario. We provide all baselines their desired input formats (Part I).}\label{mag:1}
  \vspace{-4mm}  
\end{figure*}

\begin{figure*}[h]
    \centering
        \includegraphics[width=1.\linewidth]{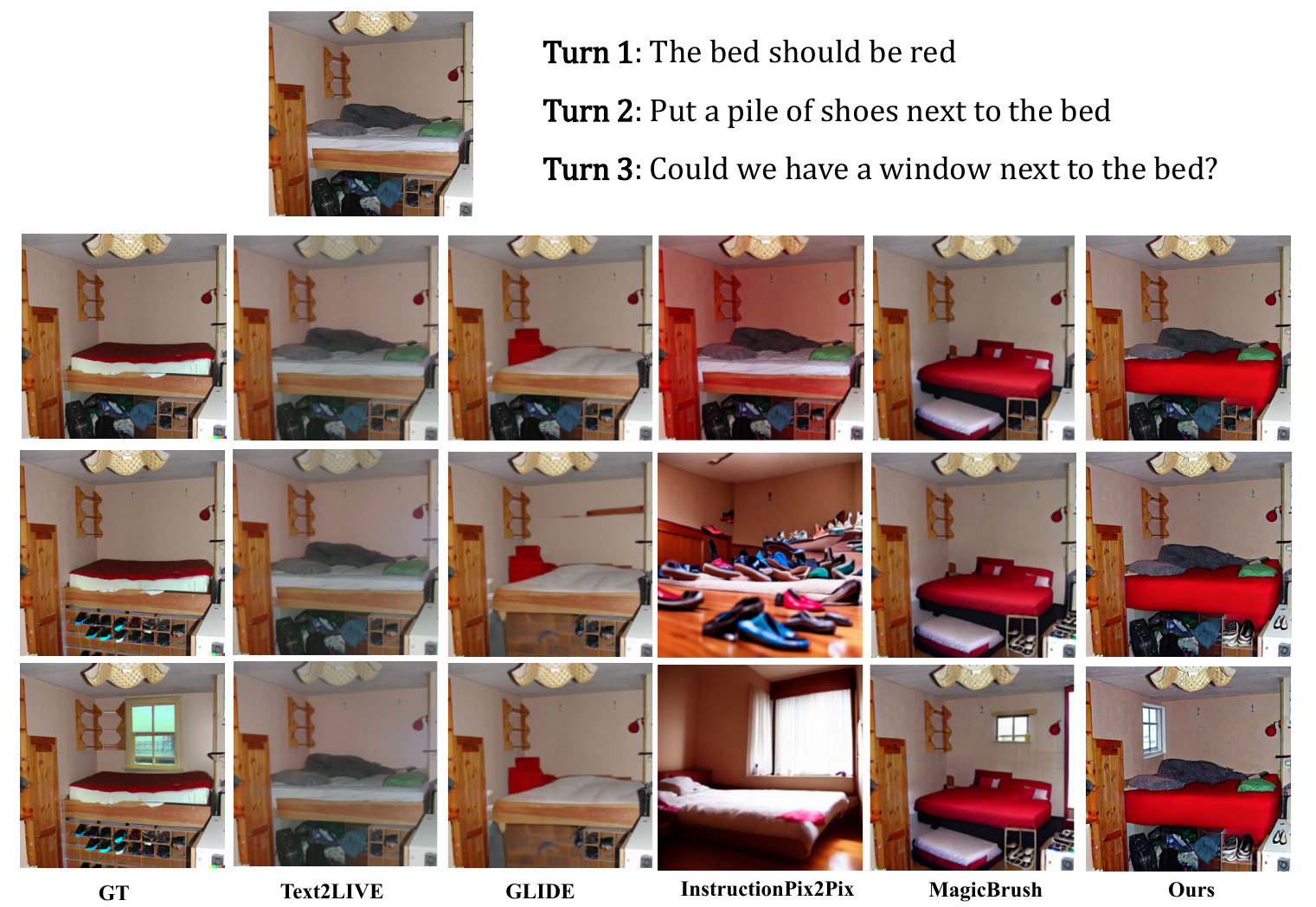}
    \caption{Qualitative evaluation of multi-turn editing scenario. We provide all baselines their desired input formats (Part II).}\label{mag:2}
  \vspace{-4mm}  
\end{figure*}

\noindent\textbf{Multi-turn in MagicBrush}. Figures~\ref{mag:1} and~\ref{mag:2} illustrate the performance of our AnySD model in multi-turn editing. Compared to Text2LIVE~\citep{xu2022designing}, GLIDE~\citep{nichol2021glide}, InstructPix2Pix, and MagicBrush, our model demonstrates stronger consistency, maintaining greater similarity to the original image even in the final editing rounds. Our results even surpass the ground truth provided by MagicBrush, further affirming the high quality of the AnyEdit dataset.

\begin{figure*}[h]
    \centering
        \includegraphics[width=1.\linewidth]{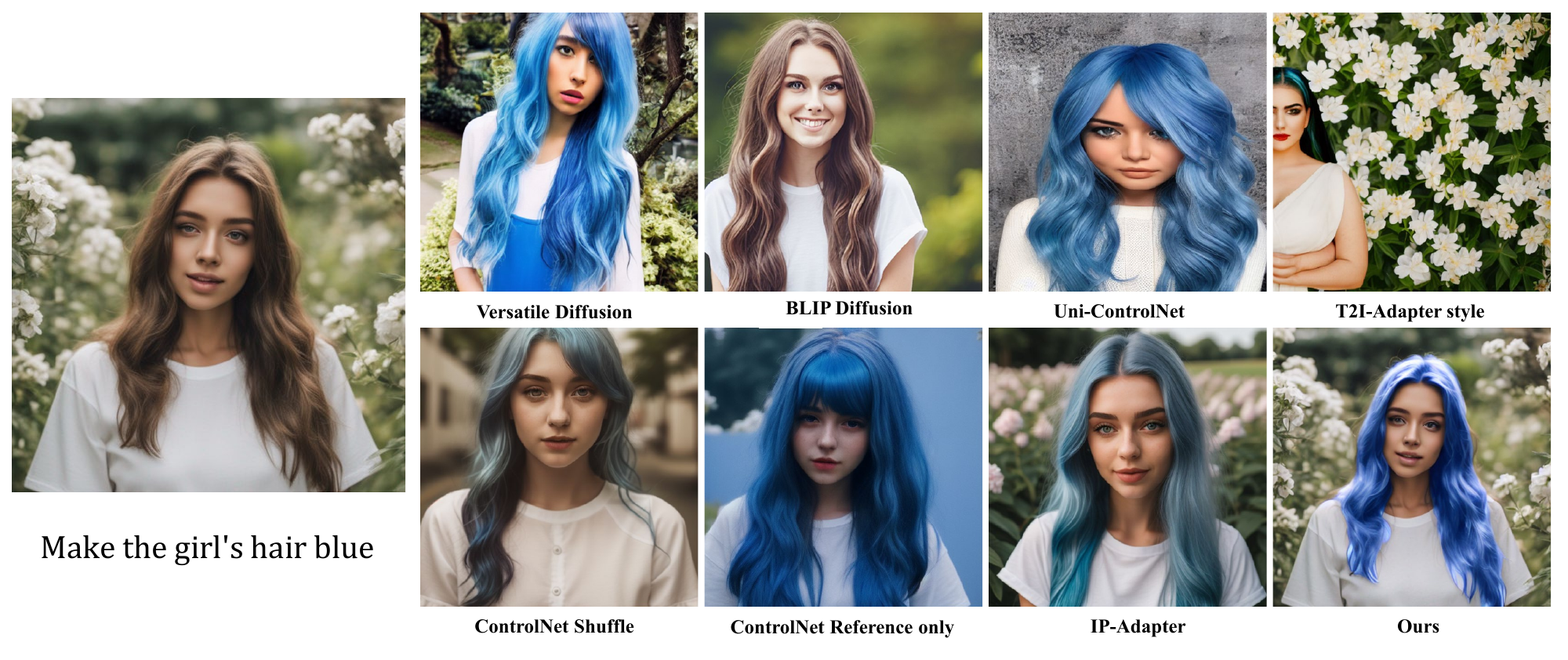}
        \vspace{5mm} 
        \includegraphics[width=1.\linewidth]{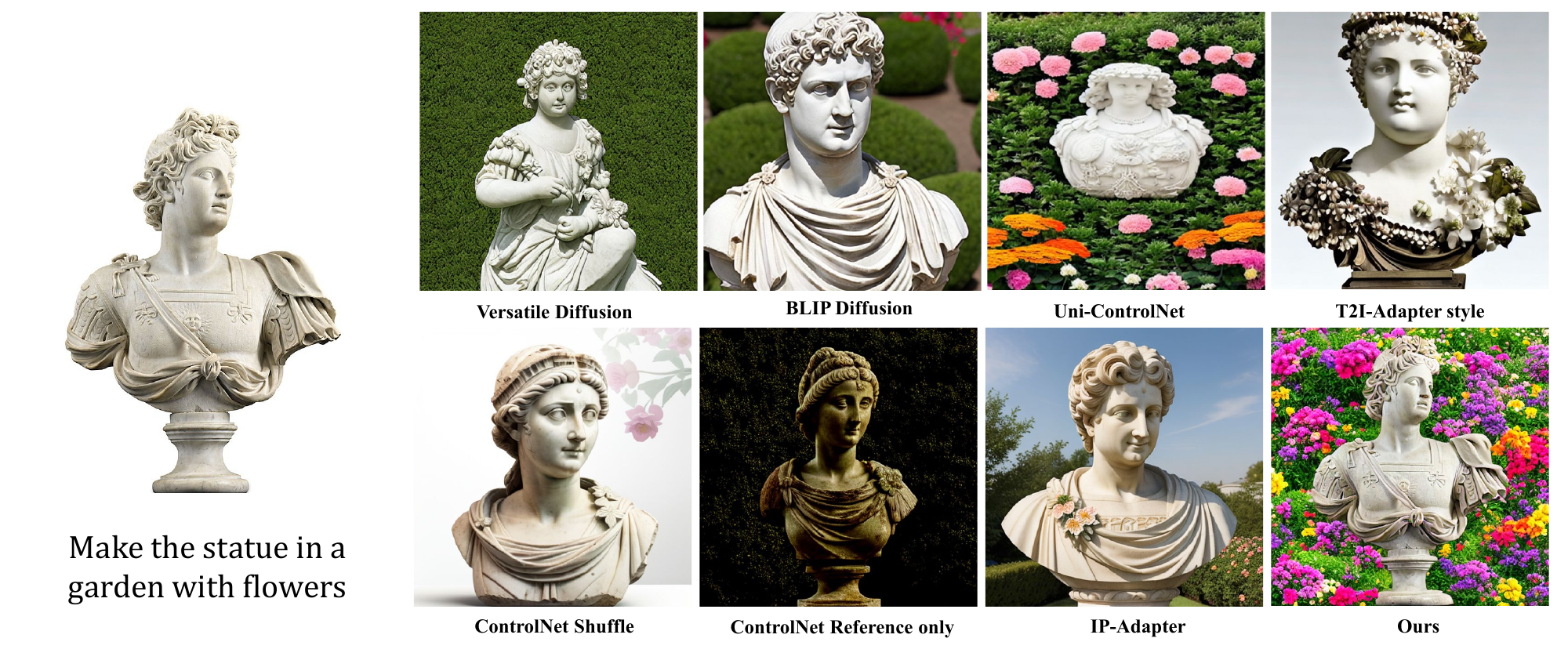}
    \caption{Comparison with more other image instruction edit methods (Part I).}\label{other:1}
  \vspace{-4mm}  
\end{figure*}

\begin{figure*}[h]
    \centering
        \includegraphics[width=1.\linewidth]{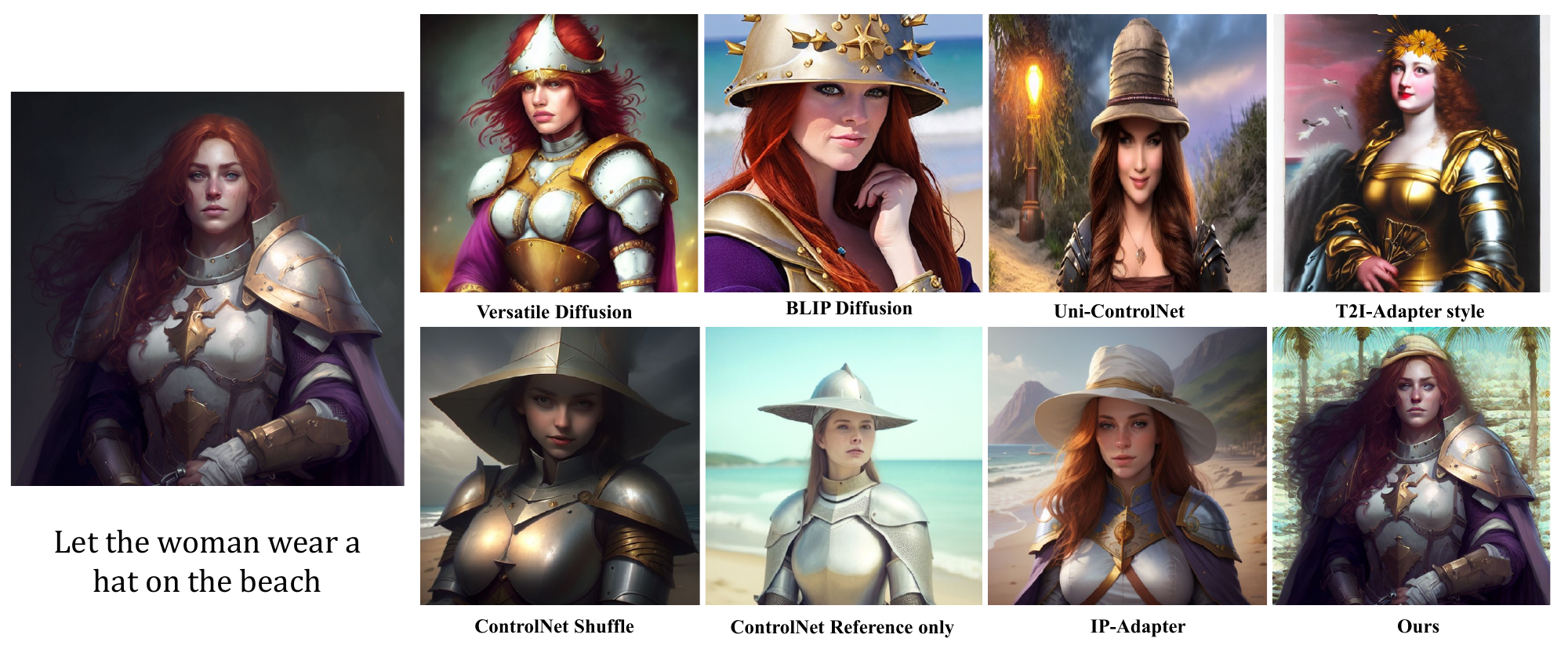}
        \vspace{5mm} 
        \includegraphics[width=1.\linewidth]{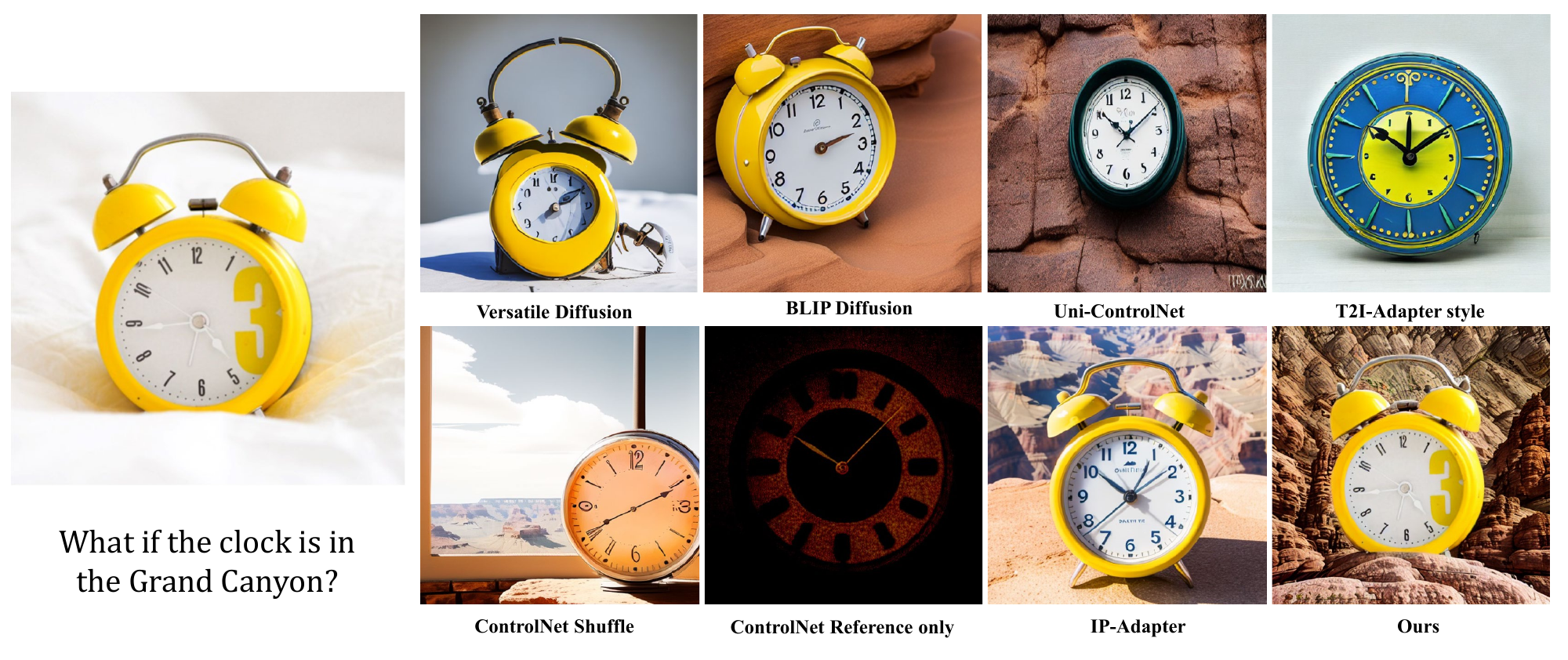}
    \caption{Comparison with more other image instruction edit methods (Part II).}\label{other:2}
  \vspace{-4mm}  
\end{figure*}

\noindent\textbf{Additional Image Editing Methods}. We also evaluate our image editing model in comparison with other approaches, including Versatile Diffusion~\citep{xu2023versatile}, BLIP Diffusion~\citep{li2023blip}, Uni-ControlNet~\citep{zhao2023uni}, T2I-Adapter~\citep{mou2024t2i}, ControlNet Shuffle~\cite{zhang2023adding}, ControlNet Reference-only~\cite{zhang2023adding}, and IP-Adapter~\citep{ye2023ip-adapter}. The comparison results are presented in Figure~\ref{other:1},~\ref{other:2}. Compared to other methods, our approach consistently produces superior results in terms of image quality and alignment with multimodal prompts.

\begin{table*}[h!]
\centering
\resizebox{\textwidth}{!}{ % 调整表格宽度为页面宽度
\begin{tabular}{ll}
\hline
 \textbf{Type} & \multicolumn{1}{c}{\textbf{Description}} \\\hline
 \textbf{Local Editing} &\\
 \quad Remove & Remove a specific object in the image and fill it with a background. \\
 \quad Replace & Replace a specific object in the image with a new object. \\
 \quad Add & Inserting a new object in the image. \\
 \quad Counting & Removing a specified number of objects to satisfy the number requirement. \\
 \quad Color Alter & Altering the color of specific objects in the image.\\
 \quad Appearance Alter & Altering the appearance~(\textit{e.g.}, decoration, texture, illumination) of specific objects in the image.\\
 \quad Action Change & Change the action of the specific object in the image. \\
 \quad Textual Change & Change the specific textual contents in the image to new textual contents \\
 \quad Material Change & Change the material of the specific object in the image. \\ \hline
 \textbf{Global Editing} & \\
\quad Background Change & Modifying the background of the entire image but not affecting the foreground objects. \\

 \quad Tone Transfer & Modifying the overall tone of the image, including changes in time, weather, and seasons. \\

\quad Style Change & Modifying the overall style of the image according to the given style word. \\\hline

\textbf{Camera Movement Editing} & \\
\quad Movement & Move the position viewpoint of a specific object in the image to the left or right or up or down. \\

\quad Resize & Zoom in or zoom out to a specific object in the image.\\

\quad Outpainting & Expanding the overall viewpoint by imagining the surroundings of the image elements of the mask. \\

\quad Rotation Change &  Rotate the overall viewpoint of the image to obtain images from different perspectives. \\ \hline
\textbf{Implicit Editing} & \\
\quad Implicit Change & Implicitly altering the contents of an image necessitates comprehension rather than explicit instructions.\\
\quad Relation Change & Change the position relationship between two objects~(\textit{i.e.}, swap their positions) in the image.\\\hline
\textbf{Visual Editing} & \\
\quad Image Reference & Replace a specific object in the image with the object in the reference image instead of any word.\\
\quad Material Transfer & Transfer the material in the reference image to the specific object in the image.\\
\quad Style Transfer & Transfer the style in the reference image to the specific image.\\
% ~(\textit{i.e.}, bounding box, scribble, segmentation, sketch, and depth image)
\quad Visual Bounding Box & Utilizing bounding box images as visual conditions images to guide the removal or replacement.\\
\quad Visual Scribble & Utilizing scribble images as visual conditions images to guide the removal or replacement.\\
\quad Visual Segmentation & Utilizing segmentation images as visual conditions images to guide the removal or replacement.\\
\quad Visual Sketch & Utilizing sketch images as visual conditions images to guide the removal or replacement.\\
\quad Visual Depth & Utilizing depth images as visual conditions images to guide the removal or replacement.\\

\hline
\end{tabular}
}
\caption{The Definition of Editing Instruction Types in AnyEdit.}\label{dataset_info}
\end{table*}

\begin{table*}[!t]
\centering
\begin{tabular}{>{\raggedright\arraybackslash}m{4cm}>{\raggedright\arraybackslash}m{10cm}}
\hline
\textbf{Element} & \multicolumn{1}{c}{\textbf{Content}} \\
\hline
\textbf{System Prompt} & You are an assistant that only speaks JSON. Do not write normal text. The assistant answer is JSON with the following string fields: \texttt{edit}, \texttt{edited object}, \texttt{output}. Here is the latest conversation between the Assistant and the User. \\
\hline
\textbf{Task Description (Model)} & Hi, My job is to take a given caption (\texttt{input}) and to output the following: an instruction for adding an object to the image (\texttt{edit}), the object to add (\texttt{edited object}), and the caption with the object (\texttt{output}). Please help me do it. I will give you the \texttt{input}, and you will help. \\
\hline
\textbf{Output Format} & When you reply, use the following format: \texttt{\{'edit': '<instruction>', 'edited object': '<object>', 'output': '<caption>'\}}. Construct the instruction with one of the following instruction words: [\texttt{place}, \texttt{add}, \texttt{include}]. Don’t include any \texttt{\textbackslash} or edit any actions in the instruction. \\
\hline
\textbf{Initial Message from LLM} & Sure, I'd be happy to help! Just provide me with the \texttt{input} (the original caption), and I'll generate the instruction, edited object, and output caption for you. Let's get started! \\
\hline
\textbf{Example Input} & User input: Beautiful cat with mojito sitting in a cafe on the street. \\
\hline
\textbf{Example Output from LLM} & \{'edit': 'add a hat to the cat', 'edited object': 'hat', 'output': 'Beautiful cat wearing a hat with mojito sitting in a cafe on the street.'\}  \\
\hline
\multicolumn{2}{c}{\textbf{Constrained Editing Instruction Generation}} \\\hline
\textbf{Input} & \texttt{\{caption\}} from real-world image-text pairs or counterfactual synthetic datasets with \texttt{\{editing type\}} requirements from users \\
\hline
\textbf{Response} & \texttt{\{model generated edit instructions in JSON format\}}. An example of editing data is: \{
  "edit": "change the airplane to green",  
  "edited object": "airplane", 
  "input": "a small airplane sits stationary on a piece of concrete.",
  "output": "A green small airplane sits stationary on a piece of concrete.",
  "edit type": "color alter",
  "visual input": "None", 
  "image file": "COCO-train2014-000000521165.jpg",
  "edited file": "xxxxx.jpg"  
\}" \\
\hline
\end{tabular}\caption{Prompts constraints for LLMs to write edit instructions and captions.}\label{prompt_template}
\end{table*}

\begin{figure*}[ht]
    \centering
        \includegraphics[width=0.9\linewidth]{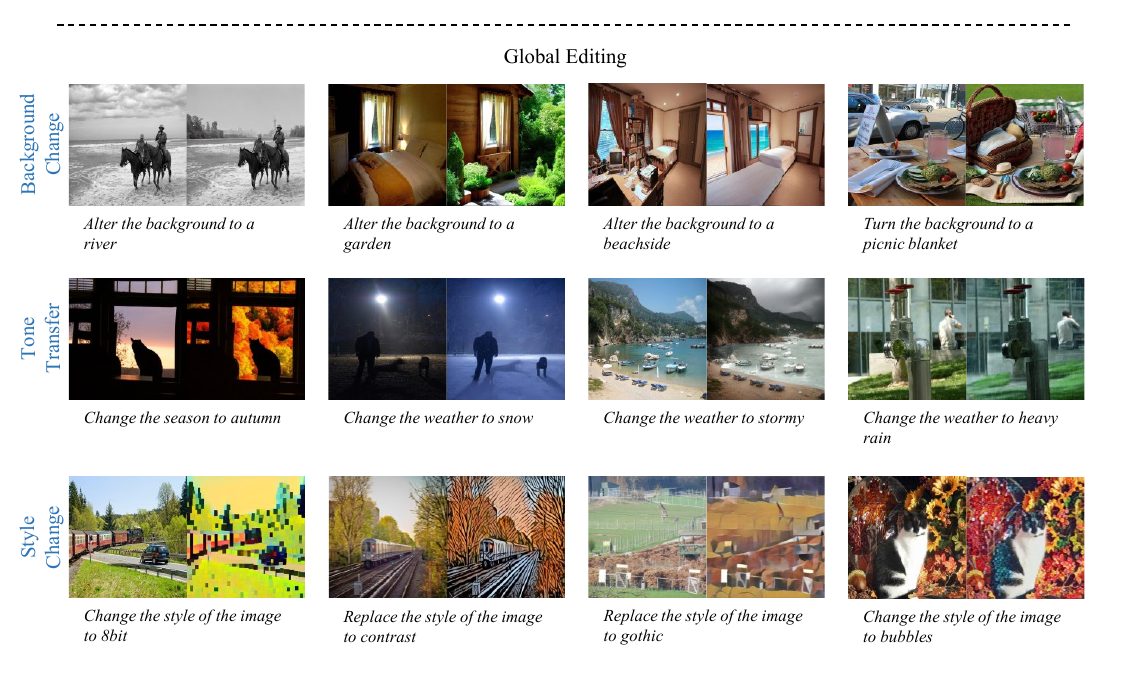}
  \vspace{-4mm}
  \caption{More Examples of AnyEdit-Test with global editing categories.}
\end{figure*}
\begin{figure*}[ht]
    \centering
        \includegraphics[width=0.9\linewidth]{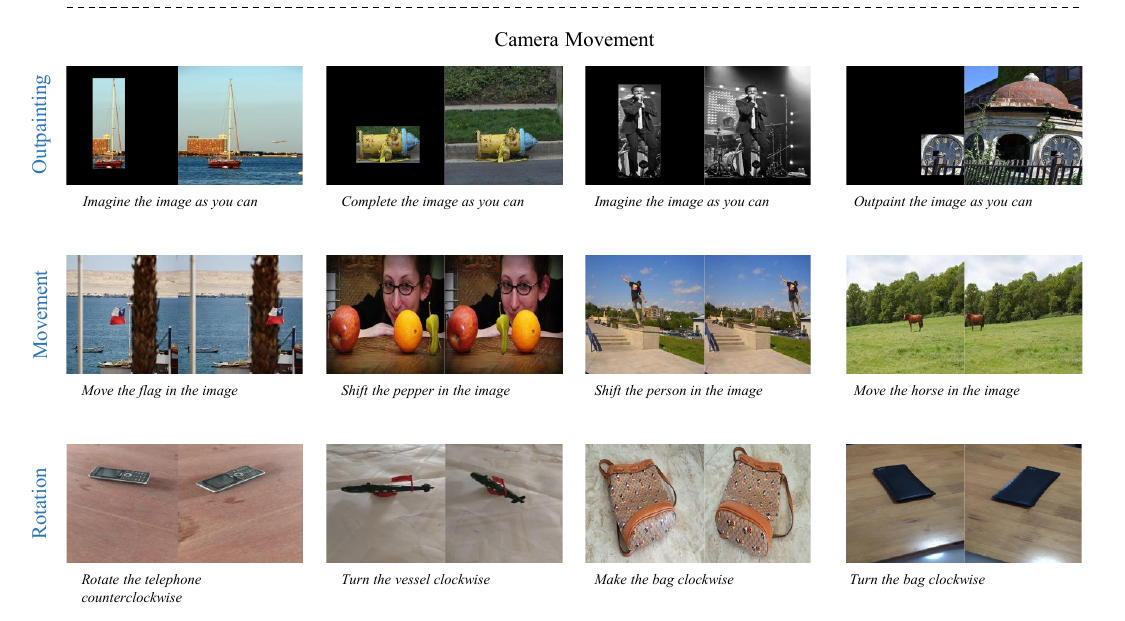}
  \vspace{-4mm}
  \caption{More Examples of AnyEdit-Test with camera movement editing categories.}
\end{figure*}
\begin{figure*}[ht]
    \centering
        \includegraphics[width=0.9\linewidth]{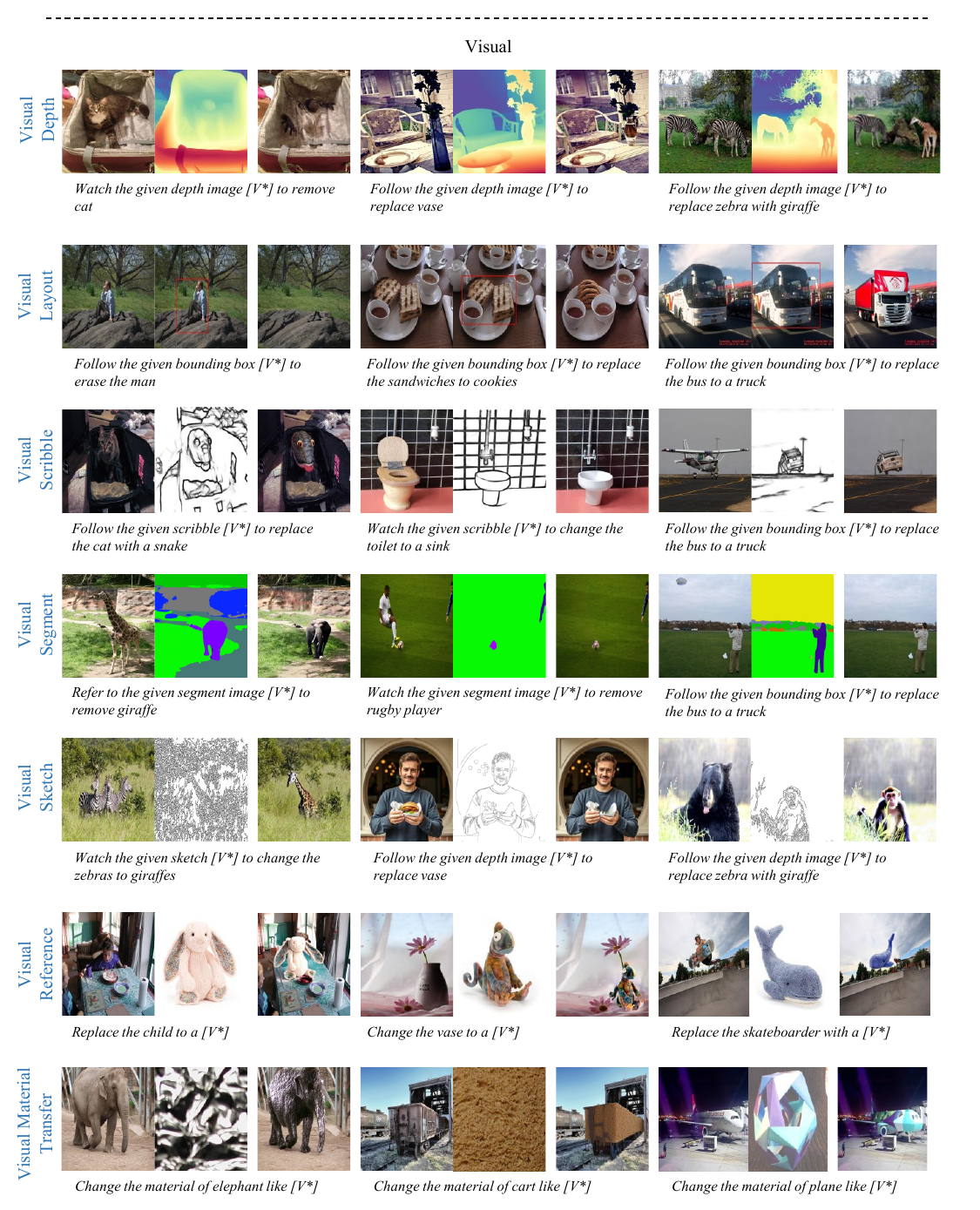}
  \vspace{-4mm}
  \caption{More Examples of AnyEdit-Test with visual editing categories.}
\end{figure*}
\clearpage
\begin{table*}[t]
\centering
\resizebox{0.99\textwidth}{!}{
    \begin{tabular}{lccccccccc}
    \hline
        \multirow{2}{*}{} & \multicolumn{9}{c}{local} \\ \cline{2-10} 
                          & remove & replace & add & color & appearance & material change & action & textual & counting \\ \hline
        \multicolumn{10}{l}{\textbf{InstructPix2Pix}~\citep{brooks2023instructpix2pix}} \\
        CLIPim $\uparrow$  &0.664 &0.779 &0.832 &0.862 &0.770 &0.700 &0.674 &0.744 &0.803 \\
        CLIPout $\uparrow$ &0.227 &0.276 &0.302 &\underline{0.318} &0.308 &- &0.228 &0.298 & \underline{0.272}\\
        L1 $\downarrow$    &0.146 &0.188 &0.134 &0.162 &0.160 &0.168 &0.167 &0.190 &0.149 \\
        DINO $\uparrow$    &0.408 &0.537 &0.706 &0.773 &0.593 &0.369 &0.413 &0.694 &0.590 \\ \hline

        \multicolumn{10}{l}{\textbf{MagicBrush}~\citep{zhang2024magicbrush}} \\
        CLIPim $\uparrow$  &\underline{0.849} &\underline{0.814}&0.930 &0.826 &0.843 &\underline{0.809} &0.754 & 0.759&0.875 \\
        CLIPout $\uparrow$ &\underline{0.264} &0.289&\underline{0.321} &0.305 &0.319 &-     &0.272 & 0.312&0.264 \\
        L1 $\downarrow$    &\textbf{0.076} &\underline{0.143}&0.071 &0.112 &\underline{0.084} &0.111 &0.203 & 0.157&0.100 \\
        DINO $\uparrow$    &\underline{0.783} &0.604&\underline{0.897} &0.667 &0.739 &0.570 &0.548 & 0.774&0.731 \\ \hline

        \multicolumn{10}{l}{\textbf{HIVE$^w$}~\citep{zhang2023hive}} \\
        CLIPim $\uparrow$  &0.750 &0.788 &0.914 &0.853 &0.819 &0.764 &0.826 &0.801 &0.866 \\
        CLIPout $\uparrow$ &0.237 &0.282 &0.312 &0.307 &\underline{0.313} &-     &0.291 &0.318 &0.266 \\
        L1 $\downarrow$    &0.118 &0.184 &0.079 &0.114 &0.147 &0.126 &0.155 &0.139 &0.122 \\
        DINO $\uparrow$    &0.586 &0.600 &0.857 &0.779 &0.690 &0.536 &0.735 &0.838 &0.738 \\ \hline

        \multicolumn{10}{l}{\textbf{HIVE$^c$}~\citep{zhang2023hive}} \\
        CLIPim $\uparrow$  &0.823 &0.778 &\underline{0.932} &\textbf{0.894} &\underline{0.864} &0.785 &\textbf{0.874} &\underline{0.807} &\textbf{0.899} \\
        CLIPout $\uparrow$ &0.254 &0.284 &0.312 &0.309 &0.309 &-     &\textbf{0.308} &\underline{0.319} &0.267 \\
        L1 $\downarrow$    &\underline{0.099} &0.167 &\underline{0.066} &0.097 &0.105 &\underline{0.103} &\underline{0.147} &\underline{0.129} &0.100 \\
        DINO $\uparrow$    &0.728 &0.584 &0.891 &\underline{0.850} &\underline{0.795} &\underline{0.594} &\textbf{0.811} &\underline{0.871} &\underline{0.800} \\ \hline
        
        \multicolumn{10}{l}{\textbf{UltraEdit (SD3)}~\citep{zhao2024ultraedit}} \\
        CLIPim $\uparrow$  &0.806 &0.805 &0.925 &0.851 &0.817 &0.764 &0.827 &\textbf{0.854} & 0.880 \\
        CLIPout $\uparrow$ &0.262 &\textbf{0.295} &0.323 &\textbf{0.320} &\textbf{0.320} &- &0.292 &\textbf{0.344 }&\textbf{0.273} \\
        L1 $\downarrow$    &0.087 &0.151 &0.072 &\underline{0.091} &0.100 &0.108 &0.158 &\textbf{0.127} &\underline{0.089} \\
        DINO $\uparrow$    &0.709 &\underline{0.615} &0.867 &0.791 &0.729 &0.522 &0.724 &\textbf{0.890} &0.764 \\ \hline

        \multicolumn{10}{l}{\textbf{Null-Text}~\citep{mokady2023null}} \\
        CLIPim $\uparrow$  &0.752 &0.710 &- &0.814 &0.785 &- &0.838 &0.764 &- \\
        CLIPout $\uparrow$ &0.250 &0.247 &- &0.274 &0.285 &- &0.298 &0.305 &- \\
        L1 $\downarrow$    &0.235 &0.253 &- &0.227 &0.239 &- &0.243 &0.275 &- \\
        DINO $\uparrow$    &0.598 &0.384 &- &0.695 &0.675 &- &0.732 &0.764 &- \\ \hline

        \multicolumn{10}{l}{\textbf{AnySD w/ AnyEdit (Ours)}} \\
        CLIPim $\uparrow$  &\textbf{0.851} &\textbf{0.853} &\textbf{0.946} &\underline{0.896} &\textbf{0.877} &\textbf{0.811} &\underline{0.873} &0.763 &\underline{0.898} \\
        CLIPout $\uparrow$ &\textbf{0.265} &\underline{0.292} &\textbf{0.322} &0.313 &0.309 &- &\underline{0.306} &0.303 &0.263 \\
        L1 $\downarrow$    &0.103 &\textbf{0.123} &\textbf{0.052} &\textbf{0.061} &\textbf{0.051} &\textbf{0.084} &\textbf{0.145} &0.136 &\textbf{0.088} \\
        DINO $\uparrow$    &\textbf{0.785} &\textbf{0.688} &\textbf{0.921} &\textbf{0.855} &\textbf{0.840} &\textbf{0.602} &\underline{0.782} &0.800 &\textbf{0.819} \\ \hline

    \end{tabular}
}
\caption{\textbf{Comparison of Methods on AnyEdit-Test (Part 1)}. '-' indicates 'not applicable'.}\label{app:exp2_table1}
\end{table*}

\clearpage
\begin{table*}[t]
\centering
\resizebox{0.99\textwidth}{!}{
    \begin{tabular}{lccccccccc}
    \hline
        \multirow{2}{*}{} & \multicolumn{3}{c}{global} & \multicolumn{4}{c}{camera} & \multicolumn{2}{c}{implicit} \\ \cline{2-10} 
                          & background & tone transfer & style change & movement & outpaint & rotation & resize & implicit & relation \\ \hline
        \multicolumn{10}{l}{\textbf{InstructPix2Pix}~\citep{brooks2023instructpix2pix}} \\
        CLIPim $\uparrow$  &0.680 &\textbf{0.860} &0.702 &0.805 &0.563 &0.675 &0.755 &0.762 &0.826 \\
        CLIPout $\uparrow$ &0.259 &\textbf{0.304} &- &- &- &- &- &- &\underline{0.288} \\
        L1 $\downarrow$    &0.221 &\textbf{0.098} &0.221 &0.131 &\underline{0.290} &0.148 &0.141 &0.212 &0.167 \\
        DINO $\uparrow$    &0.411 &\underline{0.804} &0.354 &0.639 &0.341 &0.361 &0.566 &0.538 &0.577 \\ \hline

        \multicolumn{10}{l}{\textbf{MagicBrush}~\citep{zhang2024magicbrush}} \\
        CLIPim $\uparrow$  &0.739 &0.789 &0.664 &0.863 &0.561 &0.791 &0.845 &0.819 &\underline{0.910} \\
        CLIPout $\uparrow$ &0.268 &0.287 &-     &-     &-     &-     &-     &- &0.280 \\
        L1 $\downarrow$    &0.233 &0.213 &0.252 &\underline{0.093} &0.353 &0.134 &0.101 &0.189 &0.109 \\
        DINO $\uparrow$    &0.529 &0.657 &0.292 &0.710 &0.344 &0.575 &0.725 &0.622 &0.800 \\ \hline

        \multicolumn{10}{l}{\textbf{HIVE$^w$}~\citep{zhang2023hive}} \\
        CLIPim $\uparrow$  &0.764 &0.816 &0.706 &\underline{0.872} &0.582 &0.774 &0.888 &0.784 &0.858 \\
        CLIPout $\uparrow$ &0.280 &0.293 &-     &-     &-     &-     &-     &- &0.284 \\
        L1 $\downarrow$    &0.202 &0.175 &0.212 &0.131 &0.328 &0.135 &0.107 &0.202 &0.119 \\
        DINO $\uparrow$    &0.635 &0.719 &0.383 &0.732 &0.328 &0.620 &0.796 &0.572 &0.697 \\ \hline

        \multicolumn{10}{l}{\textbf{HIVE$^c$}~\citep{zhang2023hive}} \\
        CLIPim $\uparrow$  &\textbf{0.822} &0.833 &0.705 &\textbf{0.926} &0.665 &\textbf{0.848} &\textbf{0.912} &\underline{0.809} &\textbf{0.914} \\
        CLIPout $\uparrow$ &\underline{0.294} &0.293 &-     &-     &-     &-     &-     &- &0.284 \\
        L1 $\downarrow$    &0.177 &0.182 &0.401 &0.112 &0.349 &\underline{0.129} &0.093 &0.180 &\underline{0.093} \\
        DINO $\uparrow$    &\textbf{0.777} &0.748 &0.202 &\textbf{0.866} &0.428 &\textbf{0.739} &\textbf{0.861} &0.627 &\textbf{0.829} \\ \hline

        \multicolumn{10}{l}{\textbf{UltraEdit (SD3)}~\citep{zhao2024ultraedit}} \\
        CLIPim $\uparrow$  &0.790 &0.795 &\textbf{0.730} &0.867 &\textbf{0.705} &0.765 &0.872 &\textbf{0.825} &0.887 \\
        CLIPout $\uparrow$ &0.293 &0.301 &-     &-     &-     &-     &-     &- &0.281 \\
        L1 $\downarrow$    &\underline{0.181} &0.184 &\underline{0.208} &0.106 &0.372 &0.139 &\underline{0.086} &\underline{0.176} &\underline{0.093} \\
        DINO $\uparrow$    &0.701 &0.709 &0.448 &0.762 &\underline{0.612} &0.523 &0.813 &\underline{0.642} &0.764 \\ \hline

        \multicolumn{10}{l}{\textbf{Null-Text}~\citep{mokady2023null}} \\
        CLIPim $\uparrow$  &0.755 &0.750 &-     &-     &-     &-     &-     &- &-  \\
        CLIPout $\uparrow$ &0.285 &0.269 &-     &-     &-     &-     &-     &- &-  \\
        L1 $\downarrow$    &0.251 &0.289 &-     &-     &-     &-     &-     &- &-  \\
        DINO $\uparrow$    &0.617 &0.608 &-     &-     &-     &-     &-     &- &-  \\ \hline

        \multicolumn{10}{l}{\textbf{AnySD w/ AnyEdit (Ours)}} \\
        CLIPim $\uparrow$  &\underline{0.819} &\underline{0.836} &\underline{0.710} &0.870 &\underline{0.738} &\underline{0.826} &\underline{0.898} &\textbf{0.825} &0.908 \\
        CLIPout $\uparrow$ &\textbf{0.300} &\underline{0.302} &-     &-     &-     &-     &-     &- &\textbf{0.289} \\
        L1 $\downarrow$    &\textbf{0.169} &\underline{0.115} & \textbf{0.192} &\textbf{0.069} &\textbf{0.189} &\textbf{0.122} &\textbf{0.060} &\textbf{0.169} &\textbf{0.091} \\
        DINO $\uparrow$    &\underline{0.744} &\textbf{0.811} &\underline{0.385} &\underline{0.782} &\textbf{0.682} &\underline{0.685} &\underline{0.832} &\textbf{0.643} &\underline{0.822} \\ \hline

    \end{tabular}
}
\caption{\textbf{Comparison of Methods on AnyEdit-Test (Part 2)}. '-' indicates 'not applicable'.}\label{app:exp2_table2}
\end{table*}

\begin{table*}[t]
\centering
\resizebox{0.99\textwidth}{!}{
    \begin{tabular}{lccccccc}
    \hline
        \multirow{2}{*}{} & \multicolumn{7}{c}{Visual} \\ \cline{2-8} 
                           & visual depth & visual sketch & visual scribble & visual segment & visual bbox  & material transfer & visual reference \\ \hline
        \multicolumn{8}{l}{\textbf{Uni-controlnet}~\citep{zhao2023uni}} \\
        CLIPim $\uparrow$  &0.741 &0.763 &0.770 &0.716 &0.734 &0.642 &0.652 \\
        CLIPout $\uparrow$ &0.246 &0.259 &0.253 &0.246 &0.253 &- &0.234 \\
        L1 $\downarrow$    &0.271 &0.247 &0.254 &0.281 &0.214 &0.278 &0.275 \\
        DINO $\uparrow$    &0.503 &0.576 &0.531 &0.421 &0.512 &0.241 &0.308  \\ \hline

        \multicolumn{8}{l}{\textbf{AnySD w/ AnyEdit (Ours)}} \\
        CLIPim $\uparrow$  &\textbf{0.780} &\textbf{0.803} &\textbf{0.805} &\textbf{0.770} &\textbf{0.811} &\textbf{0.849} &\textbf{0.714} \\
        CLIPout $\uparrow$ &\textbf{0.250} &\textbf{0.268} &\textbf{0.258} &\textbf{0.252} &\textbf{0.258} &- &\textbf{0.260} \\
        L1 $\downarrow$    &\textbf{0.177} &\textbf{0.164} &\textbf{0.158} &\textbf{0.181} &\textbf{0.125} &\textbf{0.090} &\textbf{0.121} \\
        DINO $\uparrow$    &\textbf{0.612} &\textbf{0.663} &\textbf{0.627} &\textbf{0.607} &\textbf{0.687} &\textbf{0.712} &\textbf{0.488}  \\ \hline

    \end{tabular}
}
\caption{\textbf{Comparison of Methods on AnyEdit-Test (Part 3)}. '-' indicates 'not applicable'.}\label{app:exp2_table3}
\end{table*}
\clearpage
\clearpage

\clearpage
\clearpage
\vspace{-8mm}
\clearpage


\clearpage
% {
%     \small
%     \bibliographystyle{ieeenat_fullname}
%     \bibliography{main}
% }
%%WARNING: do not forget to delete the supplementary pages from your submission 
%% \input{sec/X_suppl}

\end{document}